\pdfoutput=1

\documentclass[11pt]{article}

\usepackage[final]{acl}

\usepackage{times}
\usepackage{latexsym}

\usepackage[T2A,T2B,T2C,T1]{fontenc}

\usepackage[utf8]{inputenc}

\usepackage{microtype}

\usepackage{inconsolata}

\usepackage{graphicx}

\definecolor{LightGray}{rgb}{0.93, 0.93, 0.93}
\usepackage{tabularx}

\usepackage[russian,polish,swedish,norsk,hungarian,english]{babel}

\usepackage{booktabs}

\usepackage{todonotes}

\usepackage[most]{tcolorbox} 

\usepackage{makecell}
\usepackage{booktabs}
\usepackage{multirow}
\usepackage{amsmath}
\usepackage{graphicx}
\usepackage{adjustbox} 

\usepackage{longtable}
\title{Breaking Language Barriers in Visual Language Models via Multilingual Textual Regularization}

\author{
 \textbf{Iñigo Pikabea\textsuperscript{*1,2}},
 \textbf{Iñaki Lacunza\textsuperscript{*1}},
 \textbf{Oriol Pareras\textsuperscript{*1}},
\\
 \textbf{Carlos Escolano\textsuperscript{1,2}},
 \textbf{Aitor Gonzalez-Agirre\textsuperscript{1}},
 \textbf{Javier Hernando\textsuperscript{1,2}},
 \textbf{Marta Villegas\textsuperscript{1}},
\\
 \textsuperscript{1} Barcelona Supercomputing Center,
 \textsuperscript{2} Universitat Politècnica de Catalunya
\\
 \small{
   \textbf{Correspondence:} \{inigo.pikabea,inaki.lacunza,oriol.pareras\}@bsc.es
 }
$^*$Core contributors
}

\begin{document}
\maketitle
\begin{abstract}
    Rapid advancements in Visual Language Models (VLMs) have transformed multimodal understanding but are often constrained by generating English responses regardless of the input language. This phenomenon has been termed as Image-induced Fidelity Loss (IFL) and stems from limited multimodal multilingual training data. To address this, we propose a continuous multilingual integration strategy that injects text-only multilingual data during visual instruction tuning, preserving the language model’s original multilingual capabilities. Extensive evaluations demonstrate that our approach significantly improves linguistic fidelity across languages without degradation in visual performance. We also explore model merging, which improves language fidelity but comes at the cost of visual performance. In contrast, our core method achieves robust multilingual alignment without trade-offs, offering a scalable and effective path to mitigating IFL for global VLM adoption.
\end{abstract}

\section{Introduction}

\begin{figure}
    \centering
    \includegraphics[width=0.87\linewidth]{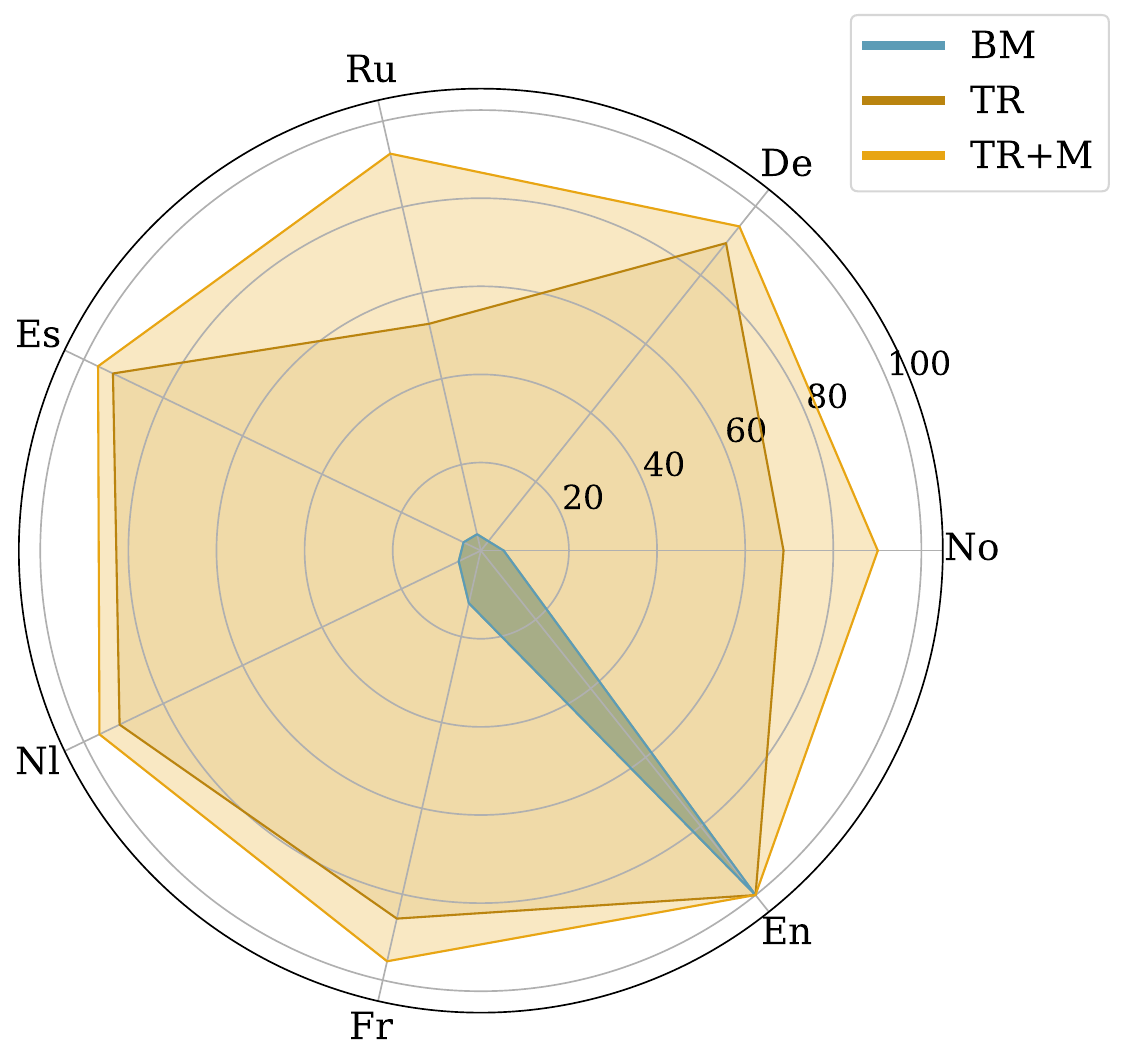}
    \caption{Language Fidelity (LF) accuracy on Crossmodal-3600. (BM: Base Model, TR: model trained with multilingual Textual Regularization, TR+M: TR and merging the final model with the original LLM Backbone)}
    \label{fig:pplot}
\end{figure}

Large Language Models (LLMs) have significantly advanced multimodal understanding, leading to the rise of VLMs, which integrate vision encoders into LLM backbones. A widely adopted paradigm is the LLaVA-style architecture~\citep{llava,llava_1_5}, where a decoder-only LLM is coupled with a vision encoder and an adapter module to align visual representations with textual embeddings.

Despite their success, VLMs exhibit a strong bias toward English due to the predominance of monolingual vision-language training data. Consequently, they often generate English responses regardless of the input language, a phenomenon termed Image-induced Fidelity Loss (or IFL)~\citep{hinck-etal-2024-llava}. This issue stems from limitations in the underlying LLM rather than the visual representations.

Ensuring multilingual capability in VLMs is essential for their adoption across diverse linguistic communities, as reliance on English-centric outputs risk erasing cultural and linguistic nuances. Prior work~\citep{qiu-etal-2022-multilingual,li2023m} has explored dataset translation, but this approach incurs high computational costs and introduces translation errors, especially in images with language-dependent elements.

In this paper, we propose an alternative solution by integrating multilingual text-only data during the visual instruction tuning process. Additionally, we explore model merging, combining the visually fine-tuned model with the original multilingual backbone LLM to further preserve linguistic fidelity. As shown in Figure~\ref{fig:pplot}, our method effectively prevents the model from defaulting to English in non-English queries. To the best of our knowledge, no previous work has achieved full multilingual competence in VLMs through such a simple and scalable approach.

Our contributions are as follows:

\begin{itemize}
    \item We systematically demonstrate that integrating multilingual text-only data during training significantly reduces IFL bias in LLaVA-style VLMs while maintaining core capabilities.
    \item We conduct an extensive analysis on the optimal proportion of text-only data required for effective multilingual adaptation.
    \item We explore a model merging strategy, combining the visually fine-tuned model with the original multilingual backbone LLM, and assess its impact on preserving linguistic fidelity.
\end{itemize}

Our findings suggest that we can develop high-quality multilingual VLMs that maintain strong performance across multiple languages in a simple and scalable way. By avoiding the need to translate or construct multimodal datasets for each language, our approach lowers the entry barrier for multilingual VLM development.  This makes it especially attractive for low-resource settings, where monolingual text is often available but collecting vision-language data is costly or impractical.

\section{Related Work}

\subsection{Multimodal Large Language Models}

VLMs typically integrate an image encoder, usually CLIP~\citep{clip,vit}, with an LLM backbone. Various strategies exist for combining these components. The predominant approach follows a decoder-only architecture, as seen in the LLaVA series, where an adapter module projects visual representations into the textual embedding space. Other methods include cross-attention mechanisms~\citep{llama3}, and some models, like NVLM~\cite{nvlm}, adopt a hybrid strategy combining both approaches.

LLaVA-style models tend to default to English due to the scarcity of multimodal training data in other languages~\citep{hinck-etal-2024-llava}. This issue arises because the LLM's parameters are updated for a distinct task, which can disrupt its original language capabilities. Llama 3~\citep{llama3} takes a different approach by freezing the LLM during training, which helps preserve its pretrained abilities while incorporating visual information. However, freezing the LLM also limits the model’s capacity to learn new visual tasks, creating a trade-off between language preservation and multimodal learning.

\subsection{Multilingual Multimodal Learning}

A widely adopted approach to improving multilinguality in VLMs is translating existing multimodal datasets. Several works~\citep{song2024missing,hu2024large} have analyzed this strategy and proposed methods to enhance its effectiveness. Several models, such as PALI~\cite{pali}, PALI-X~\citep{pali-x}, mBLIP~\citep{geigle-etal-2024-mblip}, PALO~\citep{palo} and  Pangea~\citep{pangea}, have pursued this approach. However, this strategy presents challenges, including computational overhead, translation inconsistencies, and the loss of cultural context in visual-text pairs.

Moreover, recent research~\citep{aggarwal2024towards} suggests that continual fine-tuning can harm an LLM’s performance. When a model undergoes two consecutive fine-tuning phases with differing task distributions, its ability to perform earlier tasks deteriorates. This raises concerns that direct fine-tuning solely on translated multimodal data may degrade the LLM’s original capabilities.

\subsection{Catastrophic Forgetting Prevention}

In the context of LLMs, the problem of maintaining performance across tasks while integrating new information is known as lifelong learning. This field focuses on a system’s ability to acquire, integrate, and retain knowledge without catastrophically forgetting previous information. Visual Instruction Tuning is a case of lifelong learning, and it faces the same challenges. One known mitigation strategy is episodic or experience replay~\citep{zheng2025survey}, which helps prevent catastrophic forgetting by reintroducing previously learned information.

Several studies~\citep{koyejo2022,ibrahim2024simple} have explored ways to incorporate pretraining data during fine-tuning. \citet{bethune2025scaling} further analyze the impact of this approach and suggest that even a small amount of pretraining data can help retain previously learned knowledge, reducing the risk of performance degradation.

In the case of VLMs, NVLM~\citep{nvlm} and InternVL 2.5~\citep{internvl25} demonstrate that incorporating high-quality text-only data during Visual Instruction Tuning, not only improves the overall text-generation capabilities, but also multimodal performance. Our approach builds upon these findings by integrating multilingual text-only data throughout VLM training to mitigate IFL, without requiring extensive multimodal multilingual data collection.

\subsection{Model Merging}

Model merging is a technique that involves combining two or more pre-trained models to create a new model that leverages the strengths of each. By merging a fine-tuned model with its original backbone, this process preserves the model's prior capabilities while incorporating additional refinements from further training. This strategy has been applied in various contexts, such as language transfer, where \citet{model_merging_cf} demonstrate that model merging facilitates fine-tuning for new linguistic capabilities without compromising the performance of the original LLM.

Building on this insight, we explore model merging as a means of preserving the multilingual competencies of a VLM during the visual fine-tuning process. We adopt the same model merging strategy as Aya Vision~\citep{AyaVision}, which has shown strong empirical results, and combine it with our multilingual textual regularization strategy.

\section{Experimental Setup}

\subsection{Data}
\label{sec:data}
Our training framework combines multimodal visual-language data from LLaVA-OneVision~\citep{llavaonevision} with multilingual text-only instruction data from the Salamandra family of models~\citep{gonzalez-agirre_salamandra_2025}. This hybrid approach ensures robust visual understanding while addressing IFL through explicit multilingual text supervision. All datasets are documented in Appendix \ref{app:data-sources}.


\paragraph{Visual Data}

We employ LLaVA-OneVision's  English-only visual pipeline, which is divided into two main groups:


\begin{itemize}
    \item \textbf{General and Detailed Image Captions}: This dataset comprises both basic and highly detailed image captions. The basic captions align the visual embedding space with the LLM's embedding space, while the detailed captions refine the mapping between the two providing a high-quality understanding of the images. This group comprises \(4.4M\) unique instances.

    \item \textbf{Task-Specific, Multi-Image, and Video Data}: This dataset is used to instruct the aligned model on specific tasks, including Optical Character Recognition (OCR), infographic understanding, and math \& reasoning. Additionally, multi-image and video data are incorporated to enhance the model's ability to interpret diverse visual inputs. This group comprises \(4.1M\) unique instances.
\end{itemize}

\paragraph{Multilingual Text-Only Data}

To further enhance the model’s multilingual proficiency, we incorporate 315,496 text-only samples drawn from 11 diverse datasets covering domains such as general language tasks, multilingual instructions, conversational QA, and code annotations. These sources include high-quality, human-annotated datasets, like the No Robots~\citep{no_robots} dataset, alongside multilingual instruction collections like the Aya Dataset~\citep{singh-etal-2024-aya} and FLORES-200 Instructions~\citep{costa2022no}, as well as conversational data from resources such as Databricks Dolly~\citep{DatabricksBlog2023DollyV2} and OASST~\citep{kopf2023openassistant}.

Notably, the text-only samples cover 21 of the 35 languages used in training Salamandra, ensuring extensive linguistic representation. A significant portion of this dataset is machine-translation data.

\begin{figure}
    \centering
    \includegraphics[width=0.8\linewidth]{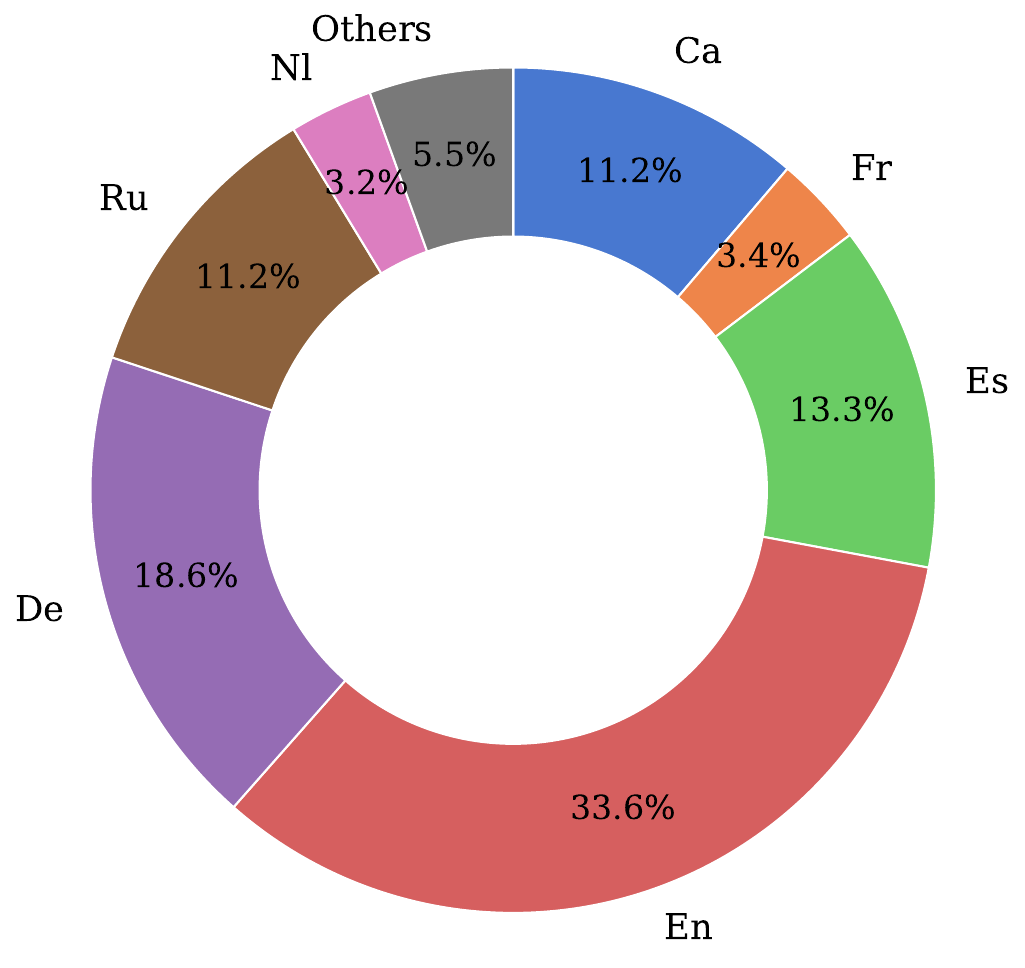}
    \caption{Distribution of the multilingual text-only data used for Textual Regularization. Languages with a volume smaller than 3\% are grouped under \textit{Others}, which collectively account for 5.5\% of the data. The most frequent languages in this group are Portuguese (2.1\%), Italian (0.7\%), Polish (0.47\%), Swedish (0.42\%), Irish (0.39\%), Lithuanian (0.29\%), Galician (0.22\%), Greek (0.20\%), and Ukrainian (0.17\%).}

    \label{fig:text-only_language_dist}
\end{figure}

Although most of the text-only data is in English, matching the language of the visual data, its inclusion remains important. This alignment reinforces the model’s linguistic foundation and facilitates the integration of multilingual supervision, ultimately ensuring balanced performance across modalities. The final distribution of languages in the text-only data, complementing LLaVA-OneVision’s training data, is shown in Figure \ref{fig:text-only_language_dist}.

\paragraph{Evaluation Data}

For evaluation, we use both monolingual and multilingual multimodal datasets. To assess visual performance, we include AI2D~\citep{ai2d}, which tests understanding of diagram-based questions; RealWorldQA\footnote{\url{https://huggingface.co/datasets/visheratin/realworldqa}}, a real-world image dataset with open-ended and multiple-choice questions; MMMU~\citep{mmmu}, a diverse multimodal reasoning benchmark; and MMStar~\citep{mmstar}, which aggregates vision-language tasks for broad multimodal evaluation.

The first two benchmarks primarily assess the exact match accuracy, quantifying the proportion of responses that exactly match the predefined ground truth. These targets are typically short-form text or multiple-choice answers. On the other hand, MMMU and MMStar are classification tasks that are measured using accuracy. 

For multilingual multimodal performance, we have selected Crossmodal-3600~\citep{xm3600}, a geographically diverse multilingual multimodal dataset for image captioning. It is particularly well-suited for our experiments as it covers the highest number of overlapping languages with Salamandra while allowing for image captioning with multiple reference targets per instance. The dataset comprises approximately 3,600 samples in 36 languages, from which we evaluate on  German (De), Russian (Ru), Spanish (Es), Dutch (Du), French (Fr), and English (En). Appendix \ref{app:cap_gen} discusses the prompts used for caption generation during evaluation and the rationale behind their selection.

\subsection{Model Framework}

Our implementation also follows the LLaVA-OneVision framework~\footnote{\url{https://github.com/LLaVA-VL/LLaVA-NeXT}}, adapted for enhanced multilingual capabilities. The architecture consists of three main components: \texttt{SigLIP-SO400M}~\footnote{\url{https://huggingface.co/google/siglip-so400m-patch14-384}}~\citep{siglip} as the visual encoder, a 2-layer MLP projection module (with GELU activation functions~\citep{gelu}), and \texttt{Salamandra-7b-instruct~\footnote{\url{https://huggingface.co/BSC-LT/salamandra-7b-instruct}}~\citep{gonzalez-agirre_salamandra_2025}} as the backbone LLM. Salamandra has been chosen for its high multilinguality, as it has been trained with 35 European languages.


We also adopt LLaVA-OneVision's curriculum learning training strategy, which progresses through four distinct stages:

\begin{itemize}
    \item \textbf{Stage 1 (Language-Image Alignment)}: In this initial phase, only the MLP projector is trained, while both the visual encoder and LLM remain frozen. General image captions are employed to establish basic cross-modal connections.
    \item \textbf{Stage 1.5 (Full Model Training)}: At this stage, all model components are unfrozen to enable end-to-end training. A high-quality set of detailed image captions is used in conjunction with an increased image resolution to enhance visual detail processing.
    \item \textbf{Stage 2 (Single-Image Instruction Tuning)}: Once the model has achieved a deep understanding of images, it is fine-tuned for a diverse set of visual tasks. The image resolution is further increased to support fine-grained visual analysis.
    \item \textbf{Stage 2.5 (Multi-Image and Video Training)}: In the final stage, multi-image and video data are incorporated to enable reasoning across multiple visual inputs. Additionally, single-image data from the previous stage is also utilized.
\end{itemize}

The key innovation in our approach lies in the strategic injection of multilingual text-only data throughout these training stages detailed in \S\ref{sec:experiments}. After the Visual Instruction Tuning, the model is merged with the baseline LLM weights using linear interpolation.

\subsection{Metrics}
\label{sec:metrics}

To evaluate language fidelity and consistency, we employ a common metric established in prior multilingual multimodal evaluation work~\citep{hinck-etal-2024-llava,m5benchmark}:
\paragraph{Language Fidelity}
We use GlotLID~\citep{glotlid} to obtain the accuracy of whether the language of the generated captions over Crossmodal-3600 images is the same as the user prompt~\footnote{Crossmodal-3600 does not include a predefined reference generation prompt. For completeness, we present and explain the employed prompt in Appendix~\ref{app:cap_gen}.}. We named this metric LF, and we observed that, in many cases, it considered as correct samples that had single words in English, or with minor code-switching errors. To address this issue, we extend this metric (LF+) by using \texttt{Llama-3.1-8B-Instruct}\footnote{\url{https://huggingface.co/meta-llama/Llama-3.1-8B-Instruct}} ~\citep{llama3} as an LLM-as-a-judge, evaluating if the samples already classified by GlotLID are entirely in the same language or not. Nevertheless, due to a majority voting strategy in its implementation (see Appendix \ref{app:text_LLM_as_a_judge}), the LLM-as-a-judge does not work perfectly, as it sometimes misclassifies correct samples as non-consistent in language. For this reason, this metric can be interpreted as a statistical lower bound of language fidelity. 

\paragraph{Visual Performance}
To evaluate visual performance, we use the English-only multimodal benchmarks detailed in Section \ref{sec:data}: AI2D, RealWorldQA, MMMU, and MMStar. AI2D and RealWorldQA are evaluated using exact match accuracy, measuring the proportion of responses identical to ground-truth answers (typically short text or multiple-choice). MMMU and MMStar, however, are treated as classification tasks and evaluated via accuracy.

To evaluate multilingual multimodal performance, we also use the same approach used in \citet{hinck-etal-2024-llava,m5benchmark}, and evaluate the captioning quality across different languages with chrF++~\citep{chrf,chrf++} over Crossmodal-3600 samples.

Further discussion on metric selection can be found in Appendix~\ref{app:metrics_discussion}.

\subsection{Experiments}
\label{sec:experiments}

We focus on testing different text-only integration strategies, analyzing the influence of data quantity, examining generalization capabilities on languages not contained during textual regularization, and assessing the effect of model merging on overall performance. To be able to quantify the results obtained with these experiments, we also trained a baseline model (BM) by only conducting the Visual Instruction Tuning, without textual regularization.

\paragraph{Multilingual Data Integration Strategies}
We explore three distinct strategies for incorporating the text-only  multilingual data (315,496 instances) during the visual instruction tuning process:
\begin{itemize}
    \item \textbf{Textual Regularization across Three Stages (TR-3S)}: Multilingual text data was distributed proportionally across the final three training stages (1.5, 2, and 2.5).
    \item \textbf{Textual Regularization across Two Stages (TR-2S)}: Multilingual text data was integrated proportionally only in the last two stages (2 and 2.5).
    \item \textbf{Textual Regularization at a Single Stage (TR-1S)}: Multilingual text data was added exclusively during the final stage (2.5).
\end{itemize}

\paragraph{Multilingual Generalization Capabilities}
To investigate whether regularization with multilingual text data extends to languages not explicitly seen during training, we train a variant of the TR-3S model where German was excluded from the multilingual text dataset. This experiment allows us to evaluate the model's generalization ability to new languages.

\paragraph{Influence of Data Balance}
We conduct experiments by varying the ratio of multilingual text data to visual data used for regularization. Specifically, we test ratios of 0.0125x, 0.025x, 0.05x (the original configuration), and 0.1x, where `x' represents the total volume of visual data (5.5M samples). This allows us to analyze how the quantity of multilingual text data affects the model's performance in terms of language fidelity.

\paragraph{Model Merging}
To explore the potential of further enhancing the multilingual capabilities of our best-performing model (TR-3S), we apply model merging. To do so, we perform a linear interpolation between the weights of the visually instructed model with those of the backbone LLM, maintaining the encoder and MLP layers. This allows us to evaluate whether model merging could combine the model's visual understanding capabilities with the language fidelity of the original model.\\

As explained in \S\ref{sec:metrics}, we evaluate multimodal performance on a suite of English benchmarks and extend it to multiple languages evaluating chrF++ on Crossmodal-3600. Moreover, we use LF to assess IFL and further analyze its bounds with LF+.

\subsection{Implementation Details}

Our experiments were conducted on custom NVIDIA H100 GPUs, each with 64GB of memory. We trained each model for 6 days in a distributed setup with 8 nodes, each containing 4 GPUs, totaling 32 GPUs per experiment. As we trained 8 models (excluding the merged model, which did not require separate training), the total compute usage amounted to 36,864 GPU hours.

For evaluation, we assessed 9 models across 6 languages, with each requiring one node for 24 hours, resulting in 5,184 GPU hours.

The training hyperparameters were largely based on those used in LLaVA-OneVision and Salamandra’s Instruction Tuning, ensuring consistency with prior work. Further details on the training process, including specific hyperparameters and configurations, can be found in Appendix~\ref{app:training-params}.

\section{Results}
This section presents the outcomes of our experimental investigation into the effectiveness of incorporating multilingual text-only data during the visual instruction tuning process for reducing IFL in VLMs.

\subsection{Quantifying the Baseline English Bias}

To better understand the starting point of our investigation, we first evaluated the baseline model (BM), trained exclusively on English visual instruction data. As anticipated, this model exhibits a pronounced English-centric behavior, responding predominantly in English even when prompted in other languages. This confirms the strong presence of IFL and underscores the necessity of multilingual regularization.

As shown in Table~\ref{tab:3s-2s-1s}, the model demonstrates a very limited capacity to generate non-English responses. Languages such as German and Spanish, for example, show particularly low consistency, often defaulting back to English. This behavior reveals how the training process strongly anchors the model to English due to the lack of multilingual signals.


Interpreting these results in context, the baseline model’s bias highlights a fundamental limitation of current VLM training pipelines, where even models based on multilingual backbones revert to English if not explicitly trained with multilingual supervision.



\subsection{Impact of Multilingual Text-Only Data Integration Strategies}

The results presented in Table \ref{tab:3s-2s-1s} clearly demonstrate that integrating multilingual text-only data substantially mitigates English bias across all evaluated strategies. Notably, proportional integration across the final three training stages (TR-3S) consistently achieves superior LF scores for most non-English languages. This suggests that continuous exposure to multilingual text throughout training stages is most effective in maintaining linguistic fidelity. The strategy of introducing multilingual data exclusively in the final stage (TR-1S) yields the least improvement, indicating that delaying multilingual exposure is insufficient to counteract the English bias ingrained during earlier training phases. The stronger performance of TR-3S can be attributed to its role as a continual regularizer. By more extensively interleaving multilingual text-only data, the model consistently reinforces its multilingual representations, thus more effectively preserving previously acquired language capabilities and reducing IFL.


\begin{table}[ht]
    \centering
    {\begin{tabular}{lrrrr}
        \toprule
        Lang. & BM & TR-3S & TR-2S & TR-1S \\
        \midrule
        De & 2.7 & \textbf{88.7} & 81.3 & 24.5 \\
        Es & 4.4 & \textbf{92.9} & 65.4 & 38.4 \\
        Fr & 12.2 & \textbf{85.7}& 74.9 & 29.9 \\
        Nl & 5.6 & \textbf{91.8} & 91.3 & 49.2 \\
        Ru & 3.8 & \textbf{52.9} & 24.8 & 50.9 \\
        En & \textbf{100.0} & \textbf{100.0} & \textbf{100.0} & \textbf{100.0} \\
        \bottomrule
    \end{tabular}}
    \caption{LF accuracy for different integration strategies. The best results are shown in bold.}
    \label{tab:3s-2s-1s}
\end{table}

\subsection{Evaluating Multilingual Generalization Capabilities}
We obtained a LF scores of 5.4\% for German in this scenario. While these results are slightly above the English-biased baseline (2.7\% LF), the performance remains very limited. This suggests that the multilingual regularization approach, in the absence of explicit exposure to the target language, does not meaningfully help mitigate IFL. In other words, the model struggles to generalize to unseen languages, and explicit inclusion during training appears necessary for achieving satisfactory multilingual fidelity.

\subsection{Analyzing the Influence of Data Quantity}

The LF score for these variations are presented on Table \ref{tab:data-quantity}.

\begin{table}[h!]
    \centering
    \footnotesize
    \resizebox{\columnwidth}{!}{\begin{tabular}{lrrrr}
        \toprule
        Lang. & 0.0125x (\%) & 0.025x (\%) & 0.05x (\%) & 0.1x (\%) \\
        \midrule
        De & 85.0 & \textbf{88.9} & 88.7 & 73.3 \\
        Es & 91.9 & 92.4 & \textbf{92.9} & 76.4 \\
        Fr & \textbf{88.6} & 83.5 & 85.7 & 69.5 \\
        Nl & 69.8 & \textbf{96.0} & 91.8 & 93.5 \\
        Ru & 91.5 & 81.9 & 52.9 & \textbf{93.3} \\
        \bottomrule
    \end{tabular}}
    \caption{LF accuracy under varying ratios of text-only to visual data. The best results are shown in bold.}
    \label{tab:data-quantity}
\end{table}
The results demonstrate a complex relationship between the text-only to visual data ratio and language fidelity, making straightforward interpretation challenging. Increasing the ratio from 0.0125x to 0.025x generally improves LF scores for most languages, suggesting a positive impact of increased text-only data within this range. However, further increasing the ratio to 0.1x does not consistently yield better results and, in some cases, significantly reduces performance, particularly for German, Spanish, and French.

Notably, none of the tested variations drastically degrade LF across all languages compared to the original 0.05x ratio. This indicates that while the optimal text-only data ratio requires careful consideration, moderate variations around the original amount do not necessarily lead to a substantial loss in language fidelity.

\subsection{Evaluating the Effect of Model Merging}

    

\begin{table}[h!]
    \centering
    
    \footnotesize
    \resizebox{0.75\columnwidth}{!}{\begin{tabular}{lcccc}
        \toprule
        & TR-3S (\%) & TR-3S + M (\%) \\
        \midrule
        De & 88.7 & \textbf{94.1} \\
        Es & 92.9 & \textbf{96.4} \\
        Fr & 85.7 & \textbf{95.5} \\
        Nl & 91.8 & \textbf{96.1} \\
        Ru & 52.9 & \textbf{92.4} \\
        \bottomrule
    \end{tabular}}
    \caption{LF score on multiple languages before (TR-3S) and after (TR-3S + M) model merging. The best results are shown in bold.}
    \label{tab:merging}
\end{table}

The results presented in Table~\ref{tab:merging} demonstrate a clear positive impact of model merging on language fidelity (LF). On average, LF improved by over 12.5 points across the five languages (5.75 points removing Russian). This improvement is consistent across all evaluated languages, indicating a robust effect of the merging strategy in enhancing multilingual fidelity, rather than a language-specific anomaly. Importantly, this analysis focuses solely on language fidelity. The impact of model merging on other VLM capabilities will be discussed in \S\ref{sec:task_performance}. These findings support the hypothesis that model merging can be a valuable tool in mitigating IFL.

\subsection{Bounding IFL}
\begin{figure}
    \centering
    \includegraphics[width=0.95\linewidth]{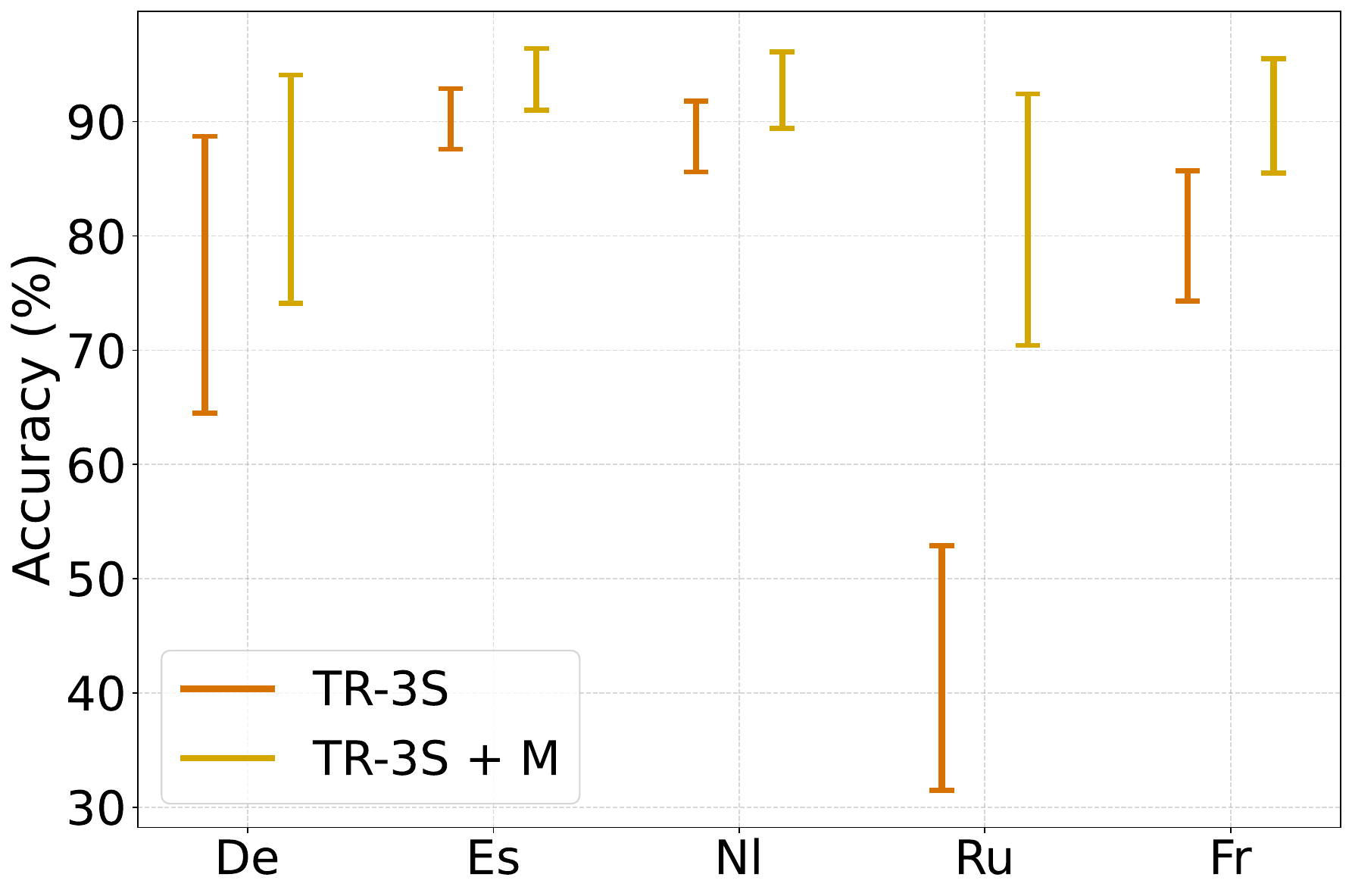}
    \caption{Interval Plot contrasting LF (upper bars) vs. LF+ (lower bars) across languages of our best-performing models.}
    \label{fig:interval_plot}
\end{figure}

Figure~\ref{fig:interval_plot} represents the upper (LF) and lower bounds (LF+) of IFL of our best-performing models. As it can be seen, in most languages we can observe a span of approximately \(10\%\) or less, except for De and Ru, which is around \(20\%\). We attribute this difference to the lower performance of the LLM-as-a-judge on these languages (detailed in Appendix~\ref{app:text_LLM_as_a_judge}). Nonetheless, it is important to note that this is not a statistical representation, so the interval width is not significant in terms of performance.

The results confirm the validity of our method, as all the lower bounds (excluding TR-3S on Russian) surpass the 65\% accuracy.

\subsection{Impact on General Tasks Performance}
\label{sec:task_performance}

\begin{table*}[t!]
    \centering
    \begin{tabular}{lcccc}
        \toprule
        Model & AI2D$_{EM}$ & RealWorldQA$_{EM}$ & MMMU$_{Acc}$ (val) & MMStar$_{Acc}$ (avg) \\
        \midrule
        BM & 73.96 & \textbf{56.99} & \textbf{34.22} & 47.33 \\
        TR-3S & \textbf{75.39} & 54.25 & 33.56 & \textbf{48.87} \\
        TR-3S + M & 57.19 & 52.03 & 34.11 & 42.25 \\
        \bottomrule
    \end{tabular}
    \caption{Performance on general VLM benchmarks (only in English). All scores are reported on a 0–100 scale. The best results are shown in bold.}
    \label{tab:performance}
\end{table*}

Our analysis shown in Table \ref{tab:performance} reveals that the models trained with our proposed regularization techniques, specifically those utilizing proportional multilingual text-only data integration, generally maintain or slightly improve performance on standard VLM benchmarks compared to the baseline English-centric model (BM). For instance, the TR-3S model, which incorporates text-only data across three training stages, exhibits an increase in AI2D and MMStar scores. This demonstrates that our method effectively mitigates IFL without sacrificing the model's core visual-language understanding capabilities. The strategic injection of multilingual text-only data appears to reinforce the LLM's inherent multilingual abilities without disrupting its ability to process and understand visual information.

The evaluation using the chrF++ metric, which measures the quality of text generation by comparing character n-grams, further supports the effectiveness of our multilingual regularization techniques. As shown in Table \ref{tab:chrf++}, the TR-3S model demonstrates improved chrF++ scores across all non-English languages compared to the baseline (BM). For instance, German improves from 15.0 to 20.4, and Spanish from 19.1 to 23.7. This indicates that the model not only maintains language fidelity but also generates more accurate and coherent text in multilingual settings.

\begin{table}[h!]
    \centering
    \footnotesize
    \resizebox{\columnwidth}{!}{\begin{tabular}{lcccccc}
        \toprule
        Model & De & Ru & Es & Nl & Fr & En \\
        \midrule
        BM & 15.0 & 9.9 & 19.1 & 16.2 & 18.1 & 27.5 \\
        TR-3S & \textbf{20.4} & \textbf{12.5} & \textbf{23.7} & \textbf{22.0} & \textbf{22.8} & \textbf{28.2} \\
        TR-3S + M & 16.1 & 10.5 & 21.6 & 14.7 & 18.5 & 25.5 \\
        \bottomrule
    \end{tabular}}
    \caption{Performance on Crossmodal-3600 by language (chrF++). The best results are shown in bold.}
    \label{tab:chrf++}
\end{table}

However, a notable observation is the performance degradation observed in the merged model. Despite achieving substantial improvements in multilingual fidelity, the TR-3S M model shows a significant decrease in performance on benchmarks such as AI2D. This decline suggests a potential trade-off between enhanced multilingual capabilities and general task performance when employing model merging techniques. We hypothesize that the merging process, while beneficial for consolidating multilingual knowledge, may introduce conflicts or misalignments in the model's learned visual representations. We further investigated alternative merging methods, including spherical linear interpolation (slerp) and both asymmetric weightings that favor the original backbone (75--25) and the visually instructed model (25--75). These variants, detailed in Appendix~\ref{app:merging-tradeoffs}, confirm the trade-offs between language fidelity and multimodal performance, without revealing a universally superior configuration.

Examples of the generation with the TR-3S model can be found in Appendix \ref{app:generation_examples}.

\section{Conclusion}

We addressed the challenge of Image-induced Fidelity Loss in VLMs, where models trained on predominantly English data tend to default to English responses. Our approach integrates multilingual text-only data into the visual instruction tuning process, preserving the multilingual abilities of the underlying language model.

Experiments show that proportional multilingual integration (TR-3S) significantly reduces English bias while maintaining core multimodal capabilities. We also analyzed data quantity effects, finding that moderate variations in text-to-visual data ratios do not compromise fidelity, though explicit inclusion of target languages remains necessary. Additionally, we introduced a model merging strategy that further improves language fidelity, albeit with some trade-offs in general task performance, highlighting the need for balance in practical applications.

Overall, our findings demonstrate that multilingual textual regularization is a simple and scalable solution to enhance VLM multilingual competence without large multimodal multilingual datasets. This paves the way for future research on optimizing data integration and refining model merging techniques to balance fidelity and overall performance.

\section*{Limitations}

\subsection*{Language Coverage}
While our approach improves multilingual alignment through text-only supervision, the language coverage remains predominantly European. This raises concerns about the model's ability to generalize to typologically diverse languages, particularly those with non-Latin scripts (e.g., Arabic, Hindi, Chinese). Future work should explore the integration of a wider array of language families and scripts to validate and expand the method's applicability.

\subsection*{Metric Reliability}
The fidelity metric (GlotLID and LF+) relies on automatic tools and heuristic judgments, including LLM-as-a-judge assessments that exhibit sensitivity to code-switching and short prompts. Despite efforts to address false positives, such metrics are not infallible and may fail to fully capture semantic fidelity across languages.

\section*{Ethical Considerations}

Our research tackles key ethical issues related to multilingual representation and inclusivity in visual language models. Enhancing multilingual capabilities promotes accessibility and fairness across diverse linguistic communities.

However, relying on machine-translated datasets may introduce biases or cultural inaccuracies. Ensuring responsible translation and ongoing refinement is crucial.

Real-world deployment also demands cultural sensitivity, especially in sectors like education, health, or governance. We emphasize the need for transparency, continuous monitoring, and collaboration with diverse communities to ensure responsible development and use.

\section*{Acknowledgements}

This work has been supported and funded by the \textit{Ministerio para la Transformación Digital y de la Función Pública} and the \textit{Plan de Recuperación, Transformación y Resiliencia} – funded by the EU through \textit{NextGenerationEU}, within the framework of the \textit{Modelos del Lenguaje} project.

\bibliography{custom}

\begin{thebibliography}{134}
\providecommand{\natexlab}[1]{#1}

\bibitem[{Acharya et~al.(2019)Acharya, Kafle, and Kanan}]{acharya2019tallyqa}
Manoj Acharya, Kushal Kafle, and Christopher Kanan. 2019.
\newblock \href {https://doi.org/10.1609/aaai.v33i01.33018076} {Tallyqa:
  Answering complex counting questions}.
\newblock \emph{Proceedings of the AAAI Conference on Artificial Intelligence},
  33(01):8076--8084.

\bibitem[{Aggarwal et~al.(2024)Aggarwal, Damle, Goyal, Lokam, and
  Sitaram}]{aggarwal2024towards}
Divyanshu Aggarwal, Sankarshan Damle, Navin Goyal, Satya Lokam, and Sunayana
  Sitaram. 2024.
\newblock \href {https://openreview.net/forum?id=1rfmxGtSjq} {Towards exploring
  continual fine-tuning for enhancing language ability in large language
  model}.
\newblock In \emph{NeurIPS 2024 Workshop on Fine-Tuning in Modern Machine
  Learning: Principles and Scalability}.

\bibitem[{Alexandrov et~al.(2024)Alexandrov, Raychev, Mueller, Zhang, Vechev,
  and Toutanova}]{model_merging_cf}
Anton Alexandrov, Veselin Raychev, Mark~Niklas Mueller, Ce~Zhang, Martin
  Vechev, and Kristina Toutanova. 2024.
\newblock \href {https://doi.org/10.18653/v1/2024.findings-emnlp.1000}
  {Mitigating catastrophic forgetting in language transfer via model merging}.
\newblock In \emph{Findings of the Association for Computational Linguistics:
  EMNLP 2024}, pages 17167--17186, Miami, Florida, USA. Association for
  Computational Linguistics.

\bibitem[{Alves et~al.(2024)Alves, Pombal, Guerreiro, Martins, Alves, Farajian,
  Peters, Rei, Fernandes, Agrawal, Colombo, de~Souza, and
  Martins}]{tower_llm_2024}
Duarte~Miguel Alves, Jos{\'e} Pombal, Nuno~M Guerreiro, Pedro~Henrique Martins,
  Jo{\~a}o Alves, Amin Farajian, Ben Peters, Ricardo Rei, Patrick Fernandes,
  Sweta Agrawal, Pierre Colombo, Jos{\'e} G.~C. de~Souza, and Andre Martins.
  2024.
\newblock \href {https://openreview.net/forum?id=EHPns3hVkj} {Tower: An open
  multilingual large language model for translation-related tasks}.
\newblock In \emph{First Conference on Language Modeling}.

\bibitem[{Amini et~al.(2019)Amini, Gabriel, Lin, Koncel-Kedziorski, Choi, and
  Hajishirzi}]{amini2019mathqa}
Aida Amini, Saadia Gabriel, Shanchuan Lin, Rik Koncel-Kedziorski, Yejin Choi,
  and Hannaneh Hajishirzi. 2019.
\newblock \href {https://doi.org/10.18653/v1/N19-1245} {{M}ath{QA}: Towards
  interpretable math word problem solving with operation-based formalisms}.
\newblock In \emph{Proceedings of the 2019 Conference of the North {A}merican
  Chapter of the Association for Computational Linguistics: Human Language
  Technologies, Volume 1 (Long and Short Papers)}, pages 2357--2367,
  Minneapolis, Minnesota. Association for Computational Linguistics.

\bibitem[{Antol et~al.(2015)Antol, Agrawal, Lu, Mitchell, Batra, Zitnick, and
  Parikh}]{antol2015vqa}
Stanislaw Antol, Aishwarya Agrawal, Jiasen Lu, Margaret Mitchell, Dhruv Batra,
  C.~Lawrence Zitnick, and Devi Parikh. 2015.
\newblock \href {https://doi.org/10.1109/ICCV.2015.279} {{VQA: Visual Question
  Answering}}.
\newblock In \emph{2015 IEEE International Conference on Computer Vision
  (ICCV)}, pages 2425--2433.

\bibitem[{Azuma et~al.(2022)Azuma, Miyanishi, Kurita, and
  Kawanabe}]{azuma2022scanqa}
Daichi Azuma, Taiki Miyanishi, Shuhei Kurita, and Motoaki Kawanabe. 2022.
\newblock \href {https://doi.org/10.1109/CVPR52688.2022.01854} {{ScanQA: 3D
  Question Answering for Spatial Scene Understanding}}.
\newblock In \emph{2022 IEEE/CVF Conference on Computer Vision and Pattern
  Recognition (CVPR)}, pages 19107--19117.

\bibitem[{Bai et~al.(2023)Bai, Mou, Likhomanenko, Cinbis, Tuzel, Huang, Shan,
  Shi, and Cao}]{bai2023vision}
Haoping Bai, Shancong Mou, Tatiana Likhomanenko, Ramazan~Gokberk Cinbis, Oncel
  Tuzel, Ping Huang, Jiulong Shan, Jianjun Shi, and Meng Cao. 2023.
\newblock \href {https://arxiv.org/abs/2306.07890} {Vision datasets: A
  benchmark for vision-based industrial inspection}.
\newblock \emph{arXiv preprint arXiv:2306.07890}.

\bibitem[{Bethune et~al.(2025)Bethune, Grangier, Busbridge, Gualdoni, Cuturi,
  and Ablin}]{bethune2025scaling}
Louis Bethune, David Grangier, Dan Busbridge, Eleonora Gualdoni, Marco Cuturi,
  and Pierre Ablin. 2025.
\newblock \href {https://arxiv.org/abs/2502.06042} {Scaling laws for forgetting
  during finetuning with pretraining data injection}.
\newblock \emph{arXiv preprint arXiv:2502.06042}.

\bibitem[{Biten et~al.(2019)Biten, Tito, Mafla, Gomez, Rusiñol, Jawahar,
  Valveny, and Karatzas}]{biten2019scene}
Ali~Furkan Biten, Rubèn Tito, Andrés Mafla, Lluis Gomez, Marçal Rusiñol,
  C.V. Jawahar, Ernest Valveny, and Dimosthenis Karatzas. 2019.
\newblock \href {https://doi.org/10.1109/ICCV.2019.00439} {Scene text visual
  question answering}.
\newblock In \emph{2019 IEEE/CVF International Conference on Computer Vision
  (ICCV)}, pages 4290--4300.

\bibitem[{Caesar et~al.(2020)Caesar, Bankiti, Lang, Vora, Liong, Xu, Krishnan,
  Pan, Baldan, and Beijbom}]{caesar2020nuscenes}
Holger Caesar, Varun Bankiti, Alex~H. Lang, Sourabh Vora, Venice~Erin Liong,
  Qiang Xu, Anush Krishnan, Yu~Pan, Giancarlo Baldan, and Oscar Beijbom. 2020.
\newblock \href {https://doi.org/10.1109/CVPR42600.2020.01164} {nuscenes: A
  multimodal dataset for autonomous driving}.
\newblock In \emph{2020 IEEE/CVF Conference on Computer Vision and Pattern
  Recognition (CVPR)}, pages 11618--11628.

\bibitem[{Carter(2024)}]{textocr-gpt4v}
Jimmy Carter. 2024.
\newblock Textocr-gpt4v.
\newblock \url{https://huggingface.co/datasets/jimmycarter/textocr-gpt4v}.

\bibitem[{Chang et~al.(2022{\natexlab{a}})Chang, Palzer, Li, Fosler-Lussier,
  and Xiao}]{chang2022mapqa}
Shuaichen Chang, David Palzer, Jialin Li, Eric Fosler-Lussier, and Ningchuan
  Xiao. 2022{\natexlab{a}}.
\newblock \href {https://arxiv.org/abs/2211.08545} {{MapQA: A Dataset for
  Question Answering on Choropleth Maps}}.
\newblock In \emph{NeurIPS 2022 First Table Representation Workshop}.

\bibitem[{Chang et~al.(2022{\natexlab{b}})Chang, Cao, Narang, Gao, Suzuki, and
  Bisk}]{chang2022webqa}
Yingshan Chang, Guihong Cao, Mridu Narang, Jianfeng Gao, Hisami Suzuki, and
  Yonatan Bisk. 2022{\natexlab{b}}.
\newblock \href {https://doi.org/10.1109/CVPR52688.2022.01600} {{WebQA:
  Multihop and Multimodal QA}}.
\newblock In \emph{2022 IEEE/CVF Conference on Computer Vision and Pattern
  Recognition (CVPR)}, pages 16474--16483.

\bibitem[{Chen et~al.(2024{\natexlab{a}})Chen, Chen, Zhang, Chen, Wu, Zhang,
  Chen, Li, Wan, and Wang}]{chen2024allava}
Guiming~Hardy Chen, Shunian Chen, Ruifei Zhang, Junying Chen, Xiangbo Wu, Zhiyi
  Zhang, Zhihong Chen, Jianquan Li, Xiang Wan, and Benyou Wang.
  2024{\natexlab{a}}.
\newblock \href {https://arxiv.org/abs/2402.11684} {{ALLaVA: Harnessing
  GPT4V-synthesized Data for A Lite Vision-Language Model}}.
\newblock \emph{Preprint}, arXiv:2402.11684.

\bibitem[{Chen et~al.(2022)Chen, Li, Qin, Lu, Lin, Chen, and
  Liang}]{chen2022unigeo}
Jiaqi Chen, Tong Li, Jinghui Qin, Pan Lu, Liang Lin, Chongyu Chen, and Xiaodan
  Liang. 2022.
\newblock \href {https://doi.org/10.18653/v1/2022.emnlp-main.218} {{U}ni{G}eo:
  Unifying geometry logical reasoning via reformulating mathematical
  expression}.
\newblock In \emph{Proceedings of the 2022 Conference on Empirical Methods in
  Natural Language Processing}, pages 3313--3323, Abu Dhabi, United Arab
  Emirates. Association for Computational Linguistics.

\bibitem[{Chen et~al.(2021)Chen, Tang, Qin, Liang, Liu, Xing, and
  Lin}]{chen2021geoqa}
Jiaqi Chen, Jianheng Tang, Jinghui Qin, Xiaodan Liang, Lingbo Liu, Eric Xing,
  and Liang Lin. 2021.
\newblock \href {https://doi.org/10.18653/v1/2021.findings-acl.46} {{G}eo{QA}:
  A geometric question answering benchmark towards multimodal numerical
  reasoning}.
\newblock In \emph{Findings of the Association for Computational Linguistics:
  ACL-IJCNLP 2021}, pages 513--523, Online. Association for Computational
  Linguistics.

\bibitem[{Chen et~al.(2025)Chen, Li, Dong, Zhang, He, Wang, Zhao, and
  Lin}]{chen2024sharegpt4v}
Lin Chen, Jinsong Li, Xiaoyi Dong, Pan Zhang, Conghui He, Jiaqi Wang, Feng
  Zhao, and Dahua Lin. 2025.
\newblock \href {https://doi.org/10.1007/978-3-031-72643-9_22} {{ShareGPT4V:
  Improving Large Multi-modal Models with Better Captions}}.
\newblock In \emph{Computer Vision -- ECCV 2024}, pages 370--387, Cham.
  Springer Nature Switzerland.

\bibitem[{Chen et~al.(2024{\natexlab{b}})Chen, Li, Dong, Zhang, Zang, Chen,
  Duan, Wang, Qiao, Lin, and Zhao}]{mmstar}
Lin Chen, Jinsong Li, Xiaoyi Dong, Pan Zhang, Yuhang Zang, Zehui Chen, Haodong
  Duan, Jiaqi Wang, Yu~Qiao, Dahua Lin, and Feng Zhao. 2024{\natexlab{b}}.
\newblock \href {https://openreview.net/forum?id=evP9mxNNxJ} {{Are We on the
  Right Way for Evaluating Large Vision-Language Models?}}
\newblock In \emph{The Thirty-eighth Annual Conference on Neural Information
  Processing Systems}.

\bibitem[{Chen et~al.(2024{\natexlab{c}})Chen, Wei, Li, Dong, Zhang, Zang,
  Chen, Duan, Lin, Tang, Yuan, Qiao, Lin, Zhao, and
  Wang}]{chen2024sharegpt4video}
Lin Chen, Xilin Wei, Jinsong Li, Xiaoyi Dong, Pan Zhang, Yuhang Zang, Zehui
  Chen, Haodong Duan, Bin Lin, Zhenyu Tang, Li~Yuan, Yu~Qiao, Dahua Lin, Feng
  Zhao, and Jiaqi Wang. 2024{\natexlab{c}}.
\newblock \href
  {https://proceedings.neurips.cc/paper_files/paper/2024/file/22a7476e4fd36818777c47e666f61a41-Paper-Datasets_and_Benchmarks_Track.pdf}
  {{ShareGPT4Video: Improving Video Understanding and Generation with Better
  Captions}}.
\newblock In \emph{Advances in Neural Information Processing Systems},
  volume~37, pages 19472--19495. Curran Associates, Inc.

\bibitem[{Chen et~al.(2024{\natexlab{d}})Chen, Djolonga, Padlewski, Mustafa,
  Changpinyo, Wu, Ruiz, Goodman, Wang, Tay, Shakeri, Dehghani, Salz, Lucic,
  Tschannen, Nagrani, Hu, Joshi, Pang, Montgomery, Pietrzyk, Ritter,
  Piergiovanni, Minderer, Pavetic, Waters, Li, Alabdulmohsin, Beyer, Amelot,
  Lee, Steiner, Li, Keysers, Arnab, Xu, Rong, Kolesnikov, Seyedhosseini,
  Angelova, Zhai, Houlsby, and Soricut}]{pali-x}
Xi~Chen, Josip Djolonga, Piotr Padlewski, Basil Mustafa, Soravit Changpinyo,
  Jialin Wu, Carlos~Riquelme Ruiz, Sebastian Goodman, Xiao Wang, Yi~Tay, Siamak
  Shakeri, Mostafa Dehghani, Daniel Salz, Mario Lucic, Michael Tschannen, Arsha
  Nagrani, Hexiang Hu, Mandar Joshi, Bo~Pang, Ceslee Montgomery, Paulina
  Pietrzyk, Marvin Ritter, AJ~Piergiovanni, Matthias Minderer, Filip Pavetic,
  Austin Waters, Gang Li, Ibrahim Alabdulmohsin, Lucas Beyer, Julien Amelot,
  Kenton Lee, Andreas~Peter Steiner, Yang Li, Daniel Keysers, Anurag Arnab,
  Yuanzhong Xu, Keran Rong, Alexander Kolesnikov, Mojtaba Seyedhosseini, Anelia
  Angelova, Xiaohua Zhai, Neil Houlsby, and Radu Soricut. 2024{\natexlab{d}}.
\newblock \href
  {https://openaccess.thecvf.com/content/CVPR2024/html/Chen_On_Scaling_Up_a_Multilingual_Vision_and_Language_Model_CVPR_2024_paper.html}
  {{On Scaling Up a Multilingual Vision and Language Model}}.
\newblock In \emph{Proceedings of the IEEE/CVF Conference on Computer Vision
  and Pattern Recognition (CVPR)}, pages 14432--14444.

\bibitem[{Chen et~al.(2023)Chen, Wang, Changpinyo, Piergiovanni, Padlewski,
  Salz, Goodman, Grycner, Mustafa, Beyer, Kolesnikov, Puigcerver, Ding, Rong,
  Akbari, Mishra, Xue, Thapliyal, Bradbury, Kuo, Seyedhosseini, Jia, Ayan,
  Ruiz, Steiner, Angelova, Zhai, Houlsby, and Soricut}]{pali}
Xi~Chen, Xiao Wang, Soravit Changpinyo, AJ~Piergiovanni, Piotr Padlewski,
  Daniel Salz, Sebastian Goodman, Adam Grycner, Basil Mustafa, Lucas Beyer,
  Alexander Kolesnikov, Joan Puigcerver, Nan Ding, Keran Rong, Hassan Akbari,
  Gaurav Mishra, Linting Xue, Ashish~V Thapliyal, James Bradbury, Weicheng Kuo,
  Mojtaba Seyedhosseini, Chao Jia, Burcu~Karagol Ayan, Carlos~Riquelme Ruiz,
  Andreas~Peter Steiner, Anelia Angelova, Xiaohua Zhai, Neil Houlsby, and Radu
  Soricut. 2023.
\newblock \href {https://openreview.net/forum?id=mWVoBz4W0u} {Pa{LI}: A
  jointly-scaled multilingual language-image model}.
\newblock In \emph{The Eleventh International Conference on Learning
  Representations}.

\bibitem[{Chen et~al.(2024{\natexlab{e}})Chen, Wang, Cao, Liu, Gao, Cui, Zhu,
  Ye, Tian, Liu et~al.}]{internvl25}
Zhe Chen, Weiyun Wang, Yue Cao, Yangzhou Liu, Zhangwei Gao, Erfei Cui, Jinguo
  Zhu, Shenglong Ye, Hao Tian, Zhaoyang Liu, et~al. 2024{\natexlab{e}}.
\newblock \href {https://arxiv.org/abs/2412.05271} {Expanding performance
  boundaries of open-source multimodal models with model, data, and test-time
  scaling}.
\newblock \emph{arXiv preprint arXiv:2412.05271}.

\bibitem[{Cheng et~al.(2022)Cheng, Dong, Wang, Jia, Guo, Gao, Han, Lou, and
  Zhang}]{cheng2022hitab}
Zhoujun Cheng, Haoyu Dong, Zhiruo Wang, Ran Jia, Jiaqi Guo, Yan Gao, Shi Han,
  Jian-Guang Lou, and Dongmei Zhang. 2022.
\newblock \href {https://doi.org/10.18653/v1/2022.acl-long.78} {{H}i{T}ab: A
  hierarchical table dataset for question answering and natural language
  generation}.
\newblock In \emph{Proceedings of the 60th Annual Meeting of the Association
  for Computational Linguistics (Volume 1: Long Papers)}, pages 1094--1110,
  Dublin, Ireland. Association for Computational Linguistics.

\bibitem[{Conover et~al.(2023)Conover, Hayes, Mathur, Xie, Wan, Shah, Ghodsi,
  Wendell, Zaharia, and Xin}]{DatabricksBlog2023DollyV2}
Mike Conover, Matt Hayes, Ankit Mathur, Jianwei Xie, Jun Wan, Sam Shah, Ali
  Ghodsi, Patrick Wendell, Matei Zaharia, and Reynold Xin. 2023.
\newblock \href
  {https://www.databricks.com/blog/2023/04/12/dolly-first-open-commercially-viable-instruction-tuned-llm}
  {Free dolly: Introducing the world's first truly open instruction-tuned llm}.

\bibitem[{Costa-jussà et~al.(2024)Costa-jussà, Cross, Çelebi, Elbayad,
  Heafield, Heffernan, Kalbassi, Lam, Licht, Maillard, Sun, Wang, Wenzek,
  Youngblood, Akula, Barrault, Gonzalez, Hansanti, Hoffman, Jarrett, Sadagopan,
  Rowe, Spruit, Tran, Andrews, Ayan, Bhosale, Edunov, Fan, Gao, Goswami,
  Guzmán, Koehn, Mourachko, Ropers, Saleem, Schwenk, Wang, and {NLLB
  Team}}]{costa2022no}
Marta~R. Costa-jussà, James Cross, Onur Çelebi, Maha Elbayad, Kenneth
  Heafield, Kevin Heffernan, Elahe Kalbassi, Janice Lam, Daniel Licht, Jean
  Maillard, Anna Sun, Skyler Wang, Guillaume Wenzek, Al~Youngblood, Bapi Akula,
  Loic Barrault, Gabriel~Mejia Gonzalez, Prangthip Hansanti, John Hoffman,
  Semarley Jarrett, Kaushik~Ram Sadagopan, Dirk Rowe, Shannon Spruit, Chau
  Tran, Pierre Andrews, Necip~Fazil Ayan, Shruti Bhosale, Sergey Edunov, Angela
  Fan, Cynthia Gao, Vedanuj Goswami, Francisco Guzmán, Philipp Koehn,
  Alexandre Mourachko, Christophe Ropers, Safiyyah Saleem, Holger Schwenk, Jeff
  Wang, and {NLLB Team}. 2024.
\newblock \href {https://doi.org/10.1038/s41586-024-07335-x} {{Scaling neural
  machine translation to 200 languages}}.
\newblock \emph{Nature}, 630(8018):841--846.

\bibitem[{Cui et~al.(2024)Cui, He, Ma, Chen, Tian, Wang, Li, Wang, Wang, Zhu,
  Lu, Lu, Wang, Wang, Qiao, and Dai}]{cui2025comprehensive}
Erfei Cui, Yinan He, Zheng Ma, Zhe Chen, Hao Tian, Weiyun Wang, Kunchang Li,
  Yi~Wang, Wenhai Wang, Xizhou Zhu, Lewei Lu, Tong Lu, Yali Wang, Limin Wang,
  Yu~Qiao, and Jifeng Dai. 2024.
\newblock \href {https://sharegpt4o.github.io/} {Sharegpt-4o: Comprehensive
  multimodal annotations with gpt-4o}.

\bibitem[{Dai et~al.(2017)Dai, Chang, Savva, Halber, Funkhouser, and
  Nießner}]{dai2017scannet}
Angela Dai, Angel~X. Chang, Manolis Savva, Maciej Halber, Thomas Funkhouser,
  and Matthias Nießner. 2017.
\newblock \href {https://doi.org/10.1109/CVPR.2017.261} {Scannet:
  Richly-annotated 3d reconstructions of indoor scenes}.
\newblock In \emph{2017 IEEE Conference on Computer Vision and Pattern
  Recognition (CVPR)}, pages 2432--2443.

\bibitem[{Dai et~al.(2024)Dai, Lee, Wang, Yang, Liu, Barker, Rintamaki,
  Shoeybi, Catanzaro, and Ping}]{nvlm}
Wenliang Dai, Nayeon Lee, Boxin Wang, Zhuolin Yang, Zihan Liu, Jon Barker,
  Tuomas Rintamaki, Mohammad Shoeybi, Bryan Catanzaro, and Wei Ping. 2024.
\newblock \href {https://arxiv.org/abs/2409.11402} {{NVLM: Open Frontier-Class
  Multimodal LLMs}}.
\newblock \emph{arXiv preprint arXiv:2409.11402}.

\bibitem[{Dash et~al.(2025)Dash, Nan, Ahmadian, and Dang}]{AyaVision}
Saurabh Dash, Yiyang Nan, Arash Ahmadian, and John Dang. 2025.
\newblock \href {https://huggingface.co/blog/aya-vision} {A deepdive into aya
  vision: Advancing the frontier of multilingual multimodality}.
\newblock Hugging Face Blog.

\bibitem[{Dosovitskiy et~al.(2021)Dosovitskiy, Beyer, Kolesnikov, Weissenborn,
  Zhai, Unterthiner, Dehghani, Minderer, Heigold, Gelly, Uszkoreit, and
  Houlsby}]{vit}
Alexey Dosovitskiy, Lucas Beyer, Alexander Kolesnikov, Dirk Weissenborn,
  Xiaohua Zhai, Thomas Unterthiner, Mostafa Dehghani, Matthias Minderer, Georg
  Heigold, Sylvain Gelly, Jakob Uszkoreit, and Neil Houlsby. 2021.
\newblock \href {https://openreview.net/forum?id=YicbFdNTTy} {An image is worth
  16x16 words: Transformers for image recognition at scale}.
\newblock In \emph{International Conference on Learning Representations}.

\bibitem[{Forbes et~al.(2019)Forbes, Kaeser-Chen, Sharma, and
  Belongie}]{forbes2019neural}
Maxwell Forbes, Christine Kaeser-Chen, Piyush Sharma, and Serge Belongie. 2019.
\newblock \href {https://doi.org/10.18653/v1/D19-1065} {Neural naturalist:
  Generating fine-grained image comparisons}.
\newblock In \emph{Proceedings of the 2019 Conference on Empirical Methods in
  Natural Language Processing and the 9th International Joint Conference on
  Natural Language Processing (EMNLP-IJCNLP)}, pages 708--717, Hong Kong,
  China. Association for Computational Linguistics.

\bibitem[{Fu et~al.(2023)Fu, Tamir, Sundaram, Chai, Zhang, Dekel, and
  Isola}]{fu2023dreamsim}
Stephanie Fu, Netanel Tamir, Shobhita Sundaram, Lucy Chai, Richard Zhang, Tali
  Dekel, and Phillip Isola. 2023.
\newblock \href {https://neurips.cc/virtual/2023/poster/72380} {Dreamsim:
  Learning new dimensions of human visual similarity using synthetic data}.
\newblock \emph{Advances in Neural Information Processing Systems},
  36:50742--50768.

\bibitem[{Gao et~al.(2023)Gao, Pi, Zhang, Ye, Zhong, Wang, Hong, Han, Xu, Li,
  and Kong}]{gao2023g}
Jiahui Gao, Renjie Pi, Jipeng Zhang, Jiacheng Ye, Wanjun Zhong, Yufei Wang,
  Lanqing Hong, Jianhua Han, Hang Xu, Zhenguo Li, and Lingpeng Kong. 2023.
\newblock \href {https://doi.org/10.48550/ARXIV.2312.11370} {G-llava: Solving
  geometric problem with multi-modal large language model}.
\newblock \emph{CoRR}, abs/2312.11370.

\bibitem[{Geigle et~al.(2024)Geigle, Jain, Timofte, and
  Glava{\v{s}}}]{geigle-etal-2024-mblip}
Gregor Geigle, Abhay Jain, Radu Timofte, and Goran Glava{\v{s}}. 2024.
\newblock \href {https://doi.org/10.18653/v1/2024.alvr-1.2} {m{BLIP}: Efficient
  bootstrapping of multilingual vision-{LLM}s}.
\newblock In \emph{Proceedings of the 3rd Workshop on Advances in Language and
  Vision Research (ALVR)}, pages 7--25, Bangkok, Thailand. Association for
  Computational Linguistics.

\bibitem[{Gonzalez-Agirre et~al.(2024)Gonzalez-Agirre, Marimon,
  Rodriguez-Penagos, Aula-Blasco, Baucells, Armentano-Oller, Palomar-Giner,
  Kulebi, and Villegas}]{gonzalez-agirre-etal-2024-building}
Aitor Gonzalez-Agirre, Montserrat Marimon, Carlos Rodriguez-Penagos, Javier
  Aula-Blasco, Irene Baucells, Carme Armentano-Oller, Jorge Palomar-Giner,
  Baybars Kulebi, and Marta Villegas. 2024.
\newblock \href {https://aclanthology.org/2024.lrec-main.231/} {Building a data
  infrastructure for a mid-resource language: The case of {C}atalan}.
\newblock In \emph{Proceedings of the 2024 Joint International Conference on
  Computational Linguistics, Language Resources and Evaluation (LREC-COLING
  2024)}, pages 2556--2566, Torino, Italia. ELRA and ICCL.

\bibitem[{Gonzalez-Agirre et~al.(2025)Gonzalez-Agirre, Pàmies, Llop, Baucells,
  Da~Dalt, Tamayo, Saiz, Espuña, Prats, Aula-Blasco, and
  {others}}]{gonzalez-agirre_salamandra_2025}
Aitor Gonzalez-Agirre, Marc Pàmies, Joan Llop, Irene Baucells, Severino
  Da~Dalt, Daniel Tamayo, José~Javier Saiz, Ferran Espuña, Jaume Prats,
  Javier Aula-Blasco, and {others}. 2025.
\newblock \href {https://arxiv.org/abs/2502.08489} {Salamandra technical
  report}.
\newblock \emph{{arXiv} preprint {arXiv}:2502.08489}.

\bibitem[{Grattafiori et~al.(2024)Grattafiori, Dubey, Jauhri, Pandey, Kadian,
  Al-Dahle, Letman, Mathur, Schelten, Vaughan et~al.}]{llama3}
Aaron Grattafiori, Abhimanyu Dubey, Abhinav Jauhri, Abhinav Pandey, Abhishek
  Kadian, Ahmad Al-Dahle, Aiesha Letman, Akhil Mathur, Alan Schelten, Alex
  Vaughan, et~al. 2024.
\newblock \href {https://arxiv.org/abs/2407.21783} {The llama 3 herd of
  models}.
\newblock \emph{arXiv preprint arXiv:2407.21783}.

\bibitem[{Grauman et~al.(2024)Grauman, Westbury, Byrne, Cartillier, Chavis,
  Furnari, Girdhar, Hamburger, Jiang, Kukreja, Liu, Liu, Martin, Nagarajan,
  Radosavovic, Ramakrishnan, Ryan, Sharma, Wray, Xu, Xu, Zhao, Bansal, Batra,
  Crane, Do, Doulaty, Erapalli, Feichtenhofer, Fragomeni, Fu, Gebreselasie,
  Gonzalez, Hillis, Huang, Huang, Jia, Khoo, Kolar, Kottur, Kumar, Landini, Li,
  Li, Li, Mangalam, Modhugu, Munro, Murrell, Nishiyasu, Price, Puentes,
  Ramazanova, Sari, Somasundaram, Southerland, Sugano, Tao, Vo, Wang, Wu, Yagi,
  Zhao, Zhu, Arbelaez, Crandall, Damen, Farinella, Fuegen, Ghanem, Ithapu,
  Jawahar, Joo, Kitani, Li, Newcombe, Oliva, Park, Rehg, Sato, Shi, Shou,
  Torralba, Torresani, Yan, and Malik}]{Grauman_2022_CVPR}
Kristen Grauman, Andrew Westbury, Eugene Byrne, Vincent Cartillier, Zachary
  Chavis, Antonino Furnari, Rohit Girdhar, Jackson Hamburger, Hao Jiang,
  Devansh Kukreja, Miao Liu, Xingyu Liu, Miguel Martin, Tushar Nagarajan, Ilija
  Radosavovic, Santhosh~Kumar Ramakrishnan, Fiona Ryan, Jayant Sharma, Michael
  Wray, Mengmeng Xu, Eric~Zhongcong Xu, Chen Zhao, Siddhant Bansal, Dhruv
  Batra, Sean Crane, Tien Do, Morrie Doulaty, Akshay Erapalli, Christoph
  Feichtenhofer, Adriano Fragomeni, Qichen Fu, Abrham Gebreselasie, Cristina
  Gonzalez, James Hillis, Xuhua Huang, Yifei Huang, Wenqi Jia, Weslie Khoo,
  Jachym Kolar, Satwik Kottur, Anurag Kumar, Federico Landini, Chao Li, Yanghao
  Li, Zhenqiang Li, Karttikeya Mangalam, Raghava Modhugu, Jonathan Munro,
  Tullie Murrell, Takumi Nishiyasu, Will Price, Paola~Ruiz Puentes, Merey
  Ramazanova, Leda Sari, Kiran Somasundaram, Audrey Southerland, Yusuke Sugano,
  Ruijie Tao, Minh Vo, Yuchen Wang, Xindi Wu, Takuma Yagi, Ziwei Zhao, Yunyi
  Zhu, Pablo Arbelaez, David Crandall, Dima Damen, Giovanni~Maria Farinella,
  Christian Fuegen, Bernard Ghanem, Vamsi~Krishna Ithapu, C.~V. Jawahar,
  Hanbyul Joo, Kris Kitani, Haizhou Li, Richard Newcombe, Aude Oliva, Hyun~Soo
  Park, James~M. Rehg, Yoichi Sato, Jianbo Shi, Mike~Zheng Shou, Antonio
  Torralba, Lorenzo Torresani, Mingfei Yan, and Jitendra Malik. 2024.
\newblock \href {https://doi.org/10.1109/TPAMI.2024.3381075} {{ Ego4D: Around
  the World in 3,000 Hours of Egocentric Video }}.
\newblock \emph{IEEE Transactions on Pattern Analysis \& Machine Intelligence},
  (01):1--32.

\bibitem[{Gupta et~al.(2018)Gupta, Schwenk, Farhadi, Hoiem, and
  Kembhavi}]{gupta2018imagine}
Tanmay Gupta, Dustin Schwenk, Ali Farhadi, Derek Hoiem, and Aniruddha Kembhavi.
  2018.
\newblock \href {https://arxiv.org/abs/1804.03608} {Imagine this! scripts to
  compositions to videos}.
\newblock In \emph{Proceedings of the European conference on computer vision
  (ECCV)}, pages 598--613.

\bibitem[{Gurari et~al.(2018)Gurari, Li, Stangl, Guo, Lin, Grauman, Luo, and
  Bigham}]{gurari2018vizwiz}
Danna Gurari, Qing Li, Abigale~J. Stangl, Anhong Guo, Chi Lin, Kristen Grauman,
  Jiebo Luo, and Jeffrey~P. Bigham. 2018.
\newblock \href {https://doi.org/10.1109/CVPR.2018.00380} {{ VizWiz Grand
  Challenge: Answering Visual Questions from Blind People }}.
\newblock In \emph{2018 IEEE/CVF Conference on Computer Vision and Pattern
  Recognition (CVPR)}, pages 3608--3617, Los Alamitos, CA, USA. IEEE Computer
  Society.

\bibitem[{Hendrycks and Gimpel(2016)}]{gelu}
Dan Hendrycks and Kevin Gimpel. 2016.
\newblock \href {https://arxiv.org/abs/1606.08415} {Bridging nonlinearities and
  stochastic regularizers with gaussian error linear units}.
\newblock \emph{CoRR}, abs/1606.08415.

\bibitem[{Hinck et~al.(2024)Hinck, Holtermann, Olson, Schneider, Yu,
  Bhiwandiwalla, Lauscher, Tseng, and Lal}]{hinck-etal-2024-llava}
Musashi Hinck, Carolin Holtermann, Matthew~Lyle Olson, Florian Schneider,
  Sungduk Yu, Anahita Bhiwandiwalla, Anne Lauscher, Shao-Yen Tseng, and Vasudev
  Lal. 2024.
\newblock \href {https://doi.org/10.18653/v1/2024.findings-emnlp.783} {Why do
  {LL}a{VA} vision-language models reply to images in {E}nglish?}
\newblock In \emph{Findings of the Association for Computational Linguistics:
  EMNLP 2024}, pages 13402--13421, Miami, Florida, USA. Association for
  Computational Linguistics.

\bibitem[{Hu et~al.(2024)Hu, Yao, Wang, WANG, Pan, Chen, Yu, Wu, Zhao, Zhang,
  Han, Lin, Xue, dahai li, Liu, and Sun}]{hu2024large}
Jinyi Hu, Yuan Yao, Chongyi Wang, SHAN WANG, Yinxu Pan, Qianyu Chen, Tianyu Yu,
  Hanghao Wu, Yue Zhao, Haoye Zhang, Xu~Han, Yankai Lin, Jiao Xue, dahai li,
  Zhiyuan Liu, and Maosong Sun. 2024.
\newblock \href {https://openreview.net/forum?id=Kuh5qgCGCp} {Large
  multilingual models pivot zero-shot multimodal learning across languages}.
\newblock In \emph{The Twelfth International Conference on Learning
  Representations}.

\bibitem[{Huang et~al.(2016)Huang, Ferraro, Mostafazadeh, Misra, Agrawal,
  Devlin, Girshick, He, Kohli, Batra, Zitnick, Parikh, Vanderwende, Galley, and
  Mitchell}]{huang2016visual}
Ting-Hao~Kenneth Huang, Francis Ferraro, Nasrin Mostafazadeh, Ishan Misra,
  Aishwarya Agrawal, Jacob Devlin, Ross Girshick, Xiaodong He, Pushmeet Kohli,
  Dhruv Batra, C.~Lawrence Zitnick, Devi Parikh, Lucy Vanderwende, Michel
  Galley, and Margaret Mitchell. 2016.
\newblock \href {https://doi.org/10.18653/v1/N16-1147} {Visual storytelling}.
\newblock In \emph{Proceedings of the 2016 Conference of the North {A}merican
  Chapter of the Association for Computational Linguistics: Human Language
  Technologies}, pages 1233--1239, San Diego, California. Association for
  Computational Linguistics.

\bibitem[{Hudson and Manning(2019)}]{hudson2019gqa}
Drew~A. Hudson and Christopher~D. Manning. 2019.
\newblock \href {https://doi.org/10.1109/CVPR.2019.00686} {Gqa: A new dataset
  for real-world visual reasoning and compositional question answering}.
\newblock In \emph{2019 IEEE/CVF Conference on Computer Vision and Pattern
  Recognition (CVPR)}, pages 6693--6702.

\bibitem[{Hui et~al.(2024)Hui, Yang, Zhao, Shi, Wang, Wang, Zhou, and
  Xie}]{hui2024hq}
Mude Hui, Siwei Yang, Bingchen Zhao, Yichun Shi, Heng Wang, Peng Wang, Yuyin
  Zhou, and Cihang Xie. 2024.
\newblock \href {https://arxiv.org/abs/2404.09990} {Hq-edit: A high-quality
  dataset for instruction-based image editing}.
\newblock \emph{CoRR}.

\bibitem[{Ibrahim et~al.(2024)Ibrahim, Th{\'e}rien, Gupta, Richter, Anthony,
  Belilovsky, Lesort, and Rish}]{ibrahim2024simple}
Adam Ibrahim, Benjamin Th{\'e}rien, Kshitij Gupta, Mats~Leon Richter,
  Quentin~Gregory Anthony, Eugene Belilovsky, Timoth{\'e}e Lesort, and Irina
  Rish. 2024.
\newblock \href {https://openreview.net/forum?id=DimPeeCxKO} {Simple and
  scalable strategies to continually pre-train large language models}.
\newblock \emph{Transactions on Machine Learning Research}.

\bibitem[{Isola et~al.(2015)Isola, Lim, and Adelson}]{isola2015discovering}
Phillip Isola, Joseph~J. Lim, and Edward~H. Adelson. 2015.
\newblock \href {https://doi.org/10.1109/CVPR.2015.7298744} {Discovering states
  and transformations in image collections}.
\newblock In \emph{2015 IEEE Conference on Computer Vision and Pattern
  Recognition (CVPR)}, pages 1383--1391.

\bibitem[{Iyyer et~al.(2017)Iyyer, Manjunatha, Guha, Vyas, Boyd-Graber, Daume,
  and Davis}]{iyyer2017amazing}
Mohit Iyyer, Varun Manjunatha, Anupam Guha, Yogarshi Vyas, Jordan Boyd-Graber,
  Hal Daume, and Larry Davis. 2017.
\newblock \href {https://doi.org/10.1109/CVPR.2017.686} {{ The Amazing
  Mysteries of the Gutter: Drawing Inferences Between Panels in Comic Book
  Narratives }}.
\newblock In \emph{2017 IEEE Conference on Computer Vision and Pattern
  Recognition (CVPR)}, pages 6478--6487, Los Alamitos, CA, USA. IEEE Computer
  Society.

\bibitem[{Jhamtani and Berg-Kirkpatrick(2018)}]{jhamtani2018learning}
Harsh Jhamtani and Taylor Berg-Kirkpatrick. 2018.
\newblock \href {https://doi.org/10.18653/v1/D18-1436} {Learning to describe
  differences between pairs of similar images}.
\newblock In \emph{Proceedings of the 2018 Conference on Empirical Methods in
  Natural Language Processing}, pages 4024--4034, Brussels, Belgium.
  Association for Computational Linguistics.

\bibitem[{Jiang et~al.(2024)Jiang, He, Zeng, Wei, Ku, Liu, and
  Chen}]{jiang2024mantis}
Dongfu Jiang, Xuan He, Huaye Zeng, Cong Wei, Max Ku, Qian Liu, and Wenhu Chen.
  2024.
\newblock \href {https://arxiv.org/abs/2405.01483} {Mantis: Interleaved
  multi-image instruction tuning}.
\newblock \emph{arXiv preprint arXiv:2405.01483}.

\bibitem[{Johnson et~al.(2017)Johnson, Hariharan, Van Der~Maaten, Fei-Fei,
  Lawrence~Zitnick, and Girshick}]{johnson2017clevr}
Justin Johnson, Bharath Hariharan, Laurens Van Der~Maaten, Li~Fei-Fei,
  C~Lawrence~Zitnick, and Ross Girshick. 2017.
\newblock \href {https://arxiv.org/abs/1612.06890} {Clevr: A diagnostic dataset
  for compositional language and elementary visual reasoning}.
\newblock In \emph{Proceedings of the IEEE conference on computer vision and
  pattern recognition}, pages 2901--2910.

\bibitem[{Kafle et~al.(2018)Kafle, Price, Cohen, and Kanan}]{kafle2018dvqa}
Kushal Kafle, Brian Price, Scott Cohen, and Christopher Kanan. 2018.
\newblock \href {https://doi.org/10.1109/CVPR.2018.00592} {Dvqa: Understanding
  data visualizations via question answering}.
\newblock In \emph{2018 IEEE/CVF Conference on Computer Vision and Pattern
  Recognition}, pages 5648--5656.

\bibitem[{Kahou et~al.(2017)Kahou, Michalski, Atkinson, K{\'a}d{\'a}r,
  Trischler, and Bengio}]{kahou2017figureqa}
Samira~Ebrahimi Kahou, Vincent Michalski, Adam Atkinson, {\'A}kos
  K{\'a}d{\'a}r, Adam Trischler, and Yoshua Bengio. 2017.
\newblock \href {https://arxiv.org/abs/1710.07300} {Figureqa: An annotated
  figure dataset for visual reasoning}.
\newblock \emph{arXiv preprint arXiv:1710.07300}.

\bibitem[{Kargaran et~al.(2023)Kargaran, Imani, Yvon, and Schuetze}]{glotlid}
Amir~Hossein Kargaran, Ayyoob Imani, Fran{\c{c}}ois Yvon, and Hinrich Schuetze.
  2023.
\newblock \href {https://doi.org/10.18653/v1/2023.findings-emnlp.410}
  {{G}lot{LID}: Language identification for low-resource languages}.
\newblock In \emph{Findings of the Association for Computational Linguistics:
  EMNLP 2023}, pages 6155--6218, Singapore. Association for Computational
  Linguistics.

\bibitem[{Kazemi et~al.(2024)Kazemi, Alvari, Anand, Wu, Chen, and
  Soricut}]{kazemi2024geomverse}
Mehran Kazemi, Hamidreza Alvari, Ankit Anand, Jialin Wu, Xi~Chen, and Radu
  Soricut. 2024.
\newblock \href {https://arxiv.org/abs/2312.12241} {Geomverse: A systematic
  evaluation of large models for geometric reasoning}.
\newblock In \emph{AI for Math Workshop@ ICML}.

\bibitem[{Kembhavi et~al.(2016)Kembhavi, Salvato, Kolve, Seo, Hajishirzi, and
  Farhadi}]{ai2d}
Aniruddha Kembhavi, Mike Salvato, Eric Kolve, Minjoon Seo, Hannaneh Hajishirzi,
  and Ali Farhadi. 2016.
\newblock \href {https://doi.org/10.1007/978-3-319-46493-0_15} {A diagram is
  worth a dozen images}.
\newblock In \emph{Computer Vision -- ECCV 2016}, pages 235--251, Cham.
  Springer International Publishing.

\bibitem[{Kembhavi et~al.(2017)Kembhavi, Seo, Schwenk, Choi, Farhadi, and
  Hajishirzi}]{kembhavi2017you}
Aniruddha Kembhavi, Minjoon Seo, Dustin Schwenk, Jonghyun Choi, Ali Farhadi,
  and Hannaneh Hajishirzi. 2017.
\newblock \href {https://doi.org/10.1109/CVPR.2017.571} {Are you smarter than a
  sixth grader? textbook question answering for multimodal machine
  comprehension}.
\newblock In \emph{2017 IEEE Conference on Computer Vision and Pattern
  Recognition (CVPR)}, pages 5376--5384.

\bibitem[{Kiela et~al.(2020)Kiela, Firooz, Mohan, Goswami, Singh, Ringshia, and
  Testuggine}]{kiela2020hateful}
Douwe Kiela, Hamed Firooz, Aravind Mohan, Vedanuj Goswami, Amanpreet Singh,
  Pratik Ringshia, and Davide Testuggine. 2020.
\newblock \href {https://arxiv.org/abs/2005.04790} {The hateful memes
  challenge: Detecting hate speech in multimodal memes}.
\newblock \emph{Advances in neural information processing systems},
  33:2611--2624.

\bibitem[{Kim et~al.(2022)Kim, Hong, Yim, Nam, Park, Yim, Hwang, Yun, Han, and
  Park}]{kim2022donut}
Geewook Kim, Teakgyu Hong, Moonbin Yim, JeongYeon Nam, Jinyoung Park, Jinyeong
  Yim, Wonseok Hwang, Sangdoo Yun, Dongyoon Han, and Seunghyun Park. 2022.
\newblock \href {https://arxiv.org/abs/2111.15664} {Ocr-free document
  understanding transformer}.
\newblock In \emph{European Conference on Computer Vision (ECCV)}.

\bibitem[{K{\"o}pf et~al.(2023)K{\"o}pf, Kilcher, Von~R{\"u}tte, Anagnostidis,
  Tam, Stevens, Barhoum, Nguyen, Stanley, Nagyfi
  et~al.}]{kopf2023openassistant}
Andreas K{\"o}pf, Yannic Kilcher, Dimitri Von~R{\"u}tte, Sotiris Anagnostidis,
  Zhi~Rui Tam, Keith Stevens, Abdullah Barhoum, Duc Nguyen, Oliver Stanley,
  Rich{\'a}rd Nagyfi, et~al. 2023.
\newblock \href {https://nips.cc/virtual/2023/poster/73573} {Openassistant
  conversations-democratizing large language model alignment}.
\newblock \emph{Advances in Neural Information Processing Systems},
  36:47669--47681.

\bibitem[{Krishna et~al.(2017)Krishna, Zhu, Groth, Johnson, Hata, Kravitz,
  Chen, Kalantidis, Li, Shamma, Bernstein, and Fei-Fei}]{krishna2017visual}
Ranjay Krishna, Yuke Zhu, Oliver Groth, Justin Johnson, Kenji Hata, Joshua
  Kravitz, Stephanie Chen, Yannis Kalantidis, Li-Jia Li, David~A. Shamma,
  Michael~S. Bernstein, and Li~Fei-Fei. 2017.
\newblock \href {https://doi.org/10.1007/s11263-016-0981-7} {Visual genome:
  Connecting language and vision using crowdsourced dense image annotations}.
\newblock \emph{International Journal of Computer Vision}, 123(1):32--73.

\bibitem[{Krojer et~al.(2022)Krojer, Adlakha, Vineet, Goyal, Ponti, and
  Reddy}]{krojer2022image}
Benno Krojer, Vaibhav Adlakha, Vibhav Vineet, Yash Goyal, Edoardo Ponti, and
  Siva Reddy. 2022.
\newblock \href {https://doi.org/10.18653/v1/2022.acl-long.241} {Image
  retrieval from contextual descriptions}.
\newblock In \emph{Proceedings of the 60th Annual Meeting of the Association
  for Computational Linguistics (Volume 1: Long Papers)}, pages 3426--3440,
  Dublin, Ireland. Association for Computational Linguistics.

\bibitem[{Lau et~al.(2018)Lau, Gayen, Abacha, and
  Demner-Fushman}]{lau2018dataset}
Jason~J. Lau, Soumya Gayen, Asma~Ben Abacha, and Dina Demner-Fushman. 2018.
\newblock \href {https://doi.org/10.1038/sdata.2018.251} {A dataset of
  clinically generated visual questions and answers about radiology images}.
\newblock \emph{Scientific Data}, 5(1):180251.

\bibitem[{Laurençon et~al.(2024)Laurençon, Tronchon, and
  Sanh}]{lauren2024unlocking}
Hugo Laurençon, Léo Tronchon, and Victor Sanh. 2024.
\newblock \href {https://arxiv.org/abs/2403.09029} {Unlocking the conversion of
  web screenshots into html code with the websight dataset}.
\newblock \emph{Preprint}, arXiv:2403.09029.

\bibitem[{Li et~al.(2025)Li, Zhang, Guo, Zhang, Li, Zhang, Zhang, Zhang, Li,
  Liu, and Li}]{llavaonevision}
Bo~Li, Yuanhan Zhang, Dong Guo, Renrui Zhang, Feng Li, Hao Zhang, Kaichen
  Zhang, Peiyuan Zhang, Yanwei Li, Ziwei Liu, and Chunyuan Li. 2025.
\newblock \href {https://openreview.net/forum?id=zKv8qULV6n}
  {{LL}a{VA}-onevision: Easy visual task transfer}.
\newblock \emph{Transactions on Machine Learning Research}.

\bibitem[{Li et~al.(2023{\natexlab{a}})Li, Pan, Ge, Gao, Ji, Zhang, Chua, Tang,
  Zhang, and Zhuang}]{li2023fine}
Juncheng Li, Kaihang Pan, Zhiqi Ge, Minghe Gao, Wei Ji, Wenqiao Zhang, Tat-Seng
  Chua, Siliang Tang, Hanwang Zhang, and Yueting Zhuang. 2023{\natexlab{a}}.
\newblock \href {https://arxiv.org/abs/2308.04152v4} {Fine-tuning multimodal
  llms to follow zero-shot demonstrative instructions}.
\newblock \emph{arXiv preprint arXiv:2308.04152}.

\bibitem[{Li et~al.(2023{\natexlab{b}})Li, Yin, Li, Chen, Wang, Ren, Li, Yang,
  Xu, Sun et~al.}]{li2023m}
Lei Li, Yuwei Yin, Shicheng Li, Liang Chen, Peiyi Wang, Shuhuai Ren, Mukai Li,
  Yazheng Yang, Jingjing Xu, Xu~Sun, et~al. 2023{\natexlab{b}}.
\newblock \href {https://arxiv.org/abs/2306.04387} {M$^3$it: A large-scale
  dataset towards multi-modal multilingual instruction tuning}.
\newblock \emph{arXiv preprint arXiv:2306.04387}.

\bibitem[{Li et~al.(2019)Li, Gan, Shen, Liu, Cheng, Wu, Carin, Carlson, and
  Gao}]{li2019storygan}
Yitong Li, Zhe Gan, Yelong Shen, Jingjing Liu, Yu~Cheng, Yuexin Wu, Lawrence
  Carin, David Carlson, and Jianfeng Gao. 2019.
\newblock \href {https://doi.org/10.1109/CVPR.2019.00649} {Storygan: A
  sequential conditional gan for story visualization}.
\newblock In \emph{2019 IEEE/CVF Conference on Computer Vision and Pattern
  Recognition (CVPR)}, pages 6322--6331.

\bibitem[{Li et~al.(2023{\natexlab{c}})Li, Wong, Stengel-Eskin, Kortylewski,
  Ma, Van~Durme, and Yuille}]{li2023super}
Zhuowan Li, Xingrui Wong, Elias Stengel-Eskin, Adam Kortylewski, Wufei Ma,
  Benjamin Van~Durme, and Alan Yuille. 2023{\natexlab{c}}.
\newblock \href {https://doi.org/10.1109/CVPR52729.2023.01437} {{ Super-CLEVR:
  A Virtual Benchmark to Diagnose Domain Robustness in Visual Reasoning }}.
\newblock In \emph{2023 IEEE/CVF Conference on Computer Vision and Pattern
  Recognition (CVPR)}, pages 14963--14973, Los Alamitos, CA, USA. IEEE Computer
  Society.

\bibitem[{Lin et~al.(2014)Lin, Maire, Belongie, Hays, Perona, Ramanan,
  Doll{\'a}r, and Zitnick}]{lin2014microsoft}
Tsung-Yi Lin, Michael Maire, Serge Belongie, James Hays, Pietro Perona, Deva
  Ramanan, Piotr Doll{\'a}r, and C~Lawrence Zitnick. 2014.
\newblock \href {https://arxiv.org/abs/1405.0312} {Microsoft coco: Common
  objects in context}.
\newblock In \emph{Computer vision--ECCV 2014: 13th European conference,
  zurich, Switzerland, September 6-12, 2014, proceedings, part v 13}, pages
  740--755. Springer.

\bibitem[{Liu et~al.(2023{\natexlab{a}})Liu, Lin, Li, Wang, Yacoob, and
  Wang}]{liu2023mitigating}
Fuxiao Liu, Kevin Lin, Linjie Li, Jianfeng Wang, Yaser Yacoob, and Lijuan Wang.
  2023{\natexlab{a}}.
\newblock \href {https://arxiv.org/abs/2306.14565} {Mitigating hallucination in
  large multi-modal models via robust instruction tuning}.
\newblock \emph{arXiv preprint arXiv:2306.14565}.

\bibitem[{Liu et~al.(2024{\natexlab{a}})Liu, Li, Li, and Lee}]{llava_1_5}
Haotian Liu, Chunyuan Li, Yuheng Li, and Yong~Jae Lee. 2024{\natexlab{a}}.
\newblock \href
  {https://openaccess.thecvf.com/content/CVPR2024/html/Liu_Improved_Baselines_with_Visual_Instruction_Tuning_CVPR_2024_paper.html}
  {Improved baselines with visual instruction tuning}.
\newblock In \emph{Proceedings of the IEEE/CVF Conference on Computer Vision
  and Pattern Recognition (CVPR)}, pages 26296--26306.

\bibitem[{Liu et~al.(2024{\natexlab{b}})Liu, Li, Li, Li, Zhang, Shen, and
  Lee}]{liu2024llavanext}
Haotian Liu, Chunyuan Li, Yuheng Li, Bo~Li, Yuanhan Zhang, Sheng Shen, and
  Yong~Jae Lee. 2024{\natexlab{b}}.
\newblock \href {https://llava-vl.github.io/blog/2024-01-30-llava-next/}
  {Llava-next: Improved reasoning, ocr, and world knowledge}.

\bibitem[{Liu et~al.(2023{\natexlab{b}})Liu, Li, Wu, and Lee}]{llava}
Haotian Liu, Chunyuan Li, Qingyang Wu, and Yong~Jae Lee. 2023{\natexlab{b}}.
\newblock \href
  {https://proceedings.neurips.cc/paper_files/paper/2023/file/6dcf277ea32ce3288914faf369fe6de0-Paper-Conference.pdf}
  {Visual instruction tuning}.
\newblock In \emph{Advances in Neural Information Processing Systems},
  volume~36, pages 34892--34916. Curran Associates, Inc.

\bibitem[{Liu et~al.(2022)Liu, Xu, Xu, Qian, Li, Ji, Chan, and
  Jin}]{koyejo2022}
Ziquan Liu, Yi~Xu, Yuanhong Xu, Qi~Qian, Hao Li, Xiangyang Ji, Antoni Chan, and
  Rong Jin. 2022.
\newblock \href
  {https://proceedings.neurips.cc/paper_files/paper/2022/file/d1c88f9790765146ec8fb5d02e5653a0-Paper-Conference.pdf}
  {Improved fine-tuning by better leveraging pre-training data}.
\newblock In \emph{Advances in Neural Information Processing Systems},
  volume~35, pages 32568--32581. Curran Associates, Inc.

\bibitem[{Lu et~al.(2021{\natexlab{a}})Lu, Gong, Jiang, Qiu, Huang, Liang, and
  Zhu}]{lu2021inter}
Pan Lu, Ran Gong, Shibiao Jiang, Liang Qiu, Siyuan Huang, Xiaodan Liang, and
  Song-Chun Zhu. 2021{\natexlab{a}}.
\newblock \href {https://doi.org/10.18653/v1/2021.acl-long.528} {{I}nter-{GPS}:
  Interpretable geometry problem solving with formal language and symbolic
  reasoning}.
\newblock In \emph{Proceedings of the 59th Annual Meeting of the Association
  for Computational Linguistics and the 11th International Joint Conference on
  Natural Language Processing (Volume 1: Long Papers)}, pages 6774--6786,
  Online. Association for Computational Linguistics.

\bibitem[{Lu et~al.(2022)Lu, Mishra, Xia, Qiu, Chang, Zhu, Tafjord, Clark, and
  Kalyan}]{lu2022learn}
Pan Lu, Swaroop Mishra, Tanglin Xia, Liang Qiu, Kai-Wei Chang, Song-Chun Zhu,
  Oyvind Tafjord, Peter Clark, and Ashwin Kalyan. 2022.
\newblock \href
  {https://proceedings.neurips.cc/paper_files/paper/2022/file/11332b6b6cf4485b84afadb1352d3a9a-Paper-Conference.pdf}
  {Learn to explain: Multimodal reasoning via thought chains for science
  question answering}.
\newblock In \emph{Advances in Neural Information Processing Systems},
  volume~35, pages 2507--2521. Curran Associates, Inc.

\bibitem[{Lu et~al.(2023)Lu, Qiu, Chang, Wu, Zhu, Rajpurohit, Clark, and
  Kalyan}]{lu2023dynamic}
Pan Lu, Liang Qiu, Kai-Wei Chang, Ying~Nian Wu, Song-Chun Zhu, Tanmay
  Rajpurohit, Peter Clark, and Ashwin Kalyan. 2023.
\newblock \href {https://arxiv.org/abs/2209.14610} {Dynamic prompt learning via
  policy gradient for semi-structured mathematical reasoning}.
\newblock In \emph{International Conference on Learning Representations
  (ICLR)}.

\bibitem[{Lu et~al.(2021{\natexlab{b}})Lu, Qiu, Chen, Xia, Zhao, Zhang, Yu,
  Liang, and Zhu}]{lu2021iconqa}
Pan Lu, Liang Qiu, Jiaqi Chen, Tony Xia, Yizhou Zhao, Wei Zhang, Zhou Yu,
  Xiaodan Liang, and Song-Chun Zhu. 2021{\natexlab{b}}.
\newblock \href {https://arxiv.org/abs/2110.13214} {Iconqa: A new benchmark for
  abstract diagram understanding and visual language reasoning}.
\newblock In \emph{Advances in Neural Information Processing Systems
  (NeurIPS)}.

\bibitem[{Maaz et~al.(2024)Maaz, Rasheed, Shaker, Khan, Cholakal, Anwer,
  Baldwin, Felsberg, and Khan}]{palo}
Muhammad Maaz, Hanoona Rasheed, Abdelrahman Shaker, Salman Khan, Hisham
  Cholakal, Rao~M Anwer, Tim Baldwin, Michael Felsberg, and Fahad~S Khan. 2024.
\newblock \href {https://arxiv.org/abs/2402.14818} {Palo: A polyglot large
  multimodal model for 5b people}.
\newblock \emph{arXiv preprint arXiv:2402.14818}.

\bibitem[{Marino et~al.(2019)Marino, Rastegari, Farhadi, and
  Mottaghi}]{marino2019ok}
Kenneth Marino, Mohammad Rastegari, Ali Farhadi, and Roozbeh Mottaghi. 2019.
\newblock \href {https://arxiv.org/abs/1906.00067} {{OK-VQA: A Visual Question
  Answering Benchmark Requiring External Knowledge}}.
\newblock In \emph{Proceedings of the IEEE/cvf conference on computer vision
  and pattern recognition}, pages 3195--3204.

\bibitem[{Marti and Bunke(2002)}]{marti2002iam}
U.-V. Marti and H.~Bunke. 2002.
\newblock \href {https://doi.org/10.1007/s100320200071} {The iam-database: an
  english sentence database for offline handwriting recognition}.
\newblock \emph{International Journal on Document Analysis and Recognition},
  5(1):39--46.

\bibitem[{Masry et~al.(2022)Masry, Do, Tan, Joty, and Hoque}]{masry2022chartqa}
Ahmed Masry, Xuan~Long Do, Jia~Qing Tan, Shafiq Joty, and Enamul Hoque. 2022.
\newblock \href {https://doi.org/10.18653/v1/2022.findings-acl.177}
  {{C}hart{QA}: A benchmark for question answering about charts with visual and
  logical reasoning}.
\newblock In \emph{Findings of the Association for Computational Linguistics:
  ACL 2022}, pages 2263--2279, Dublin, Ireland. Association for Computational
  Linguistics.

\bibitem[{Mathew et~al.(2022)Mathew, Bagal, Tito, Karatzas, Valveny, and
  Jawahar}]{mathew2022infographicvqa}
Minesh Mathew, Viraj Bagal, Rubèn Tito, Dimosthenis Karatzas, Ernest Valveny,
  and C.~V. Jawahar. 2022.
\newblock \href {https://doi.org/10.1109/WACV51458.2022.00264}
  {{InfographicVQA}}.
\newblock In \emph{2022 IEEE/CVF Winter Conference on Applications of Computer
  Vision (WACV)}, pages 2582--2591.

\bibitem[{Mathew et~al.(2021)Mathew, Karatzas, and Jawahar}]{mathew2021docvqa}
Minesh Mathew, Dimosthenis Karatzas, and C.~V. Jawahar. 2021.
\newblock \href {https://doi.org/10.1109/WACV48630.2021.00225} {Docvqa: A
  dataset for vqa on document images}.
\newblock In \emph{2021 IEEE Winter Conference on Applications of Computer
  Vision (WACV)}, pages 2199--2208.

\bibitem[{Mishra et~al.(2012)Mishra, Alahari, and Jawahar}]{mishra2012scene}
Anand Mishra, Karteek Alahari, and CV~Jawahar. 2012.
\newblock \href {https://inria.hal.science/hal-00818183/document/} {Scene text
  recognition using higher order language priors}.
\newblock In \emph{BMVC-British machine vision conference}. BMVA.

\bibitem[{Mishra et~al.(2019)Mishra, Shekhar, Singh, and
  Chakraborty}]{mishra2019ocr}
Anand Mishra, Shashank Shekhar, Ajeet~Kumar Singh, and Anirban Chakraborty.
  2019.
\newblock \href {https://doi.org/10.1109/ICDAR.2019.00156} {Ocr-vqa: Visual
  question answering by reading text in images}.
\newblock In \emph{2019 International Conference on Document Analysis and
  Recognition (ICDAR)}, pages 947--952.

\bibitem[{Obeid and Hoque(2020)}]{obeid2020chart}
Jason Obeid and Enamul Hoque. 2020.
\newblock \href {https://doi.org/10.18653/v1/2020.inlg-1.20} {Chart-to-text:
  Generating natural language descriptions for charts by adapting the
  transformer model}.
\newblock In \emph{Proceedings of the 13th International Conference on Natural
  Language Generation}, pages 138--147, Dublin, Ireland. Association for
  Computational Linguistics.

\bibitem[{Park et~al.(2019)Park, Darrell, and Rohrbach}]{park2019robust}
Dong~Huk Park, Trevor Darrell, and Anna Rohrbach. 2019.
\newblock \href {https://doi.org/10.1109/ICCV.2019.00472} {Robust change
  captioning}.
\newblock In \emph{2019 IEEE/CVF International Conference on Computer Vision
  (ICCV)}, pages 4623--4632.

\bibitem[{Pi et~al.(2024)Pi, Zhang, Zhang, Pan, Chen, and Zhang}]{pi2024image}
Renjie Pi, Jianshu Zhang, Jipeng Zhang, Rui Pan, Zhekai Chen, and Tong Zhang.
  2024.
\newblock \href {https://arxiv.org/abs/2406.07502} {Image textualization: An
  automatic framework for creating accurate and detailed image descriptions}.
\newblock \emph{arXiv preprint arXiv:2406.07502}.

\bibitem[{Popovi{\'c}(2016)}]{chrf}
Maja Popovi{\'c}. 2016.
\newblock \href {https://doi.org/10.18653/v1/W16-2341} {chr{F} deconstructed:
  beta parameters and n-gram weights}.
\newblock In \emph{Proceedings of the First Conference on Machine Translation:
  Volume 2, Shared Task Papers}, pages 499--504, Berlin, Germany. Association
  for Computational Linguistics.

\bibitem[{Popovi{\'c}(2017)}]{chrf++}
Maja Popovi{\'c}. 2017.
\newblock \href {https://doi.org/10.18653/v1/W17-4770} {chr{F}++: words helping
  character n-grams}.
\newblock In \emph{Proceedings of the Second Conference on Machine
  Translation}, pages 612--618, Copenhagen, Denmark. Association for
  Computational Linguistics.

\bibitem[{Qiu et~al.(2022)Qiu, Onea{\textcommabelow{t}}{\u{a}}, Bugliarello,
  Frank, and Elliott}]{qiu-etal-2022-multilingual}
Chen Qiu, Dan Onea{\textcommabelow{t}}{\u{a}}, Emanuele Bugliarello, Stella
  Frank, and Desmond Elliott. 2022.
\newblock \href {https://doi.org/10.18653/v1/2022.findings-emnlp.308}
  {Multilingual multimodal learning with machine translated text}.
\newblock In \emph{Findings of the Association for Computational Linguistics:
  EMNLP 2022}, pages 4178--4193, Abu Dhabi, United Arab Emirates. Association
  for Computational Linguistics.

\bibitem[{Radford et~al.(2021)Radford, Kim, Hallacy, Ramesh, Goh, Agarwal,
  Sastry, Askell, Mishkin, Clark et~al.}]{clip}
Alec Radford, Jong~Wook Kim, Chris Hallacy, Aditya Ramesh, Gabriel Goh,
  Sandhini Agarwal, Girish Sastry, Amanda Askell, Pamela Mishkin, Jack Clark,
  et~al. 2021.
\newblock \href {https://arxiv.org/abs/2103.00020} {Learning transferable
  visual models from natural language supervision}.
\newblock In \emph{International conference on machine learning}, pages
  8748--8763. PmLR.

\bibitem[{Rajani et~al.(2023)Rajani, Tunstall, Beeching, Lambert, Rush, and
  Wolf}]{no_robots}
Nazneen Rajani, Lewis Tunstall, Edward Beeching, Nathan Lambert, Alexander~M.
  Rush, and Thomas Wolf. 2023.
\newblock No robots.
\newblock \url{https://huggingface.co/datasets/HuggingFaceH4/no_robots}.

\bibitem[{Ravi et~al.(2021)Ravi, Kafle, Cohen, Brandt, and
  Kapadia}]{ravi2021aesop}
Hareesh Ravi, Kushal Kafle, Scott Cohen, Jonathan Brandt, and Mubbasir Kapadia.
  2021.
\newblock \href {https://doi.org/10.1109/ICCV48922.2021.00206} {Aesop: Abstract
  encoding of stories, objects, and pictures}.
\newblock In \emph{2021 IEEE/CVF International Conference on Computer Vision
  (ICCV)}, pages 2032--2043.

\bibitem[{Schneider and Sitaram(2024)}]{m5benchmark}
Florian Schneider and Sunayana Sitaram. 2024.
\newblock \href {https://doi.org/10.18653/v1/2024.findings-emnlp.250} {M5 {--}
  a diverse benchmark to assess the performance of large multimodal models
  across multilingual and multicultural vision-language tasks}.
\newblock In \emph{Findings of the Association for Computational Linguistics:
  EMNLP 2024}, pages 4309--4345, Miami, Florida, USA. Association for
  Computational Linguistics.

\bibitem[{Schwenk et~al.(2022)Schwenk, Khandelwal, Clark, Marino, and
  Mottaghi}]{schwenk2022okvqa}
Dustin Schwenk, Apoorv Khandelwal, Christopher Clark, Kenneth Marino, and
  Roozbeh Mottaghi. 2022.
\newblock \href {https://arxiv.org/abs/2206.01718} {A-okvqa: A benchmark for
  visual question answering using world knowledge}.
\newblock In \emph{European conference on computer vision}, pages 146--162.
  Springer.

\bibitem[{Seo et~al.(2015)Seo, Hajishirzi, Farhadi, Etzioni, and
  Malcolm}]{seo2015solving}
Minjoon Seo, Hannaneh Hajishirzi, Ali Farhadi, Oren Etzioni, and Clint Malcolm.
  2015.
\newblock \href {https://doi.org/10.18653/v1/D15-1171} {Solving geometry
  problems: Combining text and diagram interpretation}.
\newblock In \emph{Proceedings of the 2015 Conference on Empirical Methods in
  Natural Language Processing}, pages 1466--1476, Lisbon, Portugal. Association
  for Computational Linguistics.

\bibitem[{Shoemake(1985)}]{shoemake1985animating}
Ken Shoemake. 1985.
\newblock Animating rotation with quaternion curves.
\newblock In \emph{Proceedings of the 12th annual conference on Computer
  graphics and interactive techniques}, pages 245--254. ACM.

\bibitem[{Shridhar et~al.(2020)Shridhar, Thomason, Gordon, Bisk, Han, Mottaghi,
  Zettlemoyer, and Fox}]{shridhar2020alfred}
Mohit Shridhar, Jesse Thomason, Daniel Gordon, Yonatan Bisk, Winson Han,
  Roozbeh Mottaghi, Luke Zettlemoyer, and Dieter Fox. 2020.
\newblock \href {https://doi.org/10.1109/CVPR42600.2020.01075} {Alfred: A
  benchmark for interpreting grounded instructions for everyday tasks}.
\newblock In \emph{2020 IEEE/CVF Conference on Computer Vision and Pattern
  Recognition (CVPR)}, pages 10737--10746.

\bibitem[{Sidorov et~al.(2020)Sidorov, Hu, Rohrbach, and
  Singh}]{sidorov2020textcaps}
Oleksii Sidorov, Ronghang Hu, Marcus Rohrbach, and Amanpreet Singh. 2020.
\newblock \href {https://arxiv.org/abs/2003.12462} {Textcaps: a dataset for
  image captioning with reading comprehension}.
\newblock In \emph{Computer Vision--ECCV 2020: 16th European Conference,
  Glasgow, UK, August 23--28, 2020, Proceedings, Part II 16}, pages 742--758.
  Springer.

\bibitem[{Sigurdsson et~al.(2016)Sigurdsson, Varol, Wang, Farhadi, Laptev, and
  Gupta}]{charades}
Gunnar~A. Sigurdsson, G{\"u}l Varol, Xiaolong Wang, Ali Farhadi, Ivan Laptev,
  and Abhinav Gupta. 2016.
\newblock \href {https://arxiv.org/abs/1604.01753} {Hollywood in homes:
  Crowdsourcing data collection for activity understanding}.
\newblock In \emph{Computer Vision -- ECCV 2016}, pages 510--526, Cham.
  Springer International Publishing.

\bibitem[{Singh et~al.(2024)Singh, Vargus, D{'}souza, Karlsson, Mahendiran, Ko,
  Shandilya, Patel, Mataciunas, O{'}Mahony, Zhang, Hettiarachchi, Wilson,
  Machado, Moura, Krzemi{\'n}ski, Fadaei, Ergun, Okoh, Alaagib, Mudannayake,
  Alyafeai, Chien, Ruder, Guthikonda, Alghamdi, Gehrmann, Muennighoff, Bartolo,
  Kreutzer, {\"U}st{\"u}n, Fadaee, and Hooker}]{singh-etal-2024-aya}
Shivalika Singh, Freddie Vargus, Daniel D{'}souza, B{\"o}rje Karlsson, Abinaya
  Mahendiran, Wei-Yin Ko, Herumb Shandilya, Jay Patel, Deividas Mataciunas,
  Laura O{'}Mahony, Mike Zhang, Ramith Hettiarachchi, Joseph Wilson, Marina
  Machado, Luisa Moura, Dominik Krzemi{\'n}ski, Hakimeh Fadaei, Irem Ergun,
  Ifeoma Okoh, Aisha Alaagib, Oshan Mudannayake, Zaid Alyafeai, Vu~Chien,
  Sebastian Ruder, Surya Guthikonda, Emad Alghamdi, Sebastian Gehrmann, Niklas
  Muennighoff, Max Bartolo, Julia Kreutzer, Ahmet {\"U}st{\"u}n, Marzieh
  Fadaee, and Sara Hooker. 2024.
\newblock \href {https://doi.org/10.18653/v1/2024.acl-long.620} {Aya dataset:
  An open-access collection for multilingual instruction tuning}.
\newblock In \emph{Proceedings of the 62nd Annual Meeting of the Association
  for Computational Linguistics (Volume 1: Long Papers)}, pages 11521--11567,
  Bangkok, Thailand. Association for Computational Linguistics.

\bibitem[{Song et~al.(2024)Song, Khanuja, and Neubig}]{song2024missing}
Yueqi Song, Simran Khanuja, and Graham Neubig. 2024.
\newblock \href {https://arxiv.org/abs/2403.01404} {What is missing in
  multilingual visual reasoning and how to fix it}.
\newblock \emph{arXiv preprint arXiv:2403.01404}.

\bibitem[{Suhr et~al.(2019)Suhr, Zhou, Zhang, Zhang, Bai, and
  Artzi}]{suhr2019corpus}
Alane Suhr, Stephanie Zhou, Ally Zhang, Iris Zhang, Huajun Bai, and Yoav Artzi.
  2019.
\newblock \href {https://doi.org/10.18653/v1/P19-1644} {A corpus for reasoning
  about natural language grounded in photographs}.
\newblock In \emph{Proceedings of the 57th Annual Meeting of the Association
  for Computational Linguistics}, pages 6418--6428, Florence, Italy.
  Association for Computational Linguistics.

\bibitem[{Tan et~al.(2019)Tan, Dernoncourt, Lin, Bui, and
  Bansal}]{tan2019expressing}
Hao Tan, Franck Dernoncourt, Zhe Lin, Trung Bui, and Mohit Bansal. 2019.
\newblock \href {https://doi.org/10.18653/v1/P19-1182} {Expressing visual
  relationships via language}.
\newblock In \emph{Proceedings of the 57th Annual Meeting of the Association
  for Computational Linguistics}, pages 1873--1883, Florence, Italy.
  Association for Computational Linguistics.

\bibitem[{Tanaka et~al.(2021)Tanaka, Nishida, and
  Yoshida}]{tanaka2021visualmrc}
Ryota Tanaka, Kyosuke Nishida, and Sen Yoshida. 2021.
\newblock \href {https://doi.org/10.1609/aaai.v35i15.17635} {{VisualMRC:
  Machine Reading Comprehension on Document Images}}.
\newblock \emph{Proceedings of the AAAI Conference on Artificial Intelligence},
  35(15):13878--13888.

\bibitem[{Tang et~al.(2023)Tang, Boggust, and Satyanarayan}]{tang2023vistext}
Benny Tang, Angie Boggust, and Arvind Satyanarayan. 2023.
\newblock \href {https://doi.org/10.18653/v1/2023.acl-long.401} {{V}is{T}ext: A
  benchmark for semantically rich chart captioning}.
\newblock In \emph{Proceedings of the 61st Annual Meeting of the Association
  for Computational Linguistics (Volume 1: Long Papers)}, pages 7268--7298,
  Toronto, Canada. Association for Computational Linguistics.

\bibitem[{Thapliyal et~al.(2022)Thapliyal, Pont~Tuset, Chen, and
  Soricut}]{xm3600}
Ashish~V. Thapliyal, Jordi Pont~Tuset, Xi~Chen, and Radu Soricut. 2022.
\newblock \href {https://doi.org/10.18653/v1/2022.emnlp-main.45}
  {Crossmodal-3600: A massively multilingual multimodal evaluation dataset}.
\newblock In \emph{Proceedings of the 2022 Conference on Empirical Methods in
  Natural Language Processing}, pages 715--729, Abu Dhabi, United Arab
  Emirates. Association for Computational Linguistics.

\bibitem[{Tong et~al.(2024)Tong, Brown, Wu, Woo, IYER, Akula, Yang, Yang,
  Middepogu, Wang et~al.}]{tong2024cambrian}
Peter Tong, Ellis Brown, Penghao Wu, Sanghyun Woo, Adithya Jairam~Vedagiri
  IYER, Sai~Charitha Akula, Shusheng Yang, Jihan Yang, Manoj Middepogu, Ziteng
  Wang, et~al. 2024.
\newblock \href {https://neurips.cc/virtual/2024/poster/94880} {Cambrian-1: A
  fully open, vision-centric exploration of multimodal llms}.
\newblock \emph{Advances in Neural Information Processing Systems},
  37:87310--87356.

\bibitem[{Wang et~al.(2021)Wang, Li, Zhou, Chen, Grossman, and
  Li}]{wang2021screen2words}
Bryan Wang, Gang Li, Xin Zhou, Zhourong Chen, Tovi Grossman, and Yang Li. 2021.
\newblock \href {https://dl.acm.org/doi/fullHtml/10.1145/3472749.3474765}
  {Screen2words: Automatic mobile ui summarization with multimodal learning}.
\newblock In \emph{The 34th Annual ACM Symposium on User Interface Software and
  Technology}, pages 498--510.

\bibitem[{Wu et~al.(2024)Wu, Zhu, Zhang, Zhang, Chen, Liao, Li, Wang, Sun, Yan
  et~al.}]{wu2024towards}
Haoning Wu, Hanwei Zhu, Zicheng Zhang, Erli Zhang, Chaofeng Chen, Liang Liao,
  Chunyi Li, Annan Wang, Wenxiu Sun, Qiong Yan, et~al. 2024.
\newblock \href {https://arxiv.org/abs/2402.16641} {Towards open-ended visual
  quality comparison}.
\newblock In \emph{European Conference on Computer Vision}, pages 360--377.
  Springer.

\bibitem[{Xiao et~al.(2021)Xiao, Shang, Yao, and Chua}]{xiao2021next}
Junbin Xiao, Xindi Shang, Angela Yao, and Tat-Seng Chua. 2021.
\newblock \href {https://doi.org/10.1109/CVPR46437.2021.00965} {{ NExT-QA: Next
  Phase of Question-Answering to Explaining Temporal Actions }}.
\newblock In \emph{2021 IEEE/CVF Conference on Computer Vision and Pattern
  Recognition (CVPR)}, pages 9772--9781, Los Alamitos, CA, USA. IEEE Computer
  Society.

\bibitem[{Xu et~al.(2024{\natexlab{a}})Xu, Jiang, Niu, Deng, Poovendran, Choi,
  and Lin}]{xu2024magpie}
Zhangchen Xu, Fengqing Jiang, Luyao Niu, Yuntian Deng, Radha Poovendran, Yejin
  Choi, and Bill~Yuchen Lin. 2024{\natexlab{a}}.
\newblock \href {https://arxiv.org/abs/2406.08464} {Magpie: Alignment data
  synthesis from scratch by prompting aligned llms with nothing}.
\newblock \emph{arXiv preprint arXiv:2406.08464}.

\bibitem[{Xu et~al.(2024{\natexlab{b}})Xu, Feng, Shao, Ashby, Shen, Jin, Cheng,
  Wang, and Huang}]{xu2024vision}
Zhiyang Xu, Chao Feng, Rulin Shao, Trevor Ashby, Ying Shen, Di~Jin, Yu~Cheng,
  Qifan Wang, and Lifu Huang. 2024{\natexlab{b}}.
\newblock \href {https://doi.org/10.18653/v1/2024.findings-acl.905}
  {{Vision-Flan: Scaling Human-Labeled Tasks in Visual Instruction Tuning}}.
\newblock In \emph{Findings of the Association for Computational Linguistics:
  ACL 2024}, pages 15271--15342, Bangkok, Thailand. Association for
  Computational Linguistics.

\bibitem[{Yagcioglu et~al.(2018)Yagcioglu, Erdem, Erdem, and
  Ikizler-Cinbis}]{yagcioglu2018recipeqa}
Semih Yagcioglu, Aykut Erdem, Erkut Erdem, and Nazli Ikizler-Cinbis. 2018.
\newblock \href {https://doi.org/10.18653/v1/D18-1166} {{R}ecipe{QA}: A
  challenge dataset for multimodal comprehension of cooking recipes}.
\newblock In \emph{Proceedings of the 2018 Conference on Empirical Methods in
  Natural Language Processing}, pages 1358--1368, Brussels, Belgium.
  Association for Computational Linguistics.

\bibitem[{Ye et~al.(2023)Ye, Hu, Xu, Ye, Yan, Xu, Li, Tian, Qian, Zhang, Jin,
  He, Lin, and Huang}]{ye-etal-2023-ureader}
Jiabo Ye, Anwen Hu, Haiyang Xu, Qinghao Ye, Ming Yan, Guohai Xu, Chenliang Li,
  Junfeng Tian, Qi~Qian, Ji~Zhang, Qin Jin, Liang He, Xin Lin, and Fei Huang.
  2023.
\newblock \href {https://doi.org/10.18653/v1/2023.findings-emnlp.187}
  {{UR}eader: Universal {OCR}-free visually-situated language understanding
  with multimodal large language model}.
\newblock In \emph{Findings of the Association for Computational Linguistics:
  EMNLP 2023}, pages 2841--2858, Singapore. Association for Computational
  Linguistics.

\bibitem[{Yu et~al.(2016)Yu, Poirson, Yang, Berg, and Berg}]{yu2016modeling}
Licheng Yu, Patrick Poirson, Shan Yang, Alexander~C Berg, and Tamara~L Berg.
  2016.
\newblock \href {https://arxiv.org/abs/1608.00272} {Modeling context in
  referring expressions}.
\newblock In \emph{Computer Vision--ECCV 2016: 14th European Conference,
  Amsterdam, The Netherlands, October 11-14, 2016, Proceedings, Part II 14},
  pages 69--85. Springer.

\bibitem[{Yu et~al.(2019)Yu, Xu, Yu, Yu, Zhao, Zhuang, and
  Tao}]{yu2019activityqa}
Zhou Yu, Dejing Xu, Jun Yu, Ting Yu, Zhou Zhao, Yueting Zhuang, and Dacheng
  Tao. 2019.
\newblock \href {https://arxiv.org/abs/1906.02467} {Activitynet-qa: A dataset
  for understanding complex web videos via question answering}.
\newblock In \emph{AAAI}, pages 9127--9134.

\bibitem[{Yuan et~al.(2022)Yuan, Liu, Dikubab, Liu, Ji, Wu, and
  Bai}]{yuan2022syntax}
Ye~Yuan, Xiao Liu, Wondimu Dikubab, Hui Liu, Zhilong Ji, Zhongqin Wu, and Xiang
  Bai. 2022.
\newblock \href {https://doi.org/10.1109/CVPR52688.2022.00451} {Syntax-aware
  network for handwritten mathematical expression recognition}.
\newblock In \emph{2022 IEEE/CVF Conference on Computer Vision and Pattern
  Recognition (CVPR)}, pages 4543--4552.

\bibitem[{Yue et~al.(2024)Yue, Ni, Zhang, Zheng, Liu, Zhang, Stevens, Jiang,
  Ren, Sun, Wei, Yu, Yuan, Sun, Yin, Zheng, Yang, Liu, Huang, Sun, Su, and
  Chen}]{mmmu}
Xiang Yue, Yuansheng Ni, Kai Zhang, Tianyu Zheng, Ruoqi Liu, Ge~Zhang, Samuel
  Stevens, Dongfu Jiang, Weiming Ren, Yuxuan Sun, Cong Wei, Botao Yu, Ruibin
  Yuan, Renliang Sun, Ming Yin, Boyuan Zheng, Zhenzhu Yang, Yibo Liu, Wenhao
  Huang, Huan Sun, Yu~Su, and Wenhu Chen. 2024.
\newblock \href
  {https://openaccess.thecvf.com/content/CVPR2024/papers/Yue_MMMU_A_Massive_Multi-discipline_Multimodal_Understanding_and_Reasoning_Benchmark_for_CVPR_2024_paper.pdf}
  {Mmmu: A massive multi-discipline multimodal understanding and reasoning
  benchmark for expert agi}.
\newblock In \emph{Proceedings of the IEEE/CVF Conference on Computer Vision
  and Pattern Recognition (CVPR)}, pages 9556--9567.

\bibitem[{Yue et~al.(2025)Yue, Song, Asai, Kim, de~Dieu~Nyandwi, Khanuja,
  Kantharuban, Sutawika, Ramamoorthy, and Neubig}]{pangea}
Xiang Yue, Yueqi Song, Akari Asai, Seungone Kim, Jean de~Dieu~Nyandwi, Simran
  Khanuja, Anjali Kantharuban, Lintang Sutawika, Sathyanarayanan Ramamoorthy,
  and Graham Neubig. 2025.
\newblock \href {https://openreview.net/forum?id=a3g2l4yEys} {Pangea: A fully
  open multilingual multimodal {LLM} for 39 languages}.
\newblock In \emph{The Thirteenth International Conference on Learning
  Representations}.

\bibitem[{Zhai et~al.(2023)Zhai, Mustafa, Kolesnikov, and Beyer}]{siglip}
Xiaohua Zhai, Basil Mustafa, Alexander Kolesnikov, and Lucas Beyer. 2023.
\newblock \href
  {https://openaccess.thecvf.com/content/ICCV2023/html/Zhai_Sigmoid_Loss_for_Language_Image_Pre-Training_ICCV_2023_paper.html}
  {Sigmoid loss for language image pre-training}.
\newblock In \emph{Proceedings of the IEEE/CVF International Conference on
  Computer Vision (ICCV)}, pages 11975--11986.

\bibitem[{Zhang et~al.(2019)Zhang, Gao, Jia, Zhu, and Zhu}]{zhang2019raven}
Chi Zhang, Feng Gao, Baoxiong Jia, Yixin Zhu, and Song-Chun Zhu. 2019.
\newblock \href {https://doi.org/10.1109/CVPR.2019.00546} {Raven: A dataset for
  relational and analogical visual reasoning}.
\newblock In \emph{2019 IEEE/CVF Conference on Computer Vision and Pattern
  Recognition (CVPR)}, pages 5312--5322.

\bibitem[{Zhang et~al.(2023{\natexlab{a}})Zhang, Mo, Chen, Sun, and
  Su}]{zhang2023magicbrush}
Kai Zhang, Lingbo Mo, Wenhu Chen, Huan Sun, and Yu~Su. 2023{\natexlab{a}}.
\newblock \href {https://neurips.cc/virtual/2023/poster/73543} {Magicbrush: A
  manually annotated dataset for instruction-guided image editing}.
\newblock \emph{Advances in Neural Information Processing Systems},
  36:31428--31449.

\bibitem[{Zhang et~al.(2024)Zhang, Wei, Jiang, Zhang, Guo, Tong, Liu, Zhou,
  Wei, Zhang, Gao, and Li}]{zhang2024mavis}
Renrui Zhang, Xinyu Wei, Dongzhi Jiang, Yichi Zhang, Ziyu Guo, Chengzhuo Tong,
  Jiaming Liu, Aojun Zhou, Bin Wei, Shanghang Zhang, Peng Gao, and Hongsheng
  Li. 2024.
\newblock \href {https://arxiv.org/abs/2407.08739} {Mavis: Mathematical visual
  instruction tuning}.
\newblock \emph{Preprint}, arXiv:2407.08739.

\bibitem[{Zhang et~al.(2023{\natexlab{b}})Zhang, Zhang, Gu, Zhou, Lipka, Yang,
  and Sun}]{zhang2023llavar}
Yanzhe Zhang, Ruiyi Zhang, Jiuxiang Gu, Yufan Zhou, Nedim Lipka, Diyi Yang, and
  Tong Sun. 2023{\natexlab{b}}.
\newblock \href {https://arxiv.org/abs/2306.17107} {Llavar: Enhanced visual
  instruction tuning for text-rich image understanding}.
\newblock \emph{arXiv preprint arXiv:2306.17107}.

\bibitem[{Zhao et~al.(2023)Zhao, Zhao, Nan, Qi, Zhang, Tang, Mi, and
  Radev}]{zhao-etal-2023-robut}
Yilun Zhao, Chen Zhao, Linyong Nan, Zhenting Qi, Wenlin Zhang, Xiangru Tang,
  Boyu Mi, and Dragomir Radev. 2023.
\newblock \href {https://doi.org/10.18653/v1/2023.acl-long.334} {{R}obu{T}: A
  systematic study of table {QA} robustness against human-annotated adversarial
  perturbations}.
\newblock In \emph{Proceedings of the 61st Annual Meeting of the Association
  for Computational Linguistics (Volume 1: Long Papers)}, pages 6064--6081,
  Toronto, Canada. Association for Computational Linguistics.

\bibitem[{Zheng et~al.(2025)Zheng, Qiu, Shi, and Ma}]{zheng2025survey}
Junhao Zheng, Shengjie Qiu, Chengming Shi, and Qianli Ma. 2025.
\newblock \href {https://doi.org/10.1145/3716629} {Towards lifelong learning of
  large language models: A survey}.
\newblock \emph{ACM Comput. Surv.}, 57(8).

\bibitem[{Zhou et~al.(2018)Zhou, Xu, and Corso}]{ZhXuCoCVPR18}
Luowei Zhou, Chenliang Xu, and Jason~J Corso. 2018.
\newblock \href
  {https://www.aaai.org/ocs/index.php/AAAI/AAAI18/paper/view/17344} {Towards
  automatic learning of procedures from web instructional videos}.
\newblock In \emph{AAAI Conference on Artificial Intelligence}.

\bibitem[{Zhu et~al.(2016)Zhu, Groth, Bernstein, and Fei-Fei}]{zhu2016visual7w}
Yuke Zhu, Oliver Groth, Michael Bernstein, and Li~Fei-Fei. 2016.
\newblock \href {https://doi.org/10.1109/CVPR.2016.540} {Visual7w: Grounded
  question answering in images}.
\newblock In \emph{2016 IEEE Conference on Computer Vision and Pattern
  Recognition (CVPR)}, pages 4995--5004.

\end{thebibliography}

\appendix

\section{Data Sources}
\label{app:data-sources}
This section showcases the visual datasets (Table~\ref{tab:visual-data}) and the text-only datasets (Table~\ref{tab:data-sources-text}) used throughout this work.

\section{Language Consistency Evaluation via LLM-as-a-judge}
\label{app:text_LLM_as_a_judge}
We have observed cases of code-switched generations, and in some cases, even if most of the sentence is generated in the target language, a few words may still appear in English. The primary goal of the textual LLM-as-a-judge evaluation is to address GlotLID's limitation to classify these cases as incorrect.

To effectively evaluate each sentence, we have designed a prompt (see Figure \ref{fig:prompt_text_llm_as_a_judge}) that instructs the evaluator to perform multiple tasks beyond basic language identification, enabling the computation of a language consistency score:

\begin{enumerate}
    \item Guess the language of the sentence (to compare with GlotLID, even if we will use GlotLID’s outputs).
    \item Assign a language consistency score (between 0 and 1).
    \item Determine whether the sentence is fully in the target language (a boolean value, where False indicates that at least one word appears in another language).
    \item Generate a summary explaining the decisions made by the model.
\end{enumerate}

This method allows us to evaluate the language consistency of our model at the word level from different perspectives, both through a numerical score and a boolean indicator.

\begin{figure*}[t]
    \centering
    \begin{tcolorbox}[
        colback=gray!8, 
        colframe=gray!127, 
        boxrule=0.8pt, 
        title={Language Consistency Evaluation Prompt Template}, 
        coltitle=white, 
        colbacktitle=gray!127, 
    ]

    Analyze the following text and determine the language it is written in.\\
- Identify the most likely language.\\
- Ensure the probability score is a single value, not a range or estimate.\\
- Determine a language consistency score between 0 and 1.0, where 1.0 means the text is entirely in one language, and 0.0 means it is completely incomprehensible.\\
- Lower the score proportionally if foreign words are present, but do not assign 0.0 unless the text is nonsensical.\\
- The language score must be a single number between 0 and 1.0.\\
- Indicate whether the text is completely written in the identified language (True or False).\\
- In both language consistency metrics, do not penalize for proper nouns, brand names, or commonly used foreign terms (e.g., 'software', 'email') that do not alter the overall language structure.\\
- Avoid unnecessary explanations. Summarize the feedback (reason of the mark) in at most 30 words.\\
\\
Use the exact format below:\\
\\
- Language: [language\_guess]\\
- Language Score: [single value between 0 and 1.0]\\
- Fully in Language: [True/False]\\
- Summary: [Concise explanation (max 30 words)]\\
\\
Keep your answer short and concise. The sentence to analyze is the following:\\
{\normalsize \textbf{\textit{<CAPTION GENERATED BY VISUAL SALAMANDRA>}}}\vspace{0.1cm}
    \end{tcolorbox}
    \caption{Prompt used to evaluate language consistency via LLM-as-a-judge. The evaluator model assesses the language fidelity of the caption generated by the VLM using multiple criteria. Note that this evaluation focuses solely on language fidelity, not the overall quality of the caption.}
  \label{fig:prompt_text_llm_as_a_judge}
    \vspace{-12pt}
\end{figure*}

For additional robustness, we have performed the text evaluations using three different generation configurations (defined in Table \ref{tab:parameters_text_llm_as_a_judge}) and then applied a majority voting.

\begin{table}[h!]
\centering
\small
\begin{tabular}{cccc}
\toprule
Conf. & Temperature & Top\_p & Max new tokens \\
\midrule
A & 0.6         & 0.7    & 50 \\
B & 0.8         & 0.6    & 50 \\
C & 1.0         & 0.5    & 50 \\
\bottomrule
\end{tabular}
\caption{Generation parameter settings for LLM-as-a-Judge evaluation.}
\label{tab:parameters_text_llm_as_a_judge}
\end{table}

As shown in Figure \ref{fig:prompt_text_llm_as_a_judge}, a structured response format was explicitly requested to ensure that each field could be reliably extracted from the output. The responses that did not conform to the expected format were replaced with "N/A".

To compute the final scores, we averaged the results across the three different configurations. For numerical scores, when a value was missing, we computed the average using the available values. For boolean scores, we applied a majority voting approach. In cases where one score was missing and the remaining two were True and False, we defaulted to False as the final instance score. This ensures that our results provide a lower bound, making the evaluation more conservative and reliable. All the results are shown in Table~\ref{tab:llm_judge_results}.

\begin{table*}[htbp]
\centering
\footnotesize
\resizebox{\textwidth}{!}{\begin{tabular}{llcccc|llcccc}
\toprule
\multicolumn{6}{c|}{} & \multicolumn{6}{c}{} \\
\cmidrule(lr){1-6} \cmidrule(lr){7-12}
\textbf{Model} & \textbf{Lang.} & \textbf{GL} & \textbf{LLM-L} & \textbf{LLM-S}  & \textbf{LLM-B} & \textbf{Model} & \textbf{Lang.} & \textbf{GL }& \textbf{LLM-L} & \textbf{LLM-S}  & \textbf{LLM-B}\\
\midrule

\multicolumn{12}{l}{\textbf{Normal Models}} \\
\midrule

\multirow{7}{*}{BM} & De & 2.7 & 2.4 & \textbf{85.8} & 73.5 & \multirow{7}{*}{TR-1S} & De & 24.5 & 18.3 & 84.0 & \textbf{75.9} \\
& En & \textbf{100.0} & 98.4 & 99.0 & 99.3 & & En & \textbf{100.0} & 98.1 & 98.7 & 99.1 \\
& Es & 4.4 & 4.7 & 91.5 & 87.2 & & Es & 38.4 & 37.9 & 94.4 & \textbf{95.4} \\
& Fr & 12.2 & 10.8 & 90.3 & 88.1 & & Fr & 29.9 & 26.8 & \textbf{91.2} & \textbf{91.1} \\
& Nl & 5.6 & 6.02 & 93.8 & 92.5 & & Nl & 49.2 & 48.1 & 93.4 & 93.1 \\
& No & 5.2 & 3.84 & 92.0 & 89.6 & & No & 46.8 & 30.3 & 92.8 & 91.5 \\
& Ru & 3.8 & 2.75 & 82.2 & 58.5 & & Ru & 50.9 & 28.4 & 82.2 & 62.9 \\

\cmidrule(lr){1-6} \cmidrule(lr){7-12}

\multirow{7}{*}{TR-2S} & De & 81.3 & 54.4 & 81.9 & 73.2 & \multirow{7}{*}{TR-3S} & De & 88.7 & \textbf{61.3} & 82.8 & 72.7 \\
& En & \textbf{100.0} & 98.3 & 98.9 & 99.4 & & En & \textbf{100.0} & \textbf{98.7} & \textbf{99.1} & \textbf{99.5} \\
& Es & 65.4 & 61.8 & 94.1 & 93.7 & & Es & \textbf{92.9} & \textbf{86.8} & \textbf{94.6} & 94.3 \\
& Fr & 74.9 & 63.8 & 90.5 & 89.5 & & Fr & 85.7 & 71.4 & 89.6 & 86.7 \\
& Nl & 91.3 & 84.9 & 93.4 & 94.3 & & Nl & 91.8 & 86.7 & 92.9 & 93.2 \\
& No & 52.0 & 40.4 & 93.7 & 93.3 & & No & 69.5 & 49.5 & 92.6 & 91.0 \\
& Ru & 24.8 & 13.4 & 81.6 & 63.1 & & Ru & 52.9 & 31.7 & \textbf{83.1} & 59.6 \\

\cmidrule(lr){1-6} \cmidrule(lr){7-12}

\multirow{7}{*}{TR-3S-0.0125x} & De & \textbf{85.0} & 58.4 & 82.1 & 75.2 & \multirow{7}{*}{TR-3S-0.025x} & De & 88.9 & 58.9 & 81.2 & 72.8 \\
& En & \textbf{100.0 }& 98.1 & 98.8 & 99.1 & & En & \textbf{100.0} & 98.4 & 98.7 & 99.4 \\
& Es & 91.9 & 85.9 & 94.1 & 94.8 & & Es & 92.4 & 85.9 & 94.0 & 93.7 \\
& Fr & \textbf{88.6} & \textbf{76.3} & 90.7 & 89.8 & & Fr & 83.5 & 71.2 & 90.0 & 88.8 \\
& Nl & 69.8 & 65.3 & 93.8 & 95.3 & & Nl & \textbf{96.0} & \textbf{90.6} & \textbf{94.1} & \textbf{96.0} \\
& No & 53.2 & 41.2 & \textbf{94.0} &\textbf{ 94.6 }& & No & \textbf{82.7} & \textbf{62.6} & 93.5 & 93.6 \\
& Ru & 91.5 & 48.5 & 81.6 & 63.2 & & Ru & 81.9 & 43.3 & 81.1 & 62.7 \\

\cmidrule(lr){1-6} \cmidrule(lr){7-12}

\multirow{7}{*}{TR-3S-0.1x} & De & 73.3 & 48.6 & 81.0 & 72.8 & \multirow{7}{*}{} &  &  &  &  &  \\
& En & \textbf{100.0} & \textbf{98.7} & 98.7 & 99.4 & &  &  &  &  &  \\
& Es & 76.4 & 71.7 & 94.3 & 95.2 & &  &  &  &  &  \\
& Fr & 69.5 & 59.1 & 90.0 & 88.6 & &  &  &  &  &  \\
& Nl & 93.5 & 88.3 & 93.8 & 95.9 & &  &  &  &  &  \\
& No & 67.1 & 53.7 & 93.9 & 94.0 & &  &  &  &  &  \\
& Ru & \textbf{93.3} & \textbf{48.9} & 81.3 & \textbf{64.2} & & &  &  &  &  \\

\midrule
\multicolumn{12}{l}{\textbf{Merged Models (+M)}} \\
\midrule

\multirow{7}{*}{BM+M} & De & 15.0 & 12.8 & 84.1 & 76.6 & \multirow{7}{*}{TR-1S+M} & De & 76.5 & 58.3 & \textbf{84.5} & 77.4 \\
& En & \textbf{100.0 }& 98.1 & \textbf{98.9} & 99.5 & & En & \textbf{100.0} & \textbf{98.8} & 98.8 & \textbf{99.7 }\\
& Es & 37.0 & 36.9 & 93.4 & 91.5 & & Es & 86.3 & 81.9 & \textbf{93.6} & 93.7 \\
& Fr & 47.4 & 40.0 & 89.8 & 86.8 & & Fr & 87.9 & 74.3 & 90.2 & 89.4 \\
& Nl & 9.3 & 10.0 & \textbf{93.1} & 92.5 & & Nl & 63.7 & 65.3 & 92.0 & 91.4 \\
& No & 14.7 & 10.9 & \textbf{91.8} & 90.4 & & No & 58.1 & 47.3 & 89.5 & 87.6 \\
& Ru & 15.4 & 10.8 & 82.2 & 60.8 & & Ru & 77.7 & 56.1 & 85.2 & 73.1 \\

\cmidrule(lr){1-6} \cmidrule(lr){7-12}

\multirow{7}{*}{TR-2S+M} & De & \textbf{95.1} & 66.5 & 84.1 & 77.1 & \multirow{7}{*}{TR-3S+M} & De & 94.1 & \textbf{69.6 }& 84.4 & \textbf{78.7} \\
& En & \textbf{100.0 }& 97.8 & 98.8 & 99.2 & & En & \textbf{100.0 }& 98.0 & \textbf{98.9} & 99.2 \\
& Es & \textbf{97.4} & \textbf{89.1} & 94.8 & \textbf{95.3} & & Es & 96.4 & 88.9 & 94.5 & 94.4 \\
& Fr & \textbf{97.7} & \textbf{83.1} & \textbf{91.8} & \textbf{90.1} & & Fr & 95.5 & 80.8 & 91.1 & 89.4 \\
& Nl & 95.6 & 88.8 & 92.7 & \textbf{93.1} & & Nl & \textbf{96.1} & \textbf{90.5} & 92.3 & 93.0 \\
& No & \textbf{97.6} & \textbf{72.7} & 90.1 & 85.7 & & No & 90.1 & 70.9 & 91.7 & \textbf{90.7} \\
& Ru & \textbf{96.7} & 60.8 & 85.2 & 74.0 & & Ru & 92.4 & \textbf{65.7} & \textbf{86.7} & \textbf{76.2} \\

\midrule
\multicolumn{12}{l}{\textbf{Crossmodal-3600 reference samples (Evaluating the LLM-as-a-judge)}} \\
\midrule

\multirow{7}{*}{references} & De & 100.0 & 83.3 & 85.4 & 91.7 & & & & & & \\
& En & 99.7 & 99.2 & 97.7 & 99.5 & & & & & & \\
& Es & 99.6 & 97.2& 95.2 & 99.0 & & & & & & \\
& Fr & 99.9 & 90.5 & 90.8 & 95.2 & & & & & & \\
& Nl & 99.4 & 95.8 & 94.5 & 98.7 & & & & & & \\
& No & 99.1 & 91.4 & 93.5 & 97.9 & & & & & & \\
& Ru & 99.9 & 76.0 & 79.8 & 83.3 & & & & & & \\

\bottomrule
\end{tabular}}
\caption{Model comparison showing GlotLID detection percentages for the target language (GL) and scores obtained using LLM-as-a-judge. LLM-L represents the target language detection, LLM-S indicates the numerical language consistency score, and LLM-B denotes the binary language consistency score. The GlotLID+LLM score is calculated as the product of the GlotLID score and LLM-B: $\text{GlotLID+LLM} = \text{GlotLID} \times \text{LLM-B}$. For each group except the LLM-as-a-judge evaluator evaluation, the highest scores are marked in bold.}
\label{tab:llm_judge_results}
\end{table*}

\subsection{Judging the Judge: Evaluation of the LLM-as-a-judge}

Even though we have implemented measures for robustness (such as using different generation configurations for the LLM-as-a-judge evaluators and requiring evaluators to provide summaries to justify their scores) we aim to further ensure the fairness of the provided scores by evaluating them against the reference captions in the dataset.

To assess the reliability of the language consistency evaluator model, we scored the reference captions from the evaluation dataset (crossmodal-3600) using the same evaluation process applied to the captions generated by our models. We have evaluated all the available references (up to 3 per image) and only used a single generation configuration (configuration B in Table \ref{tab:parameters_text_llm_as_a_judge}).

As shown in Table \ref{tab:llm_judge_results}, most of the obtained scores are above 90\%, demonstrating the effectiveness of the chosen model as an evaluator. However, some minor errors are present, which can be attributed to multilingual limitations. \texttt{Llama-3.1-8B-Instruct} officially supports seven languages in addition to English (French, German, Hindi, Italian, Portuguese, Spanish, and Thai). While this allows it to handle most European languages, it is expected that the model may occasionally struggle with languages outside its primary training set, leading to some misclassifications.

In terms of language consistency, we have discarded selected the binary score due to its higher scores. The evaluator classifies the samples as correct more than the 95\% of times in the majority of languages. The lower performance in German (91.7\%) can be attributed to the fact that it is a Germanic language that shares a large amount of words with English, what may induce classification errors. In the case of Russian (83.3\%), the lower performance may be explained by the limited support for languages using the Cyrillic alphabet in the LLM.

\vspace{-0.15cm}

\section{Caption Generation Prompt}
\label{app:cap_gen}
The Crossmodal-3600 dataset does not specify an explicit prompt for caption generation. However, in their work they provide instructions for generating captions, which we used as a guideline. Based on these instructions, we formulated a simplified captioning approach. The prompts used for generating image captions in our evaluation are presented in Figure~\ref{fig:cap_gen_prompt}. The same prompt was applied consistently across all selected languages.

\begin{figure*}[t]
    \centering
    \begin{tcolorbox}[
        colback=gray!8, 
        title={Caption Generation Prompts}, 
        coltitle=white, 
        colbacktitle=gray!127, 
    ]

    \textbf{[English]}
    
    Give me a brief summary of the following image, without too many details. The description should be general and have a maximum of 10 words. To do this, identify the most relevant object or person in the image, the main relationship between the highlighted objects, the most important activity represented, the most outstanding attributes of the main object or person, and the context in which the scene takes place. Then, synthesize everything into a single descriptive and concise sentence, without including additional text.\\ \\

    \textbf{[Spanish]} 
    
    Dame un breve resumen de la siguiente imagen, sin dar muchos detalles. La descripción debe ser general y tener un máximo de 10 palabras. Para ello, identifica el objeto o persona más relevante en la imagen, la relación principal entre los objetos destacados, la actividad más importante representada, los atributos más sobresalientes del objeto o persona principal y el contexto en el que ocurre la escena. Luego, sintetiza todo en una sola frase descriptiva y concisa, sin incluir texto adicional.\\ \\

    \textbf{[French]} 
    
    Donne-moi un bref résumé de l’image suivante, sans trop de détails. La description doit être générale et contenir un maximum de 10 mots. Pour cela, identifie l’objet ou la personne la plus importante dans l’image, la relation principale entre les objets mis en avant, l’activité la plus significative représentée, les attributs les plus marquants de l’objet ou de la personne principale et le contexte dans lequel la scène se déroule. Ensuite, synthétise tout en une seule phrase descriptive et concise, sans ajouter de texte supplémentaire.\\ \\

    \textbf{[German]} 
    
    Gib mir eine kurze Zusammenfassung des folgenden Bildes, ohne zu viele Details. Die Beschreibung sollte allgemein sein und maximal 10 Wörter umfassen. Identifiziere dazu das relevanteste Objekt oder die wichtigste Person im Bild, die Hauptbeziehung zwischen den hervorgehobenen Objekten, die wichtigste dargestellte Aktivität, die auffälligsten Merkmale des Hauptobjekts oder der Hauptperson und den Kontext, in dem die Szene stattfindet. Fasse dann alles in einem einzigen prägnanten und beschreibenden Satz zusammen, ohne zusätzlichen Text hinzuzufügen.\\ \\

    \textbf{[Italian]} 
    
    Dammi un breve riassunto della seguente immagine, senza troppi dettagli. La descrizione deve essere generale e avere un massimo di 10 parole. Per farlo, identifica l'oggetto o la persona più rilevante nell'immagine, la relazione principale tra gli oggetti evidenziati, l'attività più importante rappresentata, gli attributi più evidenti dell'oggetto o della persona principale e il contesto in cui si svolge la scena. Quindi, sintetizza tutto in un'unica frase descrittiva e concisa, senza includere testo aggiuntivo.

\end{tcolorbox}
    \caption*{\hfill \textit{(Continues on next page...)}}
  \label{fig:cap_gen_prompt}
    \vspace{-12pt}
\end{figure*}

\begin{figure*}[t]
    \centering
    \begin{tcolorbox}[
    ]

    \textbf{[Dutch]} 
    
    Geef me een korte samenvatting van de volgende afbeelding, zonder te veel details. De beschrijving moet algemeen zijn en maximaal 10 woorden bevatten. Identificeer hiervoor het meest relevante object of de belangrijkste persoon in de afbeelding, de hoofdrelatie tussen de uitgelichte objecten, de belangrijkste weergegeven activiteit, de meest opvallende kenmerken van het hoofdobject of de belangrijkste persoon en de context waarin de scène zich afspeelt. Vat vervolgens alles samen in één beschrijvende en beknopte zin, zonder extra tekst toe te voegen.\\ \\

    \textbf{[Portuguese]} 

    Dá-me um breve resumo da seguinte imagem, sem muitos detalhes. A descrição deve ser geral e ter um máximo de 10 palavras. Para isso, identifica o objeto ou a pessoa mais relevante na imagem, a principal relação entre os objetos destacados, a atividade mais importante representada, os atributos mais marcantes do objeto ou da pessoa principal e o contexto em que ocorre a cena. Depois, sintetiza tudo numa única frase descritiva e concisa, sem incluir texto adicional.\\ \\

    \textbf{[Russian]}

    \foreignlanguage{russian}{Дайте мне краткое описание следующего изображения без лишних деталей. Описание должно быть общим и содержать не более 10 слов. Для этого определите наиболее важный объект или человека на изображении, основную связь между выделенными объектами, главное изображенное действие, самые заметные характеристики главного объекта или человека и контекст, в котором происходит сцена. Затем объедините все это в одно описательное и краткое предложение, не добавляя лишнего текста.}
    
    \end{tcolorbox}
    \caption{Prompts used to evaluate via LLM-as-a-judge the language consistency of the caption provided by the model.}
  \label{fig:cap_gen_prompt}
    \vspace{-12pt}
\end{figure*}

\begin{table*}[ht!]
    \centering
    \footnotesize
    \begin{tabular}{llcccc}    \toprule
        & & \textbf{Stage-1} & \textbf{Stage-1.5} & \textbf{Stage-2} & \textbf{OneVision} \\
        \midrule
        \multicolumn{2}{l}{%
  \begin{tabular}{ll}
    \rotatebox[origin=c]{90}{\textbf{Vision}} & 
    \begin{tabular}{@{}l@{}}
      \textbf{Resolution} \\
      \# Tokens
    \end{tabular}
  \end{tabular}%
} &

    \begin{tabular}{@{}l@{}}
      384 \\
      729
    \end{tabular}
        & \begin{tabular}{@{}c@{}}
      AnyRes Max 5 \\
      Max \(729\times5\)
    \end{tabular}
        & \begin{tabular}{@{}c@{}}
      AnyRes Max 9\\
      Max \(729\times10\)
    \end{tabular}
        & \begin{tabular}{@{}c@{}}
      AnyRes Max 9\\
      Max \(729\times10\)
    \end{tabular} \\
        \midrule
        \multicolumn{2}{l}{%
  \begin{tabular}{ll}
    \rotatebox[origin=c]{90}{\textbf{Data}} & 
    \begin{tabular}{@{}l@{}}
      \textbf{Dataset} \\
      \# Vision Samples
    \end{tabular}
  \end{tabular}%
} &
    \begin{tabular}{@{}c@{}}
      Single-Image \\
      558K
    \end{tabular}
        & \begin{tabular}{@{}c@{}}
      Single-Image \\
      3.8M
    \end{tabular}
        & \begin{tabular}{@{}c@{}}
      Single-Image \\
      3.1M
    \end{tabular}
        & \begin{tabular}{@{}c@{}}
      Single/Multi-Image, Video \\
      1.6M
    \end{tabular} \\
        \midrule

        \multicolumn{2}{l}{%
  \begin{tabular}{cc}
    \rotatebox[origin=c]{90}{\textbf{Model}} & 
    \begin{tabular}{@{}l@{}}
      \textbf{Trainable} \\
      7.8B LLM
    \end{tabular}
  \end{tabular}%
} &
    \begin{tabular}{@{}c@{}}
      Projector \\
      20.0M
    \end{tabular}
        & \begin{tabular}{@{}c@{}}
      Full Model \\
      8.2B
    \end{tabular}
        & \begin{tabular}{@{}c@{}}
      Full Model \\
      8.2B
    \end{tabular}
        & \begin{tabular}{@{}c@{}}
      Full Model \\
      8.2B
    \end{tabular} \\
        \midrule

        \multicolumn{2}{l}{%
  \begin{tabular}{cc}
    \rotatebox[origin=c]{90}{\textbf{Training}} & 
    \begin{tabular}{@{}l@{}l@{}l@{}}
      \textbf{Batch Size} \\
      \textbf{LR: $\psi_{vision}$} \\
      \textbf{LR: \{$\theta_{proj}, \phi_{LLM}$\}} \\
      Epoch \\
      Warmup Ratio \\
      LR Scheduler \\
      Grad. Accum. \\
    \end{tabular}
  \end{tabular}%
} &
    \begin{tabular}{@{}c@{}}
      128 \\
      \(1\times10^{-3}\) \\
      \(1\times10^{-3}\) \\
      1 \\
      0.03 \\
      Cosine \\
      1 \\
    \end{tabular}
        & \begin{tabular}{@{}c@{}}
      64 \\
      \(2\times10^{-6}\) \\
      \(1\times10^{-5}\) \\
      1 \\
      0.03 \\
      Cosine \\
      2 \\
    \end{tabular}
        & \begin{tabular}{@{}c@{}}
      64 \\
      \(2\times10^{-6}\) \\
      \(1\times10^{-5}\) \\
      1 \\
      0.03 \\
      Cosine \\
      2 \\
    \end{tabular}
        & \begin{tabular}{@{}c@{}}
      64 \\
      \(2\times10^{-6}\) \\
      \(1\times10^{-5}\) \\
      1 \\
      0.03 \\
      Cosine \\
      2 \\
    \end{tabular} \\

        \bottomrule
    \end{tabular}
    \caption{Detailed configuration for each training stage of the LLaVA-OneVision model. For a detailed explanation of AnyRes Max, refer to \citep{llavaonevision}. Anyres Max 5: 384 \(\times \{2\times2, 1\times\{2,3\}, {2,3}\times1\}\). AnyRes Max 9: 384 \(\times\{ \{1\times1\}, ... , \{6\times6\}\}\).}
    \label{tab:onevision-stages}
\end{table*}

\section{Metrics Discussion}
\label{app:metrics_discussion}

In this study, we chose not to use teacher forcing loss or perplexity as evaluation metrics due to their inherent limitations in interpretability and comparative analysis across models.

We selected chrF++ as our primary evaluation metric rather than BLEU or ROUGE due to its suitability for multilingual assessments. BLEU relies heavily on exact n-gram matching, often penalizing legitimate linguistic variations common in multilingual contexts, while ROUGE primarily measures recall and is optimized for summarization tasks, making it suboptimal for assessing generative multilingual output quality. In contrast, chrF++ evaluates based on character-level n-gram overlaps, accommodating linguistic diversity and morphological richness across multiple languages, thus providing a more robust and linguistically sensitive assessment for multilingual visual language models.

Additionally, during the study, we employed VLM-as-a-judge to evaluate the quality of multilingual generations. However, we found that chrF++ effectively addressed the limitations related to multilingual performance evaluation inherent in other metrics, thereby serving as a comprehensive solution for our assessment needs.

\section{Training Hyperparameters}
\label{app:training-params}

The training hyperparameters used during the training of the models evaluated in this work are detailed in Table~\ref{tab:onevision-stages}.

\section{Alternative Merging Strategies}
\label{app:merging-tradeoffs}
To better understand the trade-offs involved in model merging, we conducted a series of additional experiments comparing different interpolation methods and weight ratios. In particular, we investigated:

\paragraph{Linear Interpolation (lerp)} This method interpolates model weights using the standard formula $w = (1-\alpha) w_1 + \alpha w_2$, where $w_1$ and $w_2$ are the weights of the visually instructed and backbone models, respectively, and $\alpha$ is the interpolation ratio.

\paragraph{Spherical Linear Interpolation (slerp)} Unlike \texttt{lerp}, slerp~\cite{shoemake1985animating} interpolates weights along a great arc on the hypersphere, preserving the norm and relative directionality. It is computed as:
\begin{equation}
\text{slerp}(w_1, w_2, \alpha) = \frac{\sin((1 - \alpha)\theta)}{\sin(\theta)} w_1 + \frac{\sin(\alpha \theta)}{\sin(\theta)} w_2
\end{equation}
where $\theta$ is the angle between the two weight vectors. This method can yield smoother transitions in weight space, especially when the models differ significantly.

We evaluated merged models using both interpolation methods under three weighting scenarios:
\begin{itemize}
    \item \textbf{50--50}, giving equal weight to the visually instructed and original backbone models.
    \item \textbf{75--25}, favoring the original backbone to preserve pretrained language capabilities.
    \item \textbf{25--75}, prioritizing the visually instructed model to reinforce vision-language alignment.
\end{itemize}

The results are presented in Table~\ref{tab:merging_tradeoffs}, showing both language fidelity and downstream task performance.

\begin{table*}[h]
\centering
\caption{
Merged model results: language fidelity (left) and task performance (right).
\texttt{lerp\_050} and \texttt{slerp\_050} correspond to 50--50 merges using linear and spherical interpolation, respectively; \texttt{slerp\_050} corresponds to our main model \texttt{TR-3S+M}. 
\texttt{lerp\_075} and \texttt{slerp\_075} are asymmetric 75--25 merges favoring the base text model. 
\texttt{lerp\_025} and \texttt{slerp\_025} invert this ratio to prioritize the visually instructed model.
}
\label{tab:merging_tradeoffs}
\small
\begin{tabular}{lrrrrr|r|rrrr|r}
\toprule
\textbf{Model} & \textbf{DE} & \textbf{ES} & \textbf{FR} & \textbf{NL} & \textbf{RU} & \textbf{Avg.} & \textbf{AI2D} & \textbf{MMMU} & \textbf{MMStar} & \textbf{RWQA} & \textbf{Avg.} \\
\midrule
\texttt{lerp\_075}  & 98.75 & 99.89 & 99.92 & 99.64 & 98.66 & 99.37 & 41.84 & 27.00 & 32.56 & 37.39 & 34.70 \\
\texttt{slerp\_075} & 99.08 & 99.86 & 99.89 & 99.64 & 99.02 & 99.50 & 42.16 & 27.33 & 32.66 & 36.99 & 34.79 \\
\texttt{lerp\_050}  & 92.83 & 96.42 & 95.83 & 94.78 & 91.11 & 94.19 & 57.03 & 34.44 & 42.06 & 52.03 & 46.39 \\
\texttt{slerp\_050} & 94.10 & 96.40 & 95.50 & 96.10 & 92.40 & 94.90 & 57.19 & 34.11 & 42.25 & 52.16 & 46.43 \\
\texttt{lerp\_025}  & 89.32 & 93.06 & 91.06 & 92.83 & 63.56 & 85.97 & 72.51 & 34.89 & 47.13 & 55.95 & 52.62 \\
\texttt{slerp\_025} & 90.11 & 93.53 & 90.81 & 94.44 & 66.30 & 87.44 & 72.38 & 34.89 & 47.27 & 56.08 & 52.66 \\

\bottomrule
\end{tabular}
\end{table*}

\paragraph{Discussion.}  
As shown, asymmetric merges favoring the backbone (e.g., \texttt{lerp\_075} and \texttt{slerp\_075}) achieve near-perfect language fidelity but show weaker performance in multimodal benchmarks. Conversely, merges favoring the visually instructed model (e.g., \texttt{lerp\_025} and \texttt{slerp\_025}) lead to substantially improved task performance, but at the cost of lower fidelity in certain languages. The \texttt{slerp\_050} model—corresponding to our main \texttt{TR-3S+M}—offers a more balanced trade-off.

Overall, no single merging configuration yields a clearly optimal trade-off. The best strategy depends on the intended use case: 75\% visually instructed weights are preferable when multilingual fidelity is critical, while 25\% weights better support general multimodal performance.



\section{Generation Examples}
\label{app:generation_examples}

Figures~\ref{fig:gen2}-\ref{fig:gen4} present examples generated using the TR-3S-0.05x model across various languages and diverse tasks.

\paragraph{Story Generation} In Figure \ref{fig:gen2}, the model is prompted to generate a story from an image. It accurately reads text within the image to identify characters and establish the setting, demonstrating its ability to craft diverse narratives consistently across different languages—even when the text is in English.

\paragraph{Image Description} Figure \ref{fig:gen1} presents a brief image description task. Although the image shows a salamander perched on a person’s hand, some language outputs mistakenly label it as an insect or a predator. Despite these inaccuracies, the descriptions remain largely appropriate.

\paragraph{OCR and Translation} Figure \ref{fig:gen3} showcases a task combining OCR with translation. The model extracts text from an image and then translates it into a target language. This two-step process: OCR followed by translation, highlights the model’s ability to merge visual analysis with its linguistic capabilities. Minor errors do occur, particularly in languages not extensively represented during training, resulting in slightly erroneous translations or defaulting to English.

\paragraph{Multi-Image Reasoning} Finally, Figure~\ref{fig:gen4} illustrates a multi-image scenario where the model must comprehend the content of several images and reason to provide an appropriate answer. This example further confirms the model’s effectiveness in real-world applications.

In general, these examples demonstrate how the VLMs instructed via our approach perform optimally across a wide range of tasks, especially for languages where text-only data was incorporated during the visual instruction process.

\begin{figure*}[h!] 
\centering 
\begin{tcolorbox}[ colback=gray!8, 
title={Story Generation}, 
coltitle=white, 
colbacktitle=gray!127, 
] 

 \begin{center}
        \includegraphics[width=0.3\textwidth]{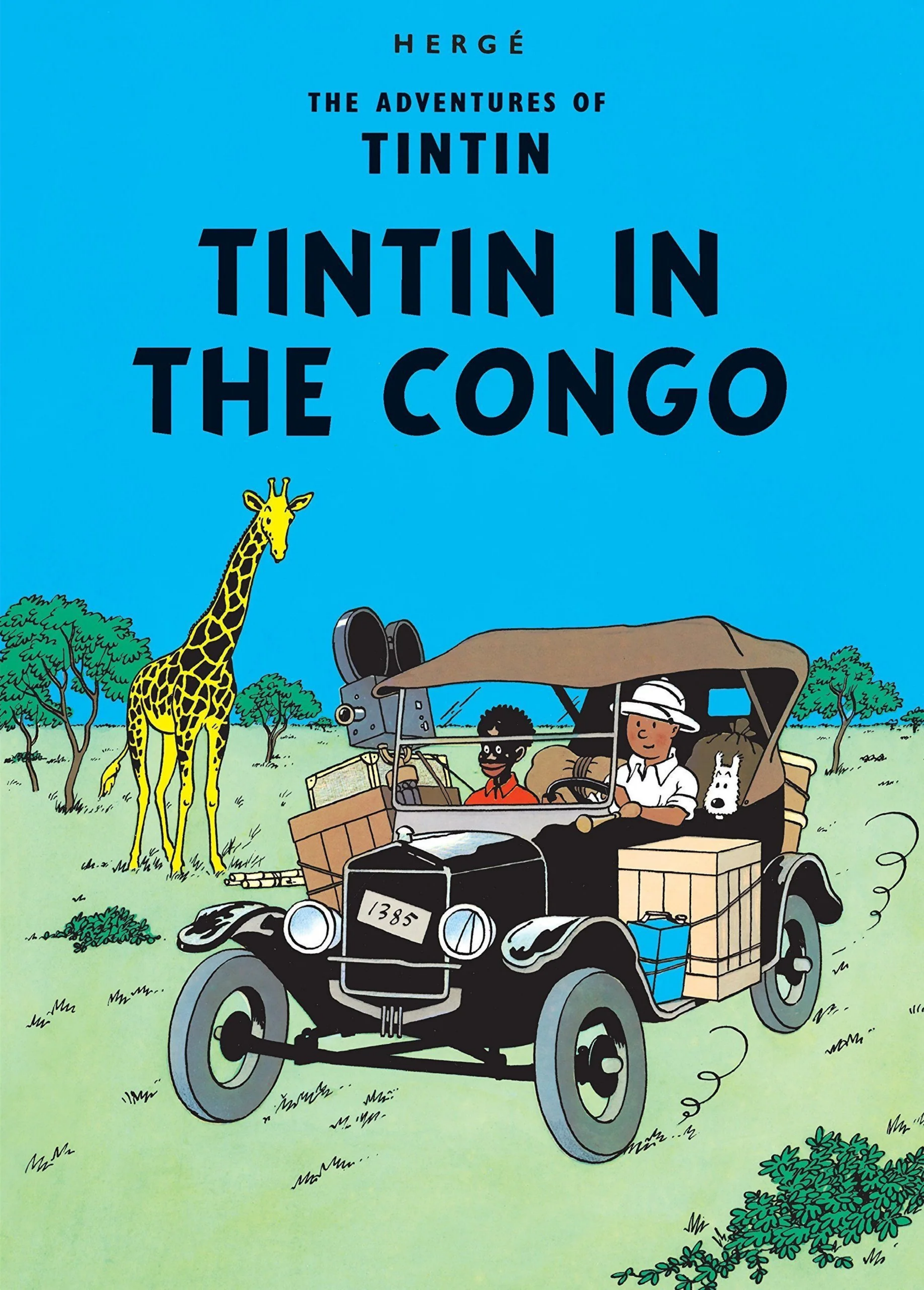} 
\end{center}

\textbf{[English]} \\ \textit{User:} Create a short story using the following image as source. \\ \textit{VLM:} Once upon a time, in the heart of the African savannah, there was a young boy named Tintin. Tintin was an adventurous soul, always eager to explore new places and meet new people. One day, he heard about a mysterious place called the Congo, a land of exotic animals and uncharted territories. Tintin decided to embark on an adventure to discover the secrets of the Congo.\\
As he traveled through the dense jungle, he encountered a variety of animals, including a majestic giraffe and a wise elephant. He also met a group of explorers who were searching for a hidden treasure. Tintin joined their team and together they set out on a thrilling journey through the jungle.\\
They faced many challenges and obstacles, but Tintin's curiosity and bravery never faltered. Finally, they reached the heart of the Congo and discovered a treasure—a collection of rare artifacts from different cultures, each with its own story to tell.\\
Tintin and his friends returned home with a newfound appreciation for the wonders of the world and the importance of exploration and discovery. And so, the legend of Tintin in the Congo began. \\ \\
\textbf{[Spanish]} \\
\textit{User:} Crea una historia corta usando la siguiente imagen como fuente. \\ \textit{VLM:} Había una vez un aventurero llamado Tintín que viajaba por el Congo en un coche antiguo. Lo acompañaban un guía y un perro, ambos llamados Milú. Tintín y sus amigos se encontraron con un grupo de gorilas y decidieron ayudarlos a escapar de los cazadores furtivos. Juntos, salvaron a los gorilas y regresaron a la ciudad, convirtiéndose en héroes en el Congo y continuando sus aventuras por el mundo. \\
{\textcolor{blue}{\small[English] \textit{VLM:} Once upon a time there was an adventurer named Tintin who traveled through the Congo in an antique car. He was accompanied by a guide and a dog, both named Snowy. Tintin and his friends came across a group of gorillas and decided to help them escape from poachers. Together, they saved the gorillas and returned to the city, becoming heroes in the Congo and continuing their adventures around the world.}}

\end{tcolorbox}
\caption*{\hfill \textit{(Continues on next page...)}}

\end{figure*}

\begin{figure*}[t!]
\begin{tcolorbox}[]

\textbf{[Catalan]} \\ \textit{User:} Crea una història curta utilitzant la següent imatge com a font. \\ \textit{VLM:} En Tintín i el seu gos Milú es troben en un safari a l'Àfrica. El cotxe d'en Tintín es trenca i han de demanar ajuda a un grup de nadius. Aquests reparen el cotxe i, a canvi, en Tintín i Milú els ensenyen una pel·lícula que és tot un èxit. Els nadius, encantats, els demanen que tornin l'any següent per mostrar-ne una altra. Finalment, en Tintín i Milú decideixen quedar-se a l'Àfrica i muntar un cinema per als habitants, començant així una nova vida plena d'amistat i intercanvi cultural. \\
{\textcolor{blue}{\small[English] \textit{VLM:} Tintin and his dog Snowy are on a safari in Africa. Tintin's car breaks down and they have to ask a group of natives for help. They repair the car and, in return, Tintin and Snowy show them a film that is a huge success. The natives, delighted, ask them to come back the following year to show another one. Finally, Tintin and Snowy decide to stay in Africa and set up a cinema for the inhabitants, thus beginning a new life full of friendship and cultural exchange.}}\\\\
\textbf{[French]} \\ \textit{User:} Créez une histoire courte en utilisant l'image suivante comme source. \\ \textit{VLM:} Il était une fois, dans la savane africaine, un petit garçon nommé Tintin, passionné d’aventures et rêvant d’explorer le continent. Un jour, il partit à la recherche de son idole, le célèbre explorateur Tintin, et se lança dans une aventure à travers la savane, rencontrant girafes, lions et éléphants. Finalement, il rencontra son idole dans le Congo, et ensemble, ils découvrirent de nombreux secrets et trésors, réalisant ainsi le rêve d'explorer l'Afrique. \\ 
{\textcolor{blue}{\small[English] \textit{VLM:} Once upon a time, in the African savannah, a little boy named Tintin had a passion for adventure and dreamed of exploring the continent. One day, he set off in search of his idol, the famous explorer Tintin, and embarked on an adventure across the savannah, encountering giraffes, lions and elephants. Finally, he met his idol in the Congo, and together they discovered many secrets and treasures, fulfilling the dream of exploring Africa.}}\\\\
\textbf{[German]} \\ \textit{User:} Erstelle eine kurze Geschichte, indem du das folgende Bild als Quelle verwendest. \\ \textit{VLM:} Tintin und sein Hund Milú begeben sich auf ein Abenteuer im Kongo. Während sie mit einem alten Auto durch die Savanne fahren, begegnen sie einem riesigen Giraffen, das sie fasziniert. Nachdem sie weitere Tiere wie Elefanten und Löwen getroffen haben, erleben sie eine spannende Reise, bei der sie schließlich einen Schatz entdecken und stolz auf ihre erlebten Abenteuer zurückblicken. \\ {\textcolor{blue}{\small[English] \textit{VLM:} Tintin and his dog Milú go on an adventure in the Congo. While driving through the savannah in an old car, they come across a giant giraffe that fascinates them. After meeting other animals such as elephants and lions, they experience an exciting journey during which they finally discover a treasure and look back proudly on their adventures.}}\\\\
\textbf{[Italian]} \\ \textit{User:} Crea una breve storia utilizzando la seguente immagine come fonte. \\ \textit{VLM:} Tintin e il suo cane Milú sono in un’avventura in Africa. Durante il viaggio, mentre guidano il loro veicolo, notano una grande giraffa che si avvicina e decidono di fermarsi per osservarla da vicino, vivendo un momento unico nel cuore della savana. \\
{\textcolor{blue}{\small[English] \textit{VLM:} Tintin and his dog Milú are on an adventure in Africa. On the way, while driving their vehicle, they notice a large giraffe approaching and decide to stop and observe it closely, experiencing a unique moment in the heart of the savannah.}}
 \end{tcolorbox} \caption*{\hfill \textit{(Continues on next page...)}}
\end{figure*}

\begin{figure*}[t!]
\begin{tcolorbox}[]

\textbf{[Dutch]} \\ \textit{User:} Maak een kort verhaal met behulp van de volgende afbeelding als bron. \\ \textit{VLM:} Tintin in de Congo is een stripverhaal van Hergé waarin Tintin als journalist naar het Congo reist, vergezeld door zijn beste vriend Milú, een geredde chimpansee. Ze beleven tal van avonturen in de jungle, ontdekken de schoonheid van de natuur en maken kennis met de lokale bevolking, wat leidt tot een onvergetelijke reis vol humor en spanning. \\
{\textcolor{blue}{\small[English] \textit{VLM:} Tintin in the Congo is a comic strip by Hergé in which Tintin travels to the Congo as a journalist, accompanied by his best friend Milú, a rescued chimpanzee. They have numerous adventures in the jungle, discover the beauty of nature and meet the local people, leading to an unforgettable journey full of humor and excitement.}}\\\\
\textbf{[Portuguese]} \\ \textit{User:} Crie uma curta história usando a seguinte imagem como fonte. \\ \textit{VLM:} Era uma vez um explorador chamado Tintin, que se encontrava numa aventura no Congo. Ao volante de um carro vintage, acompanhado pelo seu amigo e animal de estimação, Milú, Tintin estava em busca de um tesouro escondido. No caminho, tiveram de improvisar para contornar um enorme elefante que bloqueava a passagem, mas conseguiram seguir em frente e, no final, encontraram o tesouro que buscavam. \\ 
{\textcolor{blue}{\small[English] \textit{VLM:} Once upon a time there was an explorer called Tintin, who was on an adventure in the Congo. At the wheel of a vintage car, accompanied by his friend and pet, Milú, Tintin was looking for hidden treasure. I don't walk, you had to improvise to get around a huge elephant that was blocking the passage, but we managed to keep going and, finally, we found the treasure we were looking for.}}\\\\
\textbf{[Galician]} \\ \textit{User:} Crea unha historia curta utilizando a seguinte imaxe como fonte. \\ \textit{VLM:} Había unha vez un explorador chamado Tintin no Congo. Durante a súa viaxe, atopouse con animais exóticos como un gorila, un elefante e un león, pero o que máis chamou a súa atención foi un gran xaguar no medio do bosque. Cansado, o xaguar foi axudado por Tintin a chegar a un río para beber, e xuntos descubriron unha cova misteriosa con un tesouro oculto que decidiron compartir.\\
{\textcolor{blue}{\small[English] \textit{VLM:} Once upon a time there was an explorer named Tintin no Congo. During his journey, he encountered exotic animals such as a gorilla, an elephant and a lion, but what caught his attention the most was a large jaguar in the middle of the forest. Tired, the jaguar was helped by Tintin to reach a river to drink, and together they discovered a mysterious cave with a hidden treasure that they decided to share.}}\\\\
\textbf{[Polish]} \\
\textit{User:} \foreignlanguage{polish}{Stwórz krótką historię, używając poniższego obrazu jako źródła.}\\
\textit{VLM:} \foreignlanguage{polish}{Tytuł: ``Tintin w Kong'' Autor: Hergé Data wydania: 1931}\\
{\textcolor{blue}{\small[English] \textit{VLM:} Title: “Tintin in the Congo” Author: Hergé Publication date: 1931}} \\\\
\textbf{[Swedish]} \\ 
\textit{User:}  Skapa en kort berättelse med hjälp av följande bild som källa.\\
\textit{VLM:} Tintin och Milú är på en safari i Afrika. De sitter i en old-fashioned bil och ser på en giraffe som är i vägen. Tintin och Milú tar sig över till den andra sidan av vägen och ser på en elefant som är i vägen. De tar sig över till den andra sidan av vägen och ser på en elefant som är i vägen.  \\
{{\textcolor{blue}{\small[English] \textit{VLM:} Tintin and Milú are on a safari in Africa. They sit in an old-fashioned car and look at a giraffe that is in the road. Tintin and Milú cross to the other side of the road and look at an elephant in the road. They cross to the other side of the road and look at an elephant in the road.}}}
\end{tcolorbox} \caption*{\hfill \textit{(Continues on next page...)}}
\end{figure*}

\begin{figure*}[t!]
\begin{tcolorbox}[]

\textbf{[Russian]} \\ 
\textit{User:} \foreignlanguage{russian}{Создайте короткий рассказ, используя следующее изображение в качестве источника. }\\
\textit{VLM:} \foreignlanguage{russian}{Однажды, в далекой Африке, в самом сердце Конго, в 1920-х годах, произошло уникальное приключение, которое стало известно как "Тинтин в Конго".
Тинтин, знаменитый репортер и путешественник, был отправлен в Африку, чтобы рассказать о жизни и культуре этого загадочного континента. Он и его верные спутники, Микель и Нельсон, сели в свой старенький, но надежный автомобиль и отправились в путешествие.
Они пробирались через дикие джунгли, где их окружали дикие животные и таинственные племена. Тинтин и его команда были не только reporters, но и смелые путешественники, которые хотели узнать больше о мире, который был для них загадкой.
Они встретили множество людей, которые были для них новыми и интересными, и узнали о их культуре и традициях. Они также узнали о том, что Конго был не только домом для диких животных, но и для людей, которые жили в гармонии с природой.
В конце концов, Тинтин и его команда достигли своей цели и рассказали о своем путешествии в Европе, где их история вызвала большой интерес.
"Тинтин в Конго" стал одним из самых известных и любимых приключений Тинтина, и он всегда будет помнить об этом} \\ 
{\textcolor{blue}{\small[English] \textit{VLM:} Once upon a time, in faraway Africa, in the heart of the Congo, in the 1920s, a unique adventure took place that became known as “Tintin in the Congo.”
Tintin, a famous reporter and traveler, was sent to Africa to report on the life and culture of this mysterious continent. He and his faithful companions, Mikel and Nelson, got into their old but reliable car and set off on their journey.
They made their way through a wild jungle where they were surrounded by wild animals and mysterious tribes. Tintin and his team were not only reporters, but also brave travelers who wanted to learn more about a world that was a mystery to them.
They met many people who were new and interesting to them and learned about their culture and traditions. They also learned that the Congo was not only home to wild animals, but also to people who lived in harmony with nature.
In the end, Tintin and his team reached their goal and told their journey in Europe, where their story generated a lot of interest.
“Tintin in the Congo” became one of Tintin's most famous and beloved adventures, and he will always be remembered for it}} \\\\
\textbf{[Norwegian]} \\ 
\textit{User:}  Lag en kort historie ved å bruke følgende bilde som kilde. \\
\textit{VLM:} Tintin og Milú er på en safari i Afrika. De er i en bil og ser på en giraffe. \\
{\textcolor{blue}{\small[English] \textit{VLM:} Tintin and Milú are on a safari in Africa. They are in a car looking at a giraffe.}}\\ \\ 
\textbf{[Hungarian]} \\ 
\textit{User:} Írj egy rövid történetet a következő kép felhasználásával forrásként. \\
\textit{VLM:} Tintin a Kongóba indul, hogy megmentse a királyt és a királynőt, akiket a gorilla rabolt el. A királynő és a királyt a gorilla foglyul ejtette, és a Kongóba menekült. Tintin és a barátai elindulnak, hogy megmentsék a királyt és a királynőt. \\
{{\textcolor{blue}{\small[English] \textit{VLM:} Tintin goes to the Congo to rescue the king and queen, who have been kidnapped by the gorilla. The king and queen have been captured by the gorilla and have fled to the Congo. Tintin and his friends set out to rescue the king and queen.}}}
 \end{tcolorbox} \caption{Multilingual generation examples with Text Regularization and merged with the original backbone LLM.}
\label{fig:gen2}\end{figure*}

\begin{figure*}[h!]
    \centering
    \begin{tcolorbox}[
        colback=gray!8, 
        title={Image Description}, 
        coltitle=white, 
        colbacktitle=gray!127, 
    ]
    \begin{center}
        \includegraphics[width=0.3\textwidth]{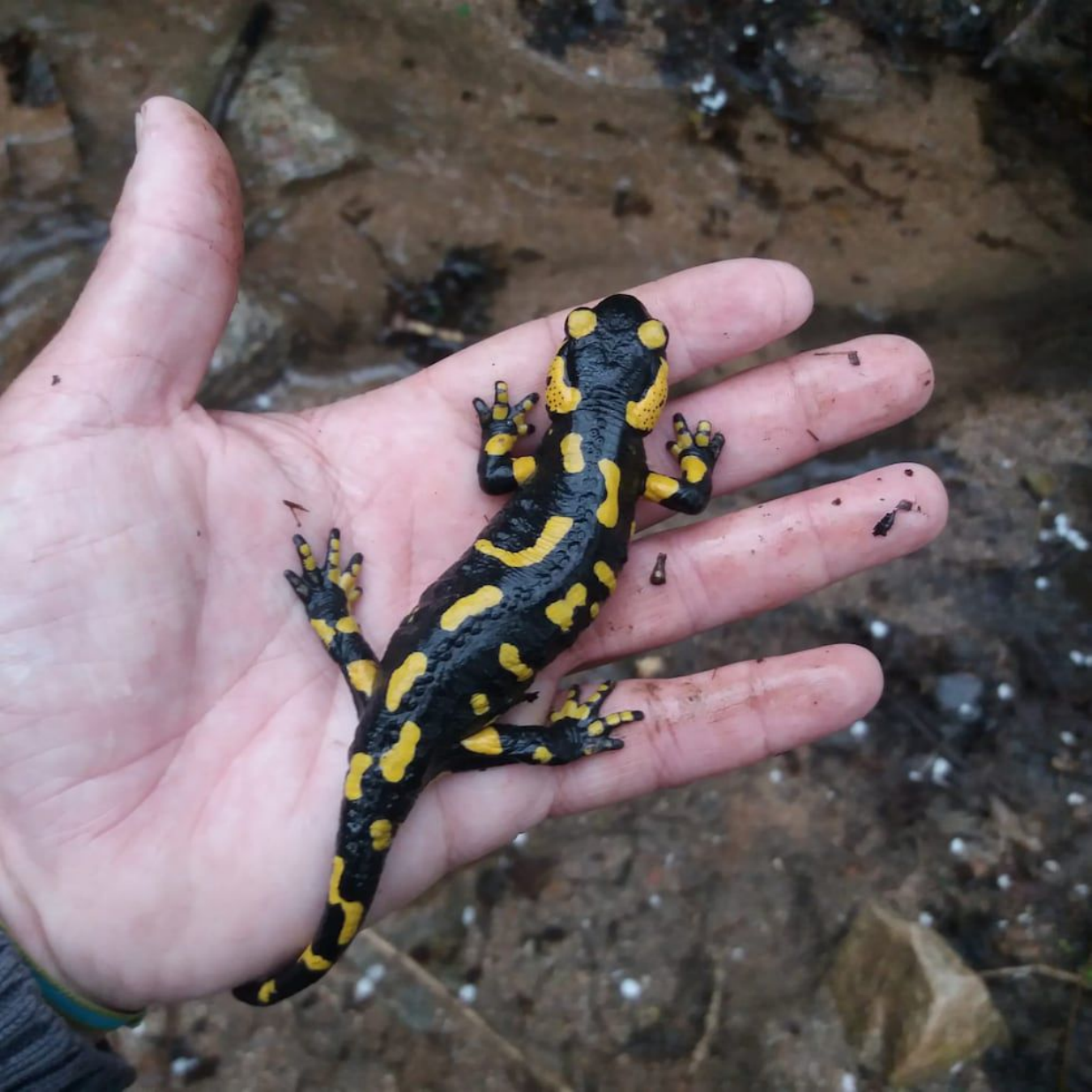} 
    \end{center}

    \textbf{[English]} \\
    \textit{User:} Briefly describe the following image, without many details and in a few words. \\
    \textit{VLM:} A person's hand holding a black and yellow striped salamander. \\\\
    \textbf{[Spanish]} \\
    \textit{User:} Describe brevemente la siguiente imagen, sin muchos detalles y en pocas palabras. \\
    \textit{VLM:} Una mano sosteniendo a un insecto, específicamente un insecto de color negro y amarillo. \\
    {\textcolor{blue}{\small[English] \textit{VLM:}A hand holding an insect, specifically a black and yellow insect.}}\\\\
    \textbf{[Catalan]}  \\
    \textit{User:} Descriu breument la següent imatge, sense molts detalls i en poques paraules. \\
    \textit{VLM:} A la mà d'una persona hi ha una salamandra de color negre i groc. \\
    {\textcolor{blue}{\small[English] \textit{VLM:} In a person's hand there is a black and yellow salamander.}}\\\\
    \textbf{[French]} \\
    \textit{User:} Décrivez brièvement l'image suivante, sans trop de détails et en quelques mots. \\
    \textit{VLM:} Un petit animal noir et blanc est dans la main d'une personne. \\
    {\textcolor{blue}{\small[English] \textit{VLM:} A small black and white animal is in a person's hand.}}\\\\
    \textbf{[German]} \\
    \textit{User:} Beschreiben Sie das folgende Bild kurz, ohne viele Details und in wenigen Worten. \\
    \textit{VLM:} In der Hand einer Person liegt ein Frillengecko. Der Gecko hat eine schwarze und gelbe markierte Haut.  Die Person hat einen Arm mit einem Armband. \\
    {\textcolor{blue}{\small[English] \textit{VLM:} A frilled gecko lies in the hand of a person. The gecko has black and yellow marked skin.  The person has an arm with a bracelet.}}\\\\
    \textbf{[Italian]} \\
    \textit{User:} Descrivi brevemente la seguente immagine, senza molti dettagli e in poche parole. \\
    \textit{VLM:} Un insetto in una mano. \\
    {\textcolor{blue}{\small[English] \textit{VLM:} An insect in a hand.}}\\\\
    \textbf{[Dutch]} \\
    \textit{User:} Beschrijf kort de volgende afbeelding, zonder veel details en in een paar woorden. \\
    \textit{VLM:} De hand van een mens met een salamander in het palm. \\
    {\textcolor{blue}{\small[English] \textit{VLM:} The hand of a human with a salamander in the palm. }}
    \end{tcolorbox}
    \caption*{\hfill\textit{(Continues on next page...)}}
\end{figure*}

\begin{figure*}[h!]
    \centering
    \begin{tcolorbox}[
        colback=gray!8, 
    ]
    \textbf{[Portuguese]} \\
    \textit{User:} Descreva resumidamente a imagem seguinte, sem muitos pormenores e em poucas palavras. \\
    \textit{VLM:} Uma pessoa está a segurar um sapo em sua mão. O sapo é de cor preta e amarela. O sapo está vivo. \\
    {\textcolor{blue}{\small[English] \textit{VLM:} A person is holding a frog in his hand. The frog is black and yellow in color. The frog is alive.}}\\\\
    \textbf{[Galician]} \\
    \textit{User:} Describe brevemente a seguinte imaxe, sen moitos detalles e en poucas palabras. \\
    \textit{VLM:} A man sostén unha larva de salamandra. A larva é de cor negra e amarela. A man que sostén a larva é branca. \\
    {\textcolor{blue}{\small[English] \textit{VLM:} A man holds a yellow and black lobster nail.}}\\\\
    \textbf{[Russian]} \\ 
    \textit{User:} \foreignlanguage{russian}{Кратко опишите следующее изображение, без подробностей и в нескольких словах.} \\
    \textit{VLM:} \foreignlanguage{russian}{В руке человека, который держит змею.}\\
    {\textcolor{blue}{\small[English] \textit{VLM:} In the hand of a man holding a snake.}}\\\\
    \textbf{[Polish]} \\
    \textit{User:} \foreignlanguage{polish}{Opisz krótko poniższy obraz, bez wielu szczegółów i w kilku słowach.} \\
    \textit{VLM:} \foreignlanguage{polish}{Oto krótki opis tego obrazu:``Czarna i żółta salamandra w dłoni.'' To jest fotografia salamandry, która jest trzymana w ręce. Salamandra ma czarne i żółte plamy na swym ciele.}\\
    {\textcolor{blue}{\small[English] \textit{VLM:} Here is a short description of this image: 'A black and yellow salamander in hand.' This is a photograph of a salamander that is being held in hand. The salamander has black and yellow spots on its body.}}\\\\
    \textbf{[Swedish]} \\
    \textit{User:} \foreignlanguage{swedish}{Beskriv kort nedanstående bild, utan många detaljer och med några få ord.} \\
    \textit{VLM:} \foreignlanguage{swedish}{En hand med en salamander i den.}\\
    {\textcolor{blue}{\small[English] \textit{VLM:} A hand with a salamander in it.}}\\\\
    \textbf{[Norwegian]} \\
    \textit{User:} \foreignlanguage{norsk}{Beskriv kort det følgende bildet, uten mange detaljer og med noen få ord.} \\
    \textit{VLM:} \foreignlanguage{norsk}{En \underline{hand holding} en salamander.}\\
    {\textcolor{blue}{\small[English] \textit{VLM:} A hand holding a salamander.}}\\\\
    \textbf{[Hungarian]} \\
    \textit{User:} \foreignlanguage{hungarian}{Röviden írja le a következő képet, sok részlet nélkül, néhány szóban.} \\
    \textit{VLM:} \foreignlanguage{hungarian}{A képen egy ember tartja a kezében egy fekete és sárga ragadozó.}\\
    {\textcolor{blue}{\small[English] \textit{VLM:} In the picture, a person holds in their hand a black and yellow predator.}}    
    \end{tcolorbox}
    \caption{Multilingual generation examples with the model trained with Text Regularization and merged with the original backbone LLM. Code switched words are underlined. 
    }
    \label{fig:gen1}
\end{figure*}

\begin{figure*}[h!] \centering \begin{tcolorbox}[ colback=gray!8, 
title={OCR \& Translation}, 
fonttitle=\bfseries, 
coltitle=white, 
colbacktitle=gray!127, 
] 
\begin{center} \includegraphics[width=0.3\textwidth]{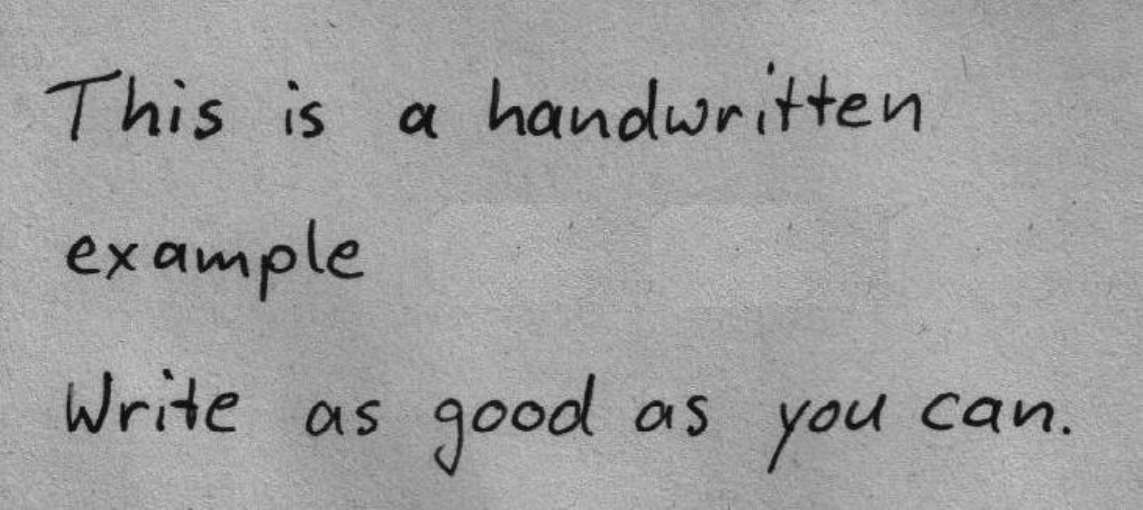} 
\end{center}
\textbf{[English]} \\
\textit{User:} Read the text that appears in the image and translate it into English. \\
\textit{VLM:} This is a handwritten example. Write as good as you can. \\ \\
\textbf{[Spanish]} \\
\textit{User:} Read the text that appears in the image and translate it into Spanish. \\
\textit{VLM:} Este es un ejemplo de escritura a mano. Escribe tan bien como puedas. \\
{\textcolor{blue}{\small[English] \textit{VLM:} This is a handwritten example. Write as good as you can.}}\\\\
\textbf{[Catalan]} \\
\textit{User:} Read the text that appears in the image and translate it into Catalan. \\
\textit{VLM:} Aquest és un exemple de text escrit a mà. Escriu tan bé com puguis. \\
{\textcolor{blue}{\small[English] \textit{VLM:} This is a handwritten example. Write as good as you can.}}\\\\
\textbf{[French]} \\
\textit{User:} Read the text that appears in the image and translate it into French. \\
\textit{VLM:} C'est un exemple de lettre écrite à la main. Ecrivez comme vous pouvez. \\
{\textcolor{blue}{\small[English] \textit{VLM:} This is an example of a handwritten letter. Write as you can.}}\\\\
\textbf{[German]} \\
\textit{User:} Read the text that appears in the image and translate it into German. \\
\textit{VLM:} Dies ist ein Beispiel für einen handgeschriebenen Text. Schreiben Sie so gut wie Sie können. \\
{\textcolor{blue}{\small[English] \textit{VLM:} This is a handwritten example. Write as good as you can.}}\\\\
\textbf{[Italian]} \\
\textit{User:} Read the text that appears in the image and translate it into Italian. \\
\textit{VLM:} Questo è un esempio di scrittura a mano. Scrivete bene come voi potete. \\
{\textcolor{blue}{\small[English] \textit{VLM:} This is a handwritten example. Write as good as you can.}}\\\\
\textbf{[Dutch]} \\
\textit{User:} Read the text that appears in the image and translate it into Dutch. \\
\textit{VLM:} Dit is een handgeschreven voorbeeld. Schrijf zo goed als je kunt. \\
{\textcolor{blue}{\small[English] \textit{VLM:} This is a handwritten example. Write as good as you can.}}\\\\
\textbf{[Portuguese]} \\
\textit{User:} Read the text that appears in the image and translate it into Portuguese. \\
\textit{VLM:} Este é um exemplo de um texto escrito à mão. Escreva como bom como você pode. \\
{\textcolor{blue}{\small[English] \textit{VLM:} This is a handwritten example. Write as good as you can.}}\\\\
\textbf{[Galician]} \\
\textit{User:} Read the text that appears in the image and translate it into Galician. \\
\textit{VLM:} Isto é un exemplo de escrita a man. Escribe ben como se pode.\\
{\textcolor{blue}{\small[English] \textit{VLM:} This is a handwritten example. Write as good as you can.}}
\end{tcolorbox} \caption*{\hfill \textit{(Continues on next page...)}}
\end{figure*}

\begin{figure*}[h!]
    \centering
    \begin{tcolorbox}[]
    \textbf{[Russian]} \\ 
    \textit{User:} Read the text that appears in the image and translate it into Russian. \\
    \textit{VLM:} \foreignlanguage{russian}{Это пример написанного от руки текста. Напишите как можно лучше, чем вы можете.}\\
    {\textcolor{blue}{\small[English] \textit{VLM:} This is an example of a handwritten text. Write as best as you can.}}\\\\
    \textbf{[Polish]} \\
    \textit{User:} Read the text that appears in the image and translate it into Polish. \\
    \textit{VLM:} \foreignlanguage{polish}{Oto tekst w języku polskim: ``To jest przykładowy handwriting example. Napisz tak, jak potrafisz'' }\\
    {\textcolor{blue}{\small[English] \textit{VLM:} Here is the text in Polish: ``This is a handwriting example. Write as much as you can'' }}\\\\
    \textbf{[Swedish]} \\
    \textit{User:} Read the text that appears in the image and translate it into Swedish. \\
    \textit{VLM:} \underline{This is a handwritten example. Write as good as you can.}\\
    {\textcolor{blue}{\small[English] \textit{VLM:} This is a handwritten example. Write as good as you can.}}\\\\
    \textbf{[Norwegian]} \\
    \textit{User:} Read the text that appears in the image and translate it into Norwegian. \\
    \textit{VLM:} \underline{This is a handwritten example. Write as good as you can.}\\
    {\textcolor{blue}{\small[English] \textit{VLM:} This is a handwritten example. Write as good as you can.}}\\\\
    \textbf{[Hungarian]} \\
    \textit{User:} Read the text that appears in the image and translate it into Hungarian. \\
    \textit{VLM:} \foreignlanguage{hungarian}{Ez egy kézzel írott példa. Írjál úgy, mint te tudsz.}\\
    {\textcolor{blue}{\small[English] \textit{VLM:} This is a handwritten example. Write the way you know how.}}

\end{tcolorbox}
\caption{Multilingual generation examples with the model trained with Text Regularization and merged with the original backbone LLM. Words in English whenever it is not the target are underlined.}
\label{fig:gen3}
\end{figure*}

\begin{figure*}[h!] \centering \begin{tcolorbox}[ colback=gray!8, 
title={Multi-Image Reasoning}, 
fonttitle=\bfseries, 
coltitle=white, 
colbacktitle=gray!127, 
] \begin{center} \includegraphics[width=0.3\textwidth]{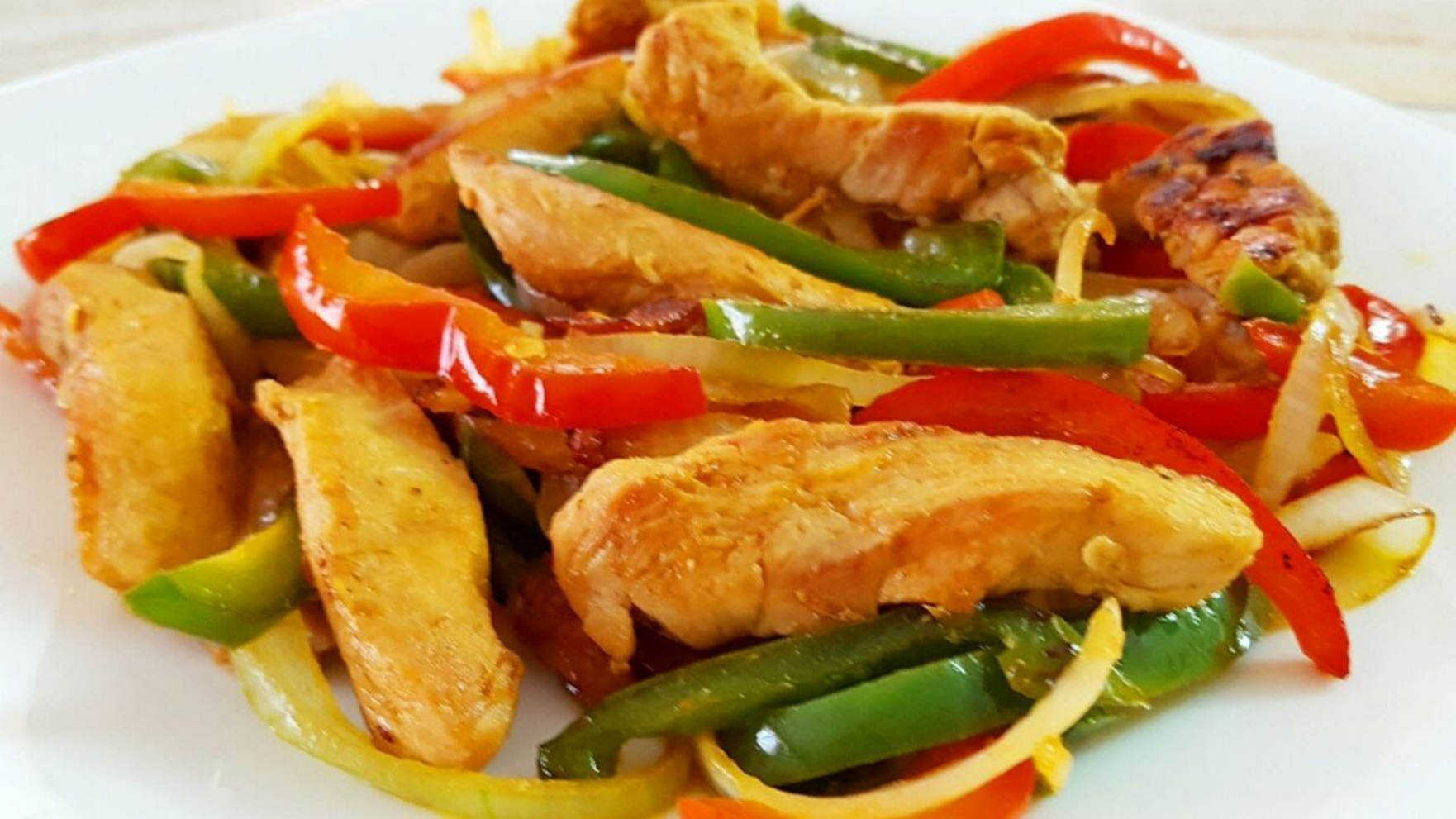} 
\includegraphics[width=0.3\textwidth]{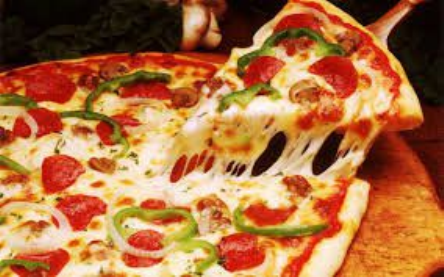} 
\end{center}

\textbf{[English]} \\
\textit{User:} Which dish from the ones shown in these images should I eat if I am lactose intolerant? \\
\textit{VLM:} If you are lactose intolerant, you should eat the dish on the left, which is a stir fry of chicken and vegetables. This dish does not contain any dairy products, making it a good option for those who are lactose intolerant. \\ \\
\textbf{[Spanish]} \\
\textit{User:} ¿Qué plato de los que aparecen en estas imágenes debería de comer si soy intolerante a la lactosa? \\
\textit{VLM:} Si eres intolerante a la lactosa, no deberías comer el plato de pizza. \\
{\textcolor{blue}{\small[English] \textit{VLM:} If you are lactose intolerant, you should not eat the pizza dish.}}\\\\
\textbf{[Catalan]} \\
\textit{User:} Quin plat de les imatges donades hauria de menjar si sóc intolerant a la lactosa? \\
\textit{VLM:} Si ets intolerant a la lactosa, hauries de menjar el plat de la primera imatge, ja que no conté productes lactis. \\
{\textcolor{blue}{\small[English] \textit{VLM:} If you are lactose intolerant, you should eat the dish in the first image, as it does not contain dairy products.}}\\\\
\textbf{[French]} \\
\textit{User:} Quel plat parmi ceux qui apparaissent dans ces images devrais-je manger si je suis intolérant au lactose ? \\
\textit{VLM:} Si vous êtes intolérant au lactose, je vous conseillerais de manger le plat de poulet grillé et de légumes. \\
{\textcolor{blue}{\small[English] \textit{VLM:} If you're lactose intolerant, I'd recommend eating the grilled chicken and vegetable dish.}}\\\\
\textbf{[Italian]} \\
\textit{User:} Quale piatto tra quelli mostrati in queste immagini dovrei mangiare se sono intollerante al lattosio? \\
\textit{VLM:} Se sei intollerante al lattosio, dovresti mangiare il primo piatto, il stir fry di pollo e verdura. \\
{\textcolor{blue}{\small[English] \textit{VLM:} If you are lactose intolerant, you should eat the first course, chicken and vegetable stir fry.}}
\end{tcolorbox}
\caption*{\hfill \textit{(Continues on next page...)}}

\end{figure*}

\begin{figure*}[t!]
\begin{tcolorbox}[]
\textbf{[German]} \\
\textit{User:} Welches Gericht aus den in diesen Bildern gezeigten sollte ich essen, wenn ich laktoseintolerant bin? \\
\textit{VLM:} Wenn Sie laktoseintolerant sind, sollten Sie das erste Bild, das Chicken and Vegetables, essen. \\ 
{\textcolor{blue}{\small[English] \textit{VLM:} If you are lactose intolerant, you should eat the first picture, the Chicken and Vegetables.}}\\\\
\textbf{[Dutch]} \\
\textit{User:} Welk gerecht van de getoonde gerechten in deze afbeeldingen zou ik moeten eten als ik lactose-intolerant ben? \\
\textit{VLM:} Als je lactose-intolerant bent, zou je het gerecht in de eerste afbeelding moeten eten. Het is een salade met een dressing die geen lactose bevat. De salade is gemaakt van grilled chicken, rode en groene peppers, en is niet gebakken of gefrituurd, wat betekent dat het geen lactose bevat. De salade is dus een lactose-vrije optie. \\
{\textcolor{blue}{\small[English] \textit{VLM:} If you are lactose intolerant, you should eat the dish in the first picture. It is a salad with a dressing that does not contain lactose. The salad is made with grilled chicken, red and green peppers, and is not baked or fried, which means that it does not contain lactose. The salad is therefore a lactose-free option.}}\\\\
\textbf{[Portuguese]} \\
\textit{User:} Qual prato entre os mostrados nestas imagens eu deveria comer se for intolerante à lactose? \\
\textit{VLM:} Se você é intolerante à lactose, o prato que você deve evitar é o pizza. A pizza é feita com queijo, que é uma fonte de lactose, e pode causar sintomas como dor abdominal, diarreia, náuseas e vômitos. Se você tem intolerância à lactose, é importante evitar alimentos que contenham lactose para evitar esses sintomas. \\
{\textcolor{blue}{\small[English] \textit{VLM:} If you are lactose intolerant, the food you should avoid is pizza. Pizza is made with cheese, which is a source of lactose, and can cause symptoms such as abdominal pain, diarrhea, nausea, and vomiting. If you are lactose intolerant, it is important to avoid foods that contain lactose to avoid these symptoms.}}\\\\
\textbf{[Galician]} \\
\textit{User:} Que prato das imaxes dadas debería comer se son intolerante á lactosa? \\
\textit{VLM:} Se é intolerante á lactosa, debería comer o prato da primeira imaxe, que é unha ensalada de pollo e verduras. \\
{\textcolor{blue}{\small[English] \textit{VLM:} If you are lactose intolerant, you should eat the dish in the first picture, which is a chicken and salad.}}\\\\
\textbf{[Polish]} \\
\textit{User:} \foreignlanguage{polish}{Jakie danie z pokazanych na tych obrazkach powinienem zjeść, jeśli jestem nietolerancyjny na laktozę?} \\
\textit{VLM:} \foreignlanguage{polish}{Jeżeli jesteś nietolerancki na laktozę, to nie powinieneś jeść pizzy. Pizza jest gotowa z mozzarella, a mozzarella jest produkowana z mleka. Jeżeli wolisz jeść potrawy bez laktozy, to możesz zjeść kurczaka z warzywami. Kurczak jest bezlaktozny, a warzywa są bezlaktozne. } \\
{\textcolor{blue}{\small[English] \textit{VLM:} If you are lactose intolerant, you should not eat pizza. Pizza is made with mozzarella, and mozzarella is made from milk. If you prefer to eat lactose-free food, you can eat chicken with vegetables. Chicken is lactose-free, and vegetables are lactose-free.}}
\end{tcolorbox}
\caption*{\hfill\textit{(Continues on next page...)}}
\label{fig:gen4}
\end{figure*}

\begin{figure*}[t!]
\begin{tcolorbox}[]
\textbf{[Russian]} \\
\textit{User:} \foreignlanguage{russian}{Какое блюдо из показанных на этих изображениях мне следует съесть, если я непереносим к лактозе?} \\
\textit{VLM:} \foreignlanguage{russian}{Если вы не переносите лактозу, то вам следует выбрать блюдо, которое не содержит молочных продуктов. На изображении 1 показаны кусочки курицы с овощами, которые не содержат лактозы. На изображении 2 показана пицца с сыром и другими toppings, которые могут содержать лактозу.
Однако, если вы хотите попробовать что-то, что не содержит лактозы, то вам следует обратиться к другим блюдам, которые не содержат молочные продукты.} \\
{\textcolor{blue}{\small[English] \textit{VLM:} If you are lactose intolerant, you should choose a dish that does not contain dairy products. Picture 1 shows chicken pieces with vegetables, which do not contain lactose. Picture 2 shows a pizza with cheese and other toppings, which may contain lactose. However, if you want to try something that does not contain lactose, you should turn to other dishes that do not contain dairy products.}}\\\\
\textbf{[Swedish]} \\
\textit{User:} \foreignlanguage{swedish}{Vilken rätt av de som visas i dessa bilder bör jag äta om jag är laktosintolerant?} \\
\textit{VLM:} \foreignlanguage{swedish}{Om du är laktosintolerant, skulle jag rekommendera att du äter den första bilden, som är en stir-fry med kyckling, redbetor, och pepper. Det har ingen laktos i sig och passar bra för dem som är laktosintoleranta.} \\
{\textcolor{blue}{\small[English] \textit{VLM:} If you are lactose intolerant, I would recommend eating the first picture, which is a stir-fry with chicken, beetroot, and pepper. It has no lactose in it and is suitable for those who are lactose intolerant. }}\\\\
\textbf{[Norwegian]} \\
\textit{User:} \foreignlanguage{norsk}{Hvilket måltid av de som vises på disse bildene bør jeg spise hvis jeg er laktoseintolerant?} \\
\textit{VLM:} \foreignlanguage{norsk}{For en person som er laktoseintolerant, er det best å spise en av de to alternativene i disse bildene:

1. **Chicken and Vegetable Stir-Fry:**
   - Chicken er en bra kilde til protein, men det er ikke laktose.
   - Vegetable Stir-Fry er full av fiber, vitaminer, og mineraler, og har ingen laktose.

2. **Pepperoni and Sausage Pizza:**
   - Pepperoni og sausage er ikke laktoseintolerant, men de har ikke så store mengder laktose som fullpizza.

Det er viktig å merke seg at disse alternativene ikke er full substitut for en fullpizza, men de er bedre enn fullpizza for en laktoseintolerant person.} \\
\textcolor{blue}{\small[English] \textit{VLM:} or a person who is lactose intolerant, it is best to eat one of the two options in these pictures:}

\textcolor{blue}{1. **Chicken and Vegetable Stir-Fry:**
   - Chicken is a good source of protein, but it's not lactose.
   - Vegetable Stir-Fry is full of fiber, vitamins, and minerals, and has no lactose.}

\textcolor{blue}{2. **Pepperoni and Sausage Pizza:**
   - Pepperoni and Sausage are not lactose intolerant, but they don't have as high amounts of lactose as full pizzas.}

{\textcolor{blue}{It's important to note that these options are not a full substitute for a full pizza, but they are better than full pizza for a lactose intolerant person.}}\\\\
\textbf{[Hungarian]} \\
\textit{User:} \foreignlanguage{hungarian}{Melyik ételt kellene enned a képeken látható ételek közül, ha laktózérzékeny vagyok?} \\
\textit{VLM:} \foreignlanguage{hungarian}{Ha laktózérzékeny vagy, akkor az ételt, amely nem tartalmaz laktózt, kellene enned. A laktózmentes alternatívaként a csirkét és a zöldségeket ajánlanám. A csirkét és a zöldségeket laktózmentesen is elkészítheted, és nem tartalmaznak laktózt.} \\
\textcolor{blue}{\small[English] \textit{VLM:} If you are lactose intolerant, you should eat the food that does not contain lactose. As a lactose-free alternative, I would recommend chicken and vegetables. You can also make chicken and vegetables lactose-free and they do not contain lactose. You can combine chicken and vegetables to get a delicious lactose-free meal.}
\end{tcolorbox}
\caption{Multilingual generation examples with the model trained with Text Regularization and merged with the original backbone LLM.}
\label{fig:gen4}
\end{figure*}

\subsection{Code Switching in Caption Generation}

Figure~\ref{fig:Codeswitching_examples} presents examples of code switching observed during caption generation for the Crossmodal-3600 dataset. We could identify two primary patterns emerge:

\begin{itemize} 
\item \textbf{Independent Words}: Certain technical or less common words are generated in English. 
\item \textbf{Language Alternation}: In some cases, once a word is switched to English, all subsequent words continue in English. 
\end{itemize}

\begin{figure*}[h!]
  \centering
  \begin{tcolorbox}[
    colback=gray!8,
    title={Generation Codeswitching Examples},
    fonttitle=\bfseries,
    coltitle=white,
    colbacktitle=gray!127,
  ]
  \begin{minipage}{.65\textwidth}
    \textit{Target Language: Spanish} \\
    Dos tacos con salsa y cilantro sobre una tortilla, \textbf{served in a paper wrapper and a lime wedge, with a white plate with a blue floral pattern underneath}.
    \end{minipage}
    \hfill
    \begin{minipage}{0.28\textwidth}
        \includegraphics[width=\linewidth]{images/tacos.pdf}
    \end{minipage}
    \\ \\
    \begin{minipage}{.65\textwidth}
    \textit{Target Language: Dutch} \\
    Het is een scenische boulevard met hotels aan de waterkant, een restaurant met \textbf{outdoor seating}, en een \textbf{sailboat} aan de promenade..
    \end{minipage}
    \hfill
    \begin{minipage}{0.28\textwidth}
        \includegraphics[width=\linewidth]{images/outdoor.pdf}
    \end{minipage}
    \\ \\
    \begin{minipage}{.65\textwidth}
    \textit{Target Language: Russian} \\
    «Postnord» \foreignlanguage{russian}{ — это название компании, которая занимается логистикой и дистрибуцией товаров}. \textbf{Company logo features bold, blue text on a wooden background, which gives a rustic and modern feel}.
    \end{minipage}
    \hfill
    \begin{minipage}{0.28\textwidth}
        \includegraphics[width=\linewidth]{images/postnord.pdf}
    \end{minipage}
    \\ \\
    \begin{minipage}{.65\textwidth}
    \textit{Target Language: German} \\
    In einem urbanen \textbf{Setting} steht ein orange VW Golf auf einem \textbf{cobblestone}. Die license plate \textbf{reads} \"HH 7293\", \textbf{and the car has a small antenna on the roof. A person stands behind the car, with a backpack on their back.}
    \end{minipage}
    \hfill
    \begin{minipage}{0.28\textwidth}
        \includegraphics[width=\linewidth]{images/coche.pdf}
    \end{minipage}

    %
    %
  \end{tcolorbox}
  \caption{Code switching examples observed when generating the captions of the Crossmodal-3600 dataset images using the Tr-3S-0.05x model. These examples showcase the need of adding a complementary evaluation apart from GlotLid's language detection in order to check language consistency. The words in English are shown in bold letters.}
  \label{fig:Codeswitching_examples}
\end{figure*}

\onecolumn

\begin{table*}[ht]

\centering
\begin{tabular}{llcc}
\toprule
\multicolumn{4}{c}{\textbf{Visual Data}} \\
\midrule
\textbf{Dataset} & \textbf{Field} & \textbf{Stage} & \textbf{Citation}  \\

\midrule
LLaVA Pretrain LCS-558K & Image Captions & 1 & \citet{llava} \\[0.2cm]
BLIP558K & Detailed Description & 1.5 & \citet{liu2024llavanext}  \\
CC3M & Detailed Description & 1.5 & \citet{liu2024llavanext} \\
COCO118K & Detailed Description & 1.5 & \citet{liu2024llavanext} \\
Evol Instruct & Math/Reasoning & 1.5 & \citet{chen2024allava} \\
UReader & OCR & 1.5 & \citet{ye-etal-2023-ureader} \\
SynthDOG & Language & 1.5 & \citet{kim2022donut} \\[0.2cm]
AI2D & Infographics & 2/2.5 & \citet{ai2d} \\
Allava Instruct & General & 2/2.5  & \citet{chen2024allava} \\
AOKVQA & General & 2/2.5  & \citet{schwenk2022okvqa} \\
Cambrian (filtered) & General & 2/2.5  & \citet{tong2024cambrian} \\
Chart2Text & Infographics & 2/2.5  & \citet{obeid2020chart} \\
ChartQA & Infographics & 2/2.5  & \citet{masry2022chartqa} \\
ChromeWriting & OCR & 2/2.5  & - \\
CLEVR & General & 2/2.5  & \citet{johnson2017clevr} \\
CLEVR-Math & Math/Reasoning & 2/2.5  & \citet{johnson2017clevr} \\
COCO Caption & General & 2/2.5  & \citet{lin2014microsoft} \\
Diagram Image2Text & Infographics & 2/2.5  & - \\
DocVQA & Infographics & 2/2.5  & \citet{mathew2021docvqa} \\
DVQA & Infographics & 2/2.5  & \citet{kafle2018dvqa} \\
FigureQA & Infographics & 2/2.5  & \citet{kahou2017figureqa} \\
GQA & Math/Reasoning & 2/2.5  & \citet{hudson2019gqa} \\
Geo170K Align & Math/Reasoning & 2/2.5  & \citet{gao2023g} \\
Geo170K QA & Math/Reasoning & 2/2.5  & \citet{gao2023g} \\
Geo3K & Math/Reasoning & 2/2.5  & - \\
Geometry3K & Math/Reasoning & 2/2.5  & \citet{lu2021inter} \\
GeoMVerse & Math/Reasoning & 2/2.5  & \citet{kazemi2024geomverse} \\
GeoQA+ & Math/Reasoning & 2/2.5  & \citet{chen2021geoqa} \\
GEOS & Math/Reasoning & 2/2.5  & \citet{seo2015solving} \\
Hateful Memes & General & 2/2.5  & \citet{kiela2020hateful} \\
HiTab & Infographics & 2/2.5  & \citet{cheng2022hitab} \\
HME100K & OCR & 2/2.5  & \citet{yuan2022syntax} \\
IAM & OCR & 2/2.5  & \citet{marti2002iam} \\
IconQA & General & 2/2.5  & \citet{lu2021iconqa} \\
IIIT5K & OCR & 2/2.5  & \citet{mishra2012scene} \\
Infographic VQA & Infographics & 2/2.5  & \citet{mathew2022infographicvqa} \\
InterGPS & General & 2/2.5  & \citet{lu2021inter} \\
Image Textualization & General & 2/2.5  & \citet{pi2024image} \\
K12 Printing & OCR & 2/2.5  & - \\
LLaVA-158K & General & 2/2.5  & \citet{llava} \\
LLaVA-Wild (train) & General & 2/2.5  & \citet{llava} \\
LLaVAR & General & 2/2.5  & \citet{zhang2023llavar} \\
LRV-Chart & Infographics & 2/2.5  & \citet{liu2023mitigating} \\
LRV-Normal & Math/Reasoning & 2/2.5  & \citet{liu2023mitigating} \\
Magpie Pro & Language & 2/2.5  & \citet{xu2024magpie} \\

\bottomrule
\end{tabular}
\caption*{\hfill \textit{(Continues on next page...)}}
\label{tab:data-sources-text}
\end{table*}

\begin{table*}[ht]

\centering
\begin{tabular}{llcc}
\toprule
\textbf{Dataset} & \textbf{Field} & \textbf{Stage} & \textbf{Citation}  \\

\midrule

MapQA & Math/Reasoning & 2/2.5  & \citet{chang2022mapqa} \\
MathQA & Math/Reasoning & 2/2.5  & \citet{amini2019mathqa} \\
MAVIS & Math/Reasoning & 2/2.5  & \citet{zhang2024mavis} \\
OKVQA & General & 2/2.5  & \citet{marino2019ok} \\
OCR-VQA & OCR & 2/2.5  & \citet{mishra2019ocr} \\
RAVEN & Math/Reasoning & 2/2.5  & \citet{zhang2019raven} \\
RefCOCO & General & 2/2.5  & \citet{yu2016modeling} \\
Rendered Text & OCR & 2/2.5  & - \\
RoBUT & Infographics & 2/2.5  & \citet{zhao-etal-2023-robut} \\
ScienceQA & General & 2/2.5  & \citet{lu2022learn} \\
Screen2Words & Infographics & 2/2.5  & \citet{wang2021screen2words} \\
ShareGPT4O & General & 2/2.5  & \citet{cui2025comprehensive} \\
ShareGPT4V & General & 2/2.5  & \citet{chen2024sharegpt4v} \\
ST-VQA & General & 2/2.5  & \citet{biten2019scene} \\
Super-CLEVR & Math/Reasoning & 2/2.5  & \citet{li2023super} \\
TabMWP & Math/Reasoning & 2/2.5  & \citet{lu2023dynamic} \\
TallyQA & General & 2/2.5  & \citet{acharya2019tallyqa} \\
TextCaps & OCR & 2/2.5  & \citet{sidorov2020textcaps} \\
TextOCR-GPT4 & OCR & 2/2.5   & \citet{textocr-gpt4v}\\
TQA & Infographics & 2/2.5  & \citet{kembhavi2017you} \\
UniGeo & Math/Reasoning & 2/2.5  & \citet{chen2022unigeo} \\
Ureader & Infographics & 2/2.5  & \citet{ye-etal-2023-ureader} \\
Vision FLAN & General & 2/2.5  & \citet{xu2024vision} \\
Visual7W & General & 2/2.5  & \citet{zhu2016visual7w} \\
Visual Genome & Math/Reasoning & 2/2.5  & \citet{krishna2017visual} \\
VisText & General & 2/2.5  & \citet{tang2023vistext} \\
VisualMRC & Infographics & 2/2.5  & \citet{tanaka2021visualmrc} \\
VizWiz & General & 2/2.5  & \citet{gurari2018vizwiz} \\
VQARAD & General & 2/2.5  & \citet{lau2018dataset} \\
VQAv2 & General & 2/2.5  & \citet{antol2015vqa} \\
VSR & General & 2/2.5  & \citet{llava} \\
WebSight & General & 2/2.5  & \citet{lauren2024unlocking} \\[0.2cm]
Spot-the-Diff & Multi-Image & 2.5  & \citet{jhamtani2018learning} \\
Birds-to-Words & Multi-Image & 2.5  & \citet{forbes2019neural} \\
CLEVR-Change & Multi-Image & 2.5  & \citet{park2019robust} \\
HQ-Edit-Diff & Multi-Image & 2.5  & \citet{hui2024hq} \\
MagicBrush-Diff & Multi-Image & 2.5  & \citet{zhang2023magicbrush} \\
IEdit & Multi-Image & 2.5  & \citet{tan2019expressing} \\
AESOP & Multi-Image & 2.5  & \citet{ravi2021aesop} \\
FlintstonesSV & Multi-Image & 2.5  & \citet{gupta2018imagine} \\
PororoSV & Multi-Image & 2.5  & \citet{li2019storygan} \\
VIST & Multi-Image & 2.5  & \citet{huang2016visual} \\
WebQA & Multi-Image & 2.5  & \citet{chang2022webqa} \\
TQA (MI) & Multi-Image & 2.5  & \citet{kembhavi2017you} \\
OCR-VQA (MI) & Multi-Image & 2.5  & \citet{mishra2019ocr} \\
DocVQA (MI) & Multi-Image & 2.5  & \citet{mathew2021docvqa} \\

\bottomrule
\end{tabular}
\caption*{\hfill \textit{(Continues on next page...)}}
\label{tab:data-sources-text}
\end{table*}

\begin{table*}[ht]

\centering
\begin{tabular}{llcc}
\toprule

\textbf{Dataset} & \textbf{Field} & \textbf{Stage} & \textbf{Citation}  \\
\midrule

MIT-StateCoherence & Multi-Image & 2.5  & \citet{isola2015discovering} \\
MIT-PropertyCoherence & Multi-Image & 2.5  & \citet{isola2015discovering} \\
RecipeQA ImageCoherence & Multi-Image & 2.5  & \citet{yagcioglu2018recipeqa} \\
VISION & Multi-Image & 2.5  & \citet{bai2023vision} \\
Multi-VQA & Multi-Image & 2.5  & \citet{li2023fine} \\
IconQA & Multi-Image & 2.5  & \citet{lu2021iconqa} \\
Co-Instruct & Multi-Image & 2.5  & \citet{wu2024towards} \\
DreamSim & Multi-Image & 2.5  & \citet{fu2023dreamsim} \\
ImageCoDe & Multi-Image & 2.5  & \citet{krojer2022image} \\
nuScenes & Multi-Image & 2.5  & \citet{caesar2020nuscenes} \\
ScanQA & Multi-Image & 2.5  & \citet{azuma2022scanqa} \\
ALFRED & Multi-Image & 2.5  & \citet{shridhar2020alfred} \\
ContrastCaption & Multi-Image & 2.5  & \citet{jiang2024mantis} \\
VizWiz (MI) & Multi-Image & 2.5  & \citet{gurari2018vizwiz} \\
ScanNet & Multi-Image & 2.5  & \citet{dai2017scannet} \\
COMICS Dialogue & Multi-Image & 2.5  & \citet{iyyer2017amazing} \\
NLVR2 & Multi-Image & 2.5  & \citet{suhr2019corpus} \\[0.2cm]
NExT-QA & Video & 2.5 & \citet{xiao2021next} \\
Ego-4D   & Video  & 2.5 & \citet{Grauman_2022_CVPR} \\
YouCook2& Video & 2.5 & \citet{ZhXuCoCVPR18} \\
ActivityNet   & Video  & 2.5 & \citet{yu2019activityqa} \\
Charades  & Video & 2.5 & \citet{charades} \\
ShareGPT4Video & Video & 2.5 & \citet{chen2024sharegpt4video} \\

\bottomrule
\end{tabular}
\caption{English only visual datasets used throughout this work. The same data as proposed in LLaVA-OneVision has been used.}
\label{tab:visual-data}
\end{table*}

\begin{table*}[ht]

\centering
\begin{tabular}{llcc}
\toprule
\multicolumn{4}{c}{\textbf{Text-Only Data}} \\
\midrule
\textbf{Dataset} & \textbf{Field} & \textbf{Stage} & \textbf{Citation}  \\

\midrule
Aya Dataset & General & 1.5/2/2.5  & \citet{singh-etal-2024-aya}  \\
CoqCat & Conversation QA & 1.5/2/2.5  & \citet{gonzalez-agirre-etal-2024-building}  \\
Databricks Dolly 15k & General & 1.5/2/2.5  & \citet{DatabricksBlog2023DollyV2}  \\
Databricks Dolly 3k CA & General & 1.5/2/2.5  & -   \\
FLORES-200 (Instructions) & Translations & 1.5/2/2.5  & \citet{costa2022no}  \\
MentorCA & General & 1.5/2/2.5  & -  \\
No Robots & General & 1.5/2/2.5  & \citet{no_robots}  \\
OASST & General & 1.5/2/2.5  & \citet{kopf2023openassistant}  \\
OASST-CA & General & 1.5/2/2.5  & -  \\
RAG Multilingual & General & 1.5/2/2.5  & -  \\
Tower-Blocks-v0.1 & Text-Insight & 1.5/2/2.5  & \citet{tower_llm_2024}  \\

\bottomrule
\end{tabular}
\caption{Multilingual text-only datasets added throughout the visual instruction process.}
\label{tab:data-sources-text}
\end{table*}

\twocolumn

\end{document}